\documentclass{article}
\usepackage{a4wide,times,amsmath, amssymb, tipa, myexamples, chicago,theorem, xspace,float,afterpage,multicol}
\usepackage[dvips]{epsfig,rotating}
\usepackage[normalem]{ulem}

\usepackage{xy}

\xyoption{all}

\CompileMatrices  

\newcommand{\ip}[1]{\textipa{#1}}
\newcommand{\gl}[1]{`{\it #1}'}
\newcommand{\finallongpage}{\enlargethispage{\baselineskip}}

\newcommand{\finalpagebreak}{\pagebreak}
\newcommand{\finalforcedpage}{\enlargethispage*{100cm}}

\theoremstyle{break}

\newcommand{\url}[1]{{\footnotesize\texttt{#1}}}

\newcommand{\ASSIGN}{$\leftarrow\,$\/}
\newcommand{\INPUT}{{\bf input }}
\newcommand{\CONDITION}{{\bf condition }}
\newcommand{\OUTPUT}{{\bf output }}
\newcommand{\REPEAT}{{\bf repeat }}
\newcommand{\IF}{{\bf if }}
\newcommand{\THEN}{{\bf then }}
\newcommand{\ELSE}{{\bf else }}
\newcommand{\ENDIF}{{\bf endif }}
\newcommand{\UNTIL}{{\bf until }}
\newcommand{\UNEQUAL}{$\neq\,$\/}
\newcommand{\AND}{$\wedge\,$\/}
\newcommand{\NOT}{$\neg\,$\/}

\newcommand{\bigemph}[1]{{\sc #1}}
\newcommand{\ling}[1]{\emph{#1}}
\newcommand{\lingnull}{$\varnothing${}}
\newcommand{\strich}{\rule{.99\linewidth}{.1pt}\\}
\newcommand{\startpiece}{\par\noindent\strich\vspace{-1.6\baselineskip}}
\newcommand{\stoppiece}{\vspace{-\baselineskip}\noindent\strich\par\noindent}

\restylefloat{figure}

\newenvironment{Ex}
{\begin{figure}[H]\begin{examples}}{\end{examples}\end{figure}}

\def\sref#1{\S\ref{#1}}
\def\code#1{{\tt #1}}
\begin{document}
\thispagestyle{empty}
\begin{center}
{\sffamily\bfseries\scshape\huge  Marburger Arbeiten zur Linguistik} \\[6mm]
\epsfig{file=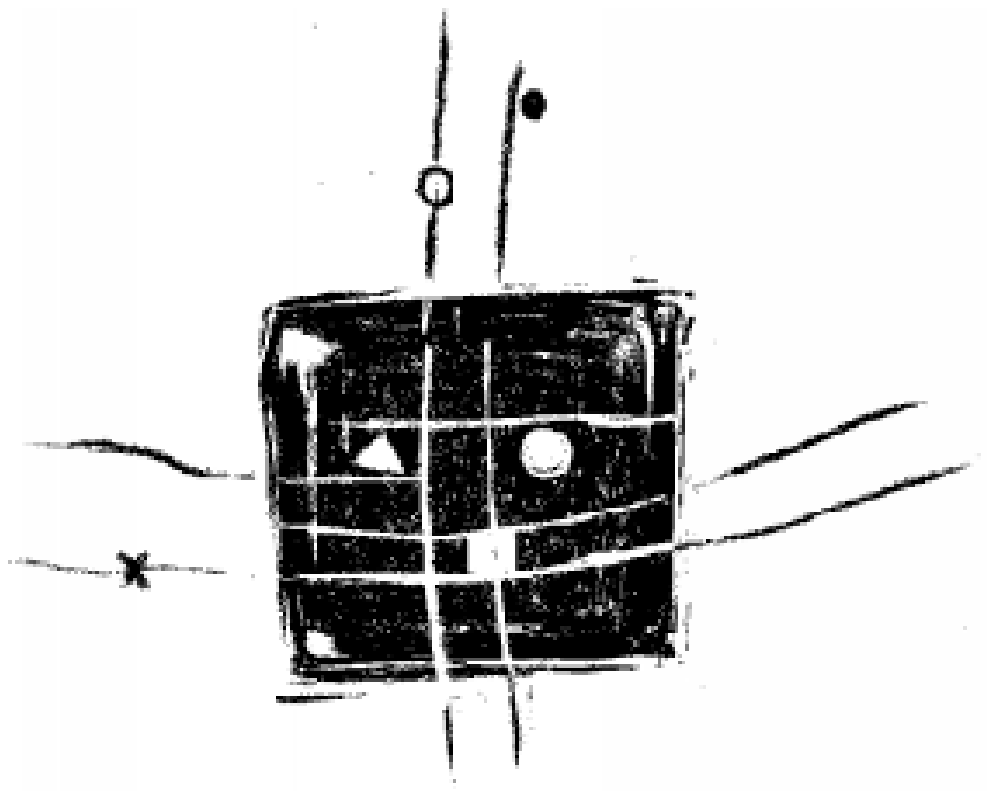}%
\\[6mm]
{\sffamily\huge
Nr. 1} \\[12mm]
{\sffamily\bfseries\huge  One-Level Prosodic Morphology} \\[12mm]
{\sffamily\huge Markus Walther} \\[15mm]
{\sffamily\large
November 1999 \\[17mm]

Institut f\"ur Germanistische Sprachwissenschaft \\
Philipps-Universit\"at Marburg \\
Wilhelm-R\"opke-Str. 6A \\
D-35032 Marburg \\[4mm]
e-mail: Markus.Walther@mailer.uni-marburg.de\\[4mm]
\underline{\phantom{XXXX}}\\[4mm]
 {\sffamily\bfseries\large archive: http://www.uni-marburg.de/linguistik/mal}}
\end{center}
\newpage
\thispagestyle{empty}
{\Large
\begin{center}
{\bf Abstract}
\end{center}

\noindent Recent developments in theoretical linguistics have lead to a
widespread acceptance of constraint-based analyses of prosodic morphology
phenomena such as truncation, infixation, floating morphemes and reduplication. Of these, 
reduplication is particularly challenging for state-of-the-art
computational morphology, since it involves copying of some part of a phonological string.  
 In this paper I argue for certain extensions to the one-level model
 of phonology and morphology (Bird \& Ellison 1994)   
to cover the computational aspects of prosodic morphology using finite-state 
methods. In a nutshell, enriched lexical representations provide
additional automaton arcs to repeat  or skip sounds and also to allow insertion of additional 
material. A kind of resource consciousness is introduced to control this 
additional freedom, distinguishing between producer and consumer
arcs. The non-finite-state copying aspect of reduplication is 
mapped to automata intersection, itself a non-finite-state operation.
Bounded local optimization prunes certain automaton arcs that fail to
contribute to linguistic optimisation criteria such as leftmostness of an infix within the word.
The paper then presents implemented case studies of Ulwa construct
state infixation, German hypocoristic truncation and Tagalog
overapplying reduplication that illustrate the expressive power
of this approach, before its merits and limitations are discussed and
possible extensions are sketched.
I conclude that the one-level approach to prosodic morphology presents
an attractive way of extending finite-state techniques to difficult
phenomena that hitherto resisted 
elegant computational analyses.
}
\newpage
\thispagestyle{empty}
\tableofcontents
\newpage
\setcounter{page}{1}
\section{Introduction}
Prosodic morphology is the study of natural language phenomena in
which the shape of words is to a 
major extent determined by 
{\em phonological} factors such as obedience to wellformed syllable
or foot structure, adjacency to stress peaks or sonority
extrema etc.%
\footnote{This work has been funded by the German research agency DFG
  under grant WI 853/4-1. I am particularly indebted to T.\/ Mark Ellison,
  who generously shared his ideas with me during a visit to Krakow. Thanks also to Richard
Wiese and Steven Bird for helpful comments and encouragement. All
remaining errors and shortcomings are mine.}%
Introductory examples from truncation, infixation,
root-and-pattern morphology and reduplication are given in \ref{intro:examples}.
\begin{Ex}
\item \label{intro:examples}
\begin{tabular}[t]{p{\linewidth}}
a. {\em German: hypocoristic i-truncation } \\[1ex] %&
 Pet{\bf ra} $>$ Pet-i, And{\bf rea} $>$ And-i,
Gorb{\bf atschow} $>$ Gorb-i \\[1ex]
b. {\em Ulwa:  construct state suffixation/infixation} \\[1ex] %& 
b\'as-{\bf ka} `his hair',
\'as-{\bf ka}-na `his clothes', kar\'as-{\bf ka}-mak `his knee' \\[1ex]
c. {\em Modern Hebrew:  vowel/\lingnull{} alternation in
  nonconcatenative verbal morphology} %
\\[1ex] %
g{\bf a}m{\bf a}r
`finished' (3sg.m), g{\bf a}mr-a (3sg.f), \newline yi-gm{\bf o}r `will finish' (3sg.m),
yi-gm{\bf e}r-u (3pl) \\[1ex]
d. {\em Mangarayi: reduplicated plural etc.} \\[1ex] %
gabuji $>$ gab{\bf ab}uji `old 
people', jimgan $>$ jimg{\bf img}an `knowledgeable people', \newline muygji $>$ 
muygj{\bf uygj}i `having a dog' %
\end{tabular}
\end{Ex}
Prosodic conditions govern the cutoff point in 
German hypocoristic truncation, serve to determine the placement of the floating 
\ling{-ka-} suffix/infix within Ulwa possessive noun constructions, control the deletion %
 of
vowels in Modern Hebrew nonconcatenative verb morphology and determine 
position and length of the reduplicant copy in Mangarayi plurals. We
will see later what some of these conditions are and how to devise computational analyses for
such phenomena.

In his extensive overview of the state of the art in computational
morphology, \citeN{sproat:92} provides ample indication that there is
still work to do with regard to these phenomena. Here is a relevant
sample of Sproat's comments:
\begin{quote}
Subtractive morphology -- presumably since it is relatively infrequent 
-- has attracted no attention. [p.170] \\
\dots computational models have been only partly successful at
analyzing infixes. [p.50] \\
From a computational point of view, one point cannot be overstressed:
the copying required in reduplication places reduplication in a class
apart from all other morphology. [p.60]\\
\dots a morphological analyzer needs to use information about prosodic 
structure. [p.170] \\
\dots there is a tendency in some quarters of the computational
morphology world to trivialize the problem, suggesting that the
problems of morphology have essentially been solved simply because
there are now working systems that are capable of doing a great deal
of morphological analysis \dots On should not be misled by such claims 
\dots there are still outstanding problems and areas which have not
received much serious attention \dots [p.123]\\
\end{quote}
There is still truth in Sproat's words seven
years after they were written.

The primary goal of this paper is therefore to start answering the challenge
posed by Sproat's comments and show how central linguistic %
insights into the way prosodic morphology works translate into an
implemented finite-state model.
The model is named \bigemph{One-Level Prosodic Morphology}
(%
OLPM), because it was developed as an
adaptation of One-Level Phonology 
(\citeNP{bird.ellison:94}, OLP) to  %
prosodic morphology. Despite some
initial attempts, it is probably fair to say that prosodic morphology as a branch of theoretical
linguistics is still rather underformalized. While \citeN[\S 5.4]{bird.ellison:94} does
contain a very brief autosegmental analysis of just the Arabic verbal stem \ling{kattab}
 -- itself a legitimate piece of prosodic morphology --, we will see that the
extensions introduced in this paper can be justified on the basis of
a broader view of what prosodic morphology encompasses. 
The model shares the essential assumption of monostratality with
One-Level Phonology, maintaining the restrictive postulate that there
should  only be {\em one level} of linguistic description. It thus
inherits two key advantages of the former model, namely easy  
integrability into monostratal frameworks of grammar such as HPSG
\cite{pollard.sag:94} and simplified machine
learning of surface-only generalizations \cite{ellison:92}.  
By furthermore retaining the finite-state
methodology of its predecessor it keeps what is computationally attractive
about the former model's properties. Combining these essential traits, 
it is clear that  OLPM  -- like OLP -- will be based on finite-state
automata (FSA) with their characteristic combinatory operation of
regular set intersection, also known as automaton product. Thus, it contrasts with models 
employing two, three or more levels (\citeNP{koskenniemi:83},
  \citeNP{touretzky.wheeler:90}, \citeNP{chomsky.halle:68}), which are
  usually implemented with finite-state transducers (FSTs) and
  characteristically employ a composition operation to combine
  individual transducers into a single overall  mapping.

The present work differs, however, in not following the specific
representational assumptions of nonlinear autosegmental phonology
\cite{goldsmith:90} embodied in OLP, using a flat
prosodic-segmental representation instead. This difference is not crucial to the task at
hand. It mainly stems from our desire to %
avoid an additional layer of
complexity that promises little return value for the immediate goal stated
above. Needless to say, however, a representational variant of OLPM using autosegmental
diagrams would not seem to be unthinkable. 

There is another difference. Existing research in
nonconcatenative finite-state morphology has primarily been concerned with
templatic aspects of Semitic languages
(e.g.\/ \citeNP{kataja.koskenniemi:88} for
Akkadian, \citeNP{beesley.buckwalter.newton:89}, \citeNP{beesley:96}
and \citeNP{kiraz:94}, \citeNP{kiraz:96b} for Arabic and 
Syriac). Alternative (feature-)logical treatments share this phenomenological
bias (\citeNP{bird.blackburn:91} on Arabic, \citeNP{klein:93} on
Sierra Miwok, \citeNP{walther:97} and \citeNP{walther:98} on Tigrinya and 
Modern Hebrew). In contrast, our focus
is on the difficult rest of prosodic morphology, where infixation,
circumfixation, truncation and in particular reduplication have not
received elegant computational analyses so far. 

The paper is organized as follows. Section \ref{back} provides some
background to the emergence of constraint-based models of prosodic
morphology, while \ref{problem} lays out the range of data to be
accounted for. As the presentation unfolds, a number of desiderata for 
formalization and implementation are formulated. The central proposals
of the paper are contained in the following section \ref{ext}, where I show
which representational assumptions must be made and which new devices
need to be incorporated into a comprehensive %
one-level model of prosodic morphology. 
In section \ref{ex} the proposals are 
evaluated in practice by developing detailed implementations of Ulwa
infixation, German truncation and Tagalog reduplication phenomena.
Section \ref{disc} concludes with a discussion of these proposals, evaluating
their merits both on internal grounds %
and in comparison to other works%
.

\section{Background}\label{back}
Since the beginning of the seventies it has been recognized that
rule-based models of prosodic morphology lack explanatory adequacy, a
fact that has come to be known as the `rule conspiracies' problem
\cite{kisseberth:70a}. Kisseberth used the vowel/zero alternation
patterns from inflected verb forms in Tonkawa to make his point \ref{back:tonkawa}. 
Tonkawa is an extinct Coahuiltecan language with CV(C) syllable structure.
\begin{Ex}
\item \label{back:tonkawa} {\sc Tonkawa verb forms}\\[1ex]
\begin{tabular}[t]{lll}
`{\em to cut}'  & `{\em to lick}'  %
\\ \cline{1-2}
\#picn-o\textglotstop\#      & \#netl-o\textglotstop\#  & %
{\footnotesize\it 3sg.obj.stem-3sg.subj.} \\  
\#we-pcen-o\textglotstop\#   & \#we-ntal-o\textglotstop\# & %
{\footnotesize\it 3pl.obj.-stem-3sg.subj.} \\
\#picna-n-o\textglotstop\#  & \#netle-n-o\textglotstop\#  &  %
{\footnotesize\it 3sg.obj.stem-prog.%
} 
 \\ \cline{1-2}
p(i)c(e)n(a)   & n(e)t(a)l(e)  & {\footnotesize\it stems}
\end{tabular}
\end{Ex}
While trying to incorporate more and more affixation patterns, Kisseberth
observed that the usual enlarge-corpus/modify-analysis cycle 
resulted in increasingly baroque levels of
complexity for the vowel deletion rules.%
\footnote{Witness e.g. the rewrite rule $ \begin{bmatrix} V \\ {+}stem \end{bmatrix}\,\rightarrow\,
  \varnothing\, /\, \left\{ \begin{matrix} V\,+\,C \\ \left\{
          \begin{matrix} \# \\ C{+} \end{matrix}\right\} CVC
    \end{matrix}\right\} \underline{\phantom{XX}} C \begin{bmatrix} V
    \\ {+}stem\end{bmatrix}$ from \citeN{kisseberth:70a}.}
 But
more to the point, these rules `conspired' to  
maintain a very simple, yet {\em global} invariant --
sequences of three consecutive consonants are banned on the
surface, symbolically $\ast$CCC (the word boundary \# acts as a
consonant). The reader may easily verify that no further deletion of vowels
is possible in \ref{back:tonkawa} without violating the
invariant. Under the rule-based analysis, however, this  
global condition is nowhere expressed directly. According to Kisseberth the failure can
be traced back to a defect of the derivational paradigm itself:
By design each rule only sees the input given to it by a prior rule
application in the rule composition cascade, being  blind to the
ultimate output consequences that surface at the end. Later
developments have decomposed segment-level constraints such as $\ast$CCC by
referring to the {\em prosodic} concept of the syllable instead -- complex syllable 
onsets and codas are disallowed in Tonkawa core syllables. At least since \citeN{kahn:76} and
\citeN{selkirk:82} this idea of an independent level of syllable structure 
superseded the SPE \cite{chomsky.halle:68} conception of purely segmental
strings in the generative literature. While the trend towards
representationalism in phonology, marked by the advent of such frameworks as
Autosegmental and Metrical Phonology (see \citeNP{goldsmith:90} for an
overview), reduced the reliance on rules and further strengthened the role of 
prosodic structure in actual analyses, the fundamental defect that Kisseberth and
others had recognized in derivational theories still awaited a
principled solution. 

That solution emerged at the beginning of the
nineties, as constraint-based models of phonology were proposed to
directly capture the missing `output orientation' that plagued its
derivational predecessors. \citeN{bird:90} and \citeN{scobbie:91} were 
among the first to use monotonic formal description languages to express surface-true
{\em non-violable constraints}, defining what has now become known
collectively as Declarative Phonology (DP; \citeNP{scobbie.et.al:96}). In DP, both lexical items
and more abstract generalizations are constraints, with constraint
conjunction being the characteristic device for formalizing constraint interaction. Shortly
thereafter \citeN{prince.smolensky:93} 
 argued for ranked {\em violable constraints} instead. Their proposal was
 named Optimality Theory (OT) and has since %
become  a much-recognized new paradigm in theoretical phonology and
 beyond.

 In OT constraints seek to capture
 conflicting universal tendencies while  strict ranking 
 imposes an extrinsic ordering relation on the set of constraints, 
 expressing which one takes precedence for purposes of conflict
 resolution. According to the OT ideal, languages differ only in how
 they rank the common pool of constraints.
 Strictness of ranking means that, in contrast to arbitrarily weighted grammars, no amount of
 positive wellformedness of an input with respect to  lower-ranked constraints can
 compensate for illformedness due to a higher-ranked constraint. Finally,
 although constraints may be gradiently violated by the set of structurally
 enriched candidates that is generated from the input, only candidates with the minimal
 number of violations are designated as grammatical.
Note that, because of this powerful mechanism of global optimization,
the OT analyst is free to propose constraints that are {\em never}
surface-true (an example would be the excessive-structure-minimizer
$\ast$STRUC `Avoid structure', \citeNP[ch.3,
fn.13]{prince.smolensky:93}; see \citeNP[13]{walther:96} for a formalization).

While DP paid considerable attention to proper formalization of 
phonology, the empirical domain of prosodic morphology so far has %
received much less attention than in OT, where the co-appearance of
\citeN{mccarthy.prince:93} with \citeN{prince.smolensky:93} marked the 
beginning of a continuous involvement with the subject. 
In particular, certain problems in the prosodic morphology of
reduplication motivated an extension to classical OT known as
Correspondence Theory \cite{mccarthy.prince:95a}. 
Here constraints are allowed to simultaneously refer to both levels of
two-level pairings for assessing gradient wellformedness, in what
appears to be rather analogous to two-level morphology \cite{koskenniemi:83}.
However, the range of correspondence-theoretic mappings --
mediated by some abstract indexation scheme -- goes beyond
Koskenniemi's original framework in that it includes
intra-level instances such as base-reduplicant correspondence with
the same word level in addition to classical cross-level mappings like
so-called input-output (i.e lexical-surface) correspondence.

Despite this growing body of
theoretical work the
average level of formalization in concrete %
OT {\em analyses} has been rather low%
\footnote{Notable exceptions include \citeN{albro:97}
  on Turkish vowel harmony, \citeN{eisner:97a} on stress systems,
  \citeN{ellison:94} on Arabic glottal stop distribution, \citeN{walther:96} on %
truncated plurals in Upper Hessian.}%
, which is a genuine problem in the light of \citeN[4]{chomsky:65}'s
definition of a generative grammar as one that must be  ``perfectly explicit'' and ``not rely on the
intelligence of the understanding reader''.%
\footnote{Cf. also the verdict of \citeN[537]{pierrehumbert.nair:96},
  who write: ``Any attempt to argue for a particular method of combining
constraints without simultaneously formalizing the constraints is
technically incoherent.''}

Note that this is not due to a lack of proposals for formalizing the abstract {\em paradigm}
of (classical) OT itself, where \citeN{ellison:94},
\citeN{walther:96}, \citeN{eisner:97}, \citeN{frank.satta:98},
and \citeN{karttunen:98} have all made contributions within 
the framework of finite-state systems. Karttunen's work
 is especially worth mentioning, because he was able to show that, under regularity
assumptions for both the constraints and the input set, optimal
outputs can be computed using only established finite-state operations 
such as transducer composition and automaton complement, without any overt
optimization component. The impact of his results is 
that it places SPE-style rule cascades and OT-style constraint rankings
into rather close proximity. Both represent ways to define
finite-state transducers, albeit with very different high-level
specification languages. Again the assumption of an underlying finite-state
architecture was the key factor in establishing Karttunen's findings. 

Given the state of the field outlined above, it seems particularly
attractive now to combine these separate 
strands of research. More precisely, the desire is to produce a model
that can (i) capture theoretical insights into the analytical requirements
of prosodic morphology without (ii) unduly compromising in the area of proper
formalization while  (iii) still ensuring effective computability with the
help of finite-state methods. To this end I chose to extend One-Level Phonology, itself a
finite-state incarnation of DP, than fleshing out one of the proposals for OT. 

The principal reason for this choice has to do with the
fact that there are much better prospects for {\em automatic constraint
acquisition} than for its two-level competitors. Corpora usually
contain the surface phonological form of words only and do not come
equipped with pairs of surface {\em and} abstract underlying representations
(SRs and URs). Theoretical linguistics offers no help either, as e.g.\/
\citeN[\S 3.4]{kenstowicz:94} argues at length that no {\em a priori} restriction on 
possible URs suffices for all scenarios. 
The OT notion of `lexicon optimization' \cite{prince.smolensky:93}, while meant to address the
same problem of deriving suitable URs, is still too vague to merit closer attention.
This lack of  either a principled or a natural, non-handcrafted source
for two-level pairings means that there may be an 
arbitrarily large gap to bridge when trying to infer a finite-state
mapping SR $\leftrightarrow$ UR from  surface-only data. It is thus no  
accident that e.g.\/ the results of \citeN{ellison:92} on learning a number
of phonological properties in a typologically balanced sample of 30 
languages were obtained by using one-level FSAs for the representation 
of inferred generalizations. Other results from the literature on machine learning 
of natural language seem to confirm this key advantage of monostratality
(e.g. \citeNP{belz:98}).
For OT, on the other hand, no substantial result is known that adresses the hard 
problem of {\em constraint}  acquisition. The prospects of remedying
this %
situation without giving up elementary OT premises are not particularly good.
As noted above, constraints are formally unrestricted and not required
to be surface true for at least some pieces of data.
Existing results on OT-based learning only deal with the much simpler problem of
inferring the {\em ranking} of constraints
given pairs of structurally annotated outputs {\em and } inputs
(\citeNP{tesar.smolensky:93}, \citeNP{boersma:98}, \citeNP{boersma.hayes:99}). 
This is probably also
due to the fact that orthodox OT itself expresses disinterest in the
question by assuming that all constraints are already given as part of
Universal Grammar.%
\footnote{But see \citeN{ellison:to_appear} for strong arguments
against the universalist interpretation of OT and \citeN{hayes:99} for initial attempts
at phonetically grounded constraint induction.}

A second reason for extending OLP is that the lack of arbitrary mapping between levels
plus the lack of global optimization forces
a healthy reexamination of existing analytical devices. Maintaining
the  restrictive set of assumptions embodied in OLP  often
leads to the discovery of new surface-true generalizations. 
Starting with the one-level approach is more illuminating for investigating the precise
nature of the tradeoff  between mono- or polystratal
analyses. Finally, this point of departure promises  
better answers to the question of which set of theory extensions is 
absolutely necessary in order to cover the
enlarged range of empirical phenomena under study. In what follows we will 
 see that this approach of `starting small' indeed yields some of these expected payoffs.
\section{The Problem}\label{problem}
In this section, I want to give a brief survey of some of the %
challenges of prosodic morphology. The focus will be on what the relevant
linguistic data suggest about minimal requirements for formalization and implementation.
The topics to be presented in turn will be reduplication, discontiguity and partial
realization of morphemes as well as cases of `floating', i.e.\/ positionally
underspecified morphemes together with their directional behaviour.
\subsection{Reduplication} \label{prob:reduplication}
Let us begin with some terminology. The original part of a word from
which reduplication copies will be called the {\em base}, while the
copy is also referred to as the {\em reduplicant}. The next bits of terminology
arise as we further classify reduplicative constructions below. The
impetus behind this classification is that it rules out some easy ways
of avoiding the full complexity of the reduplication problem.  

One such classificatory subdivision is
between {\em total} reduplications and {\em partial} ones. The former
are defined to be isomorphic to the formal language $ww$, known to be
context-sensitive, while the latter exhibit imperfect copying of one
sort or another. A frequent case of  imperfection has a truncated
portion of the base as the reduplicant. Furthermore, there 
are {\em unmodified} and {\em modified} reduplications, where in the
latter case reduplicant and base differ in the applicability of
phonological alternations. In contrast to {\em exfixing}
reduplications, where the reduplicant is a prefix or suffix, in the {\em infixing}
variant the reduplicant interrupts the immediate adjacency
relationships of the base. Also, there are {\em discontiguous}
reduplicants in addition to the more usual {\em contiguous}
ones. Figuratively speaking, discontiguity means that some segments of the base are skipped
over in constructing the reduplicant. Whereas some of the most well-known 
reduplication instances like Indonesian plural are {\em unbounded} in the sense that 
the reduplicant length is a linear function of the length of the base, 
in {\em bounded} types of reduplication  a finite, %
and often rather small, upper bound can be placed on the length of the
reduplicant. Finally, there are {\em purely reduplicative}
constructions versus their {\em fixed melody}-enriched counterparts. The former 
have reduplicants which are entirely constructed from copied base
material (possible modified in the above sense), whereas the latter
also contain base-independent segmental material as part of the construction.

Table \ref{prob:redex} shows how constructions from four languages  instantiate
the classificational scheme outlined above, illustrating each
opposition with at least one construction. Actual examples
are supplied in \ref{prob:madurese} --
\ref{prob:koasati}. Reduplicants are marked with bold face and
subscripts mark base-reduplicant correspondences where necessary.
\begin{Ex}
\item \label{prob:redex}
\begin{tabular}[t]{|l||p{1cm}|p{1cm}||p{1cm}|p{1cm}|p{1cm}|p{1cm}|}
\hline {\em Language} & \begin{turn}{90}\parbox{7em}{\scriptsize total (+)
    \newline partial (--)} \end{turn} &
\begin{turn}{90}\parbox{7em}{\scriptsize unmodified (+) \newline modified (--)}\end{turn} &
\begin{turn}{90}\parbox{7em}{\scriptsize exfixing (+) \newline  infixing (--)}\end{turn} &
\begin{turn}{90}\parbox{7em}{\scriptsize contiguous (+) \newline  discontiguous (--)}\end{turn} &
\begin{turn}{90}\parbox{7em}{\scriptsize unbounded (+) \newline  bounded (--)}\end{turn} &
\begin{turn}{90}\parbox{7em}{\scriptsize purely reduplicative (+)  \newline  fixed melody parts (--)}\end{turn}  \\ \hline
Madurese plural  & + & + & + & + & + & + \\
Mokilese progressive & -- & + & + & + & -- & + \\
Nisgha, prefixing & -- & -- & + & -- & -- & -- \\
Koasati, infixing & -- & (--) & -- & + & -- & -- \\ \hline
\end{tabular}
\end{Ex}
The %
first example from Madurese (Malayo-Polynesian) shows a case of
unbounded total reduplication \ref{prob:madurese}:  
a rather familiar type that needs no further comment here.
\begin{Ex}
\item \label{prob:madurese}
\begin{tabular}[t]{ll}
{\em Madurese plural} & \cite[35]{stevens:68}\\ \hline
sakola\textglotstop an-{\bf sakola\textglotstop an} & `schools' \\
panladhin-{\bf panladhin} & `servants' \\
panokor-{\bf panokor}(-ra) & `(his) razors'
\end{tabular}
\end{Ex}
Mokilese (Micronesian) illustrates the next case \ref{prob:mokilese},
namely prefixed reduplicants that consists of a
partial copy of the base. Since the number of segments varies as a function
of the base, simple templatic generalizations seem not to be
available here.
\begin{Ex}
\item \label{prob:mokilese}
\begin{tabular}[t]{lll}
\multicolumn{3}{l}{{\em Mokilese progressive} \cite{blevins:96}}\\ \hline
p\textopeno dok & {\bf p\textopeno d}p\textopeno dok & `plant/ing' \\
nikid & {\bf nik}nikid & `save/ing' \\
wia [wija] & {\bf wii}wia & `do/ing' \\
s\textopeno\textopeno r\textopeno k & {\bf s\textopeno\textopeno}s\textopeno\textopeno r\textopeno k &
`tear/ing' \\
onop & {\bf onn}onop & `prepare/ing' \\
andip & {\bf and}andip & `spit/ing' \\
uruur & {\bf urr}uruur & `laugh/ing'
\end{tabular}
\end{Ex}
Nisgha (Salish), shown in \ref{prob:nisgha}, differs from Mokilese in
that there may be phonological 
modifications in the reduplicant that do not affect their
correspondence partners in the base.%
\begin{Ex}
\item \label{prob:nisgha}
\begin{tabular}[t]{llll}
\multicolumn{4}{l}{\em  Nisgha  CVC prefixing reduplication}  \cite{shaw:87}\\ \hline
a. & m\'a\textlengthmark k$^w$s-k$^w$ & {\bf m$_1$is$_2$}-m$_1$\'a\textlengthmark k$^w$s$_2$-k$^w$ & `be white' \\
b. & l\'ilk$^w$ & {\bf l$_1$ux$^w _2$}-l$_1$\'ilk$^w _2$ & `to lace (shoes)' \\
c. & q\'o\textlengthmark\textglotstop os & {\bf q$_1$as$_2$}-q$_1$\'o\textlengthmark\textglotstop os$_2$ & `to be cooked'
\end{tabular}
\end{Ex}
In \ref{prob:nisgha}.b we can see that spirantization has
turned the velar stop /k$^w$/ into a velar fricative /x$^w$/. Furthermore, the vowel quality of the fixed-size CVC
reduplicant is not copied from the base, but constitutes a fixed melody part instead.%
\footnote{Actually, it may be said to be semi-fixed, since apparently
  reduplicant vowel quality is determined by the flanking consonants. The claim
  to a fixed-melody construction remains valid, however, because
  the reduplicant vowel is not identical to the corresponding base vowel.}
As it is only the first and last segment of the base that is copied, the
reduplicant is %
discontiguous as well.

Finally, Koasati (Muskogean) has a infixing reduplicative construction
depicted in \ref{prob:koasati}, where the
base-initial segment is copied to the interior of the base and
followed by a fixed-melody element /o(\textlengthmark)/. 
\begin{Ex}
\item \label{prob:koasati}
\begin{tabular}[t]{llll}
\multicolumn{4}{l}{{\em Koasati infixing aspectual reduplication }
 \cite{kimball:88}} \\ \hline
& \underline{base} & \underline{punctual} \\
a. & tah\'aspin & t$_1$ahas-{\bf t$_1$\'o\textlengthmark}-pin & `to be light in
weight' \\
b. & lap\'atkin & l$_1$apat-{\bf  l$_1$\'o\textlengthmark}-kin & `to be narrow \\
c. & akl\'atlin & a$_1$k-{\bf h$_1$o}-l\'atlin & `to be loose' \\
d. & okc\'akkon & o$_1$k-{\bf h$_1$o}-c\'akkon & `to be green or blue' \\ 
\end{tabular}
\end{Ex}
The facts are 
further complicated by the need to distinguish between consonantal
left edges and vocalic ones, where in the latter case apparently /h/
-- the voiceless equivalent of a vowel --  serves as the
\finallongpage modified copy.

Two further remarks on properties of reduplication seem appropriate
at this point. First, there are cases like reduplication in Chumash \ref{prob:chumash} which show
that copying must be phonological -- it is not in general sufficient to just
repeat a morpheme!
\begin{Ex}
\item \label{prob:chumash}
\begin{tabular}[t]{lll@{}}
{\em Chumash (Hokan)} & \cite{applegate:76} \\\hline
{\bf s}-RED-ikuk & {\bf sik}-sikuk & 3sg-cont.-chop,hack\\
s-i\v{s}-RED-expe\v{c} & s-i{\bf
  \v{s}ex}-\v{s}expe\v{c} & 3pl-dual-cont.-sing
\end{tabular}
\end{Ex}
Note that the copy includes not only some initial portion of the verb
stem but also the last consonant of the immediately preceding affix,
irrespective of its precise morphological function.

Second, there are reduplicants whose shape cannot be
statically determined as a function of the base, but which crucially require
knowledge of the reduplicant's eventual {\em surface} position. 
A case in point is again represented by the Koasati construction from
\ref{prob:koasati}, where the reduplicant has 
a long vowel  when forced to occupy  the stressed (penultimate syllable)
position in the word, but exhibits a short-vowelled allomorph when  landing
elsewhere. In fact, because of stress shifts in comparison with base-only words
\ref{prob:koasati}.a,b, base stress is a poor predictor of reduplicant position.
The lesson of Koasati is that independent lexical precomputation of
the shape of morphemes entering into a reduplicative construction cannot account for all cases.

Let us now step back from the individual cases and ask some obvious
questions: What do these examples really tell us? How can the classification that captures their
essential dimensions of variation help derive necessary and sufficient properties
of a linguistically adequate one-level account of reduplication?

First of all, reduplications of the partial, modified,
infixing, discontiguous and fixed melody type show that the model depicted
in figure \ref{prob:easy}.a  below is insufficient.
\begin{Ex}
\item \label{prob:easy} {\sc Some problematic reduplication models}\\
\epsfig{file=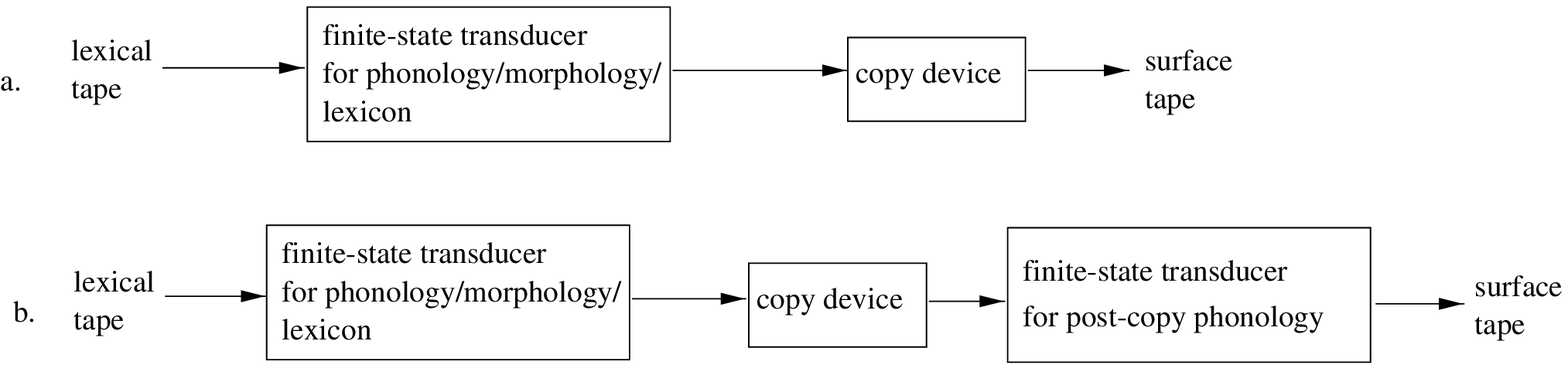,width=.9\linewidth}
\end{Ex}
The depicted attempt to
construct a hybrid model consisting of a strictly finite-state part
driving a separate copy device fails on instances of imperfect copying if the copy device is
limited to the simplest and most efficiently realizable task of
mapping input $w$ to output $ww$. Of course one might devise more
sophisticated variants of such a device (see appendix \ref{app:icopy} for discussion
of a concrete proposal). However,  hybrid models of this kind still suffer from
two rather obvious drawbacks.  First they constitute heterogenous
combinations of a finite-state device with a non-finite state component 
with unclear overall properties. Second, they are more or less
generation-oriented, not inherently reversable system. 

The modified type of reduplication shows that sometimes it is not
enough to copy object-level segments from within a word's surface
realization, but that a more abstract 
notion of identity needs to be captured. While it makes no sense to
place the copy device {\em before} the lexical transducer, in a multi-level approach one 
might envision to place the copy device in the sandwiched position of
a 2-transducer cascade \ref{prob:easy}.b. One could then copy at the level
where object identity still holds and carry out the necessary
modifications in the post-copy part. Unfortunately the aforementioned
problems of hybrid models carry over to this setting as well.
With only one level, OLPM will of course have to employ rather different means
to model modified reduplications. 

Exfixing types of reduplication allow one to leave base linearizations 
untouched; no computational means for infixation need to be provided.
As a further consequence, such constructions in principle can support
extensive arc sharing through the coexistence of simple and
reduplicated realizations in the same finite-state network. In contrast, the
infixing type with its disruption of base linearizations  {\em a priori}
permits no such memory efficiency and needs formal means to support
infixation in the first place.

Discontiguity in general poses the problem of how to model the absence 
of some stretch of segmental material in specific reduplicative constructions when the 
same material is required to be present in all other realizations of
bases. Moreover, the greater the length of the skipped-over substring
of the base in discontiguous reduplicants, the more acute is the
problem of handling even bounded-length variants \ref{prob:nisgha} of
such long-distance dependencies with finite-state automata
(\citeNP[91]{sproat:92}; \citeNP{beesley:98}).   

The bounded types of reduplications are special in that, given the
assumptions of full-form lexica and sufficient storage space, they
could, in principle, be precompiled into a finite-state network.
In practice, however, this can yield probibitively large networks for 
realistic fragments of natural language morphologies \cite[161]{sproat:92}. Also, it has been  
remarked that finite-state storage is very inefficient when it comes to simulating
even the fixed amount of global memory needed to remember the
copied portion of a base \cite{kornai:96}. In contrast, the unbounded
type cannot be modelled by precompilation at all. No
artificially imposed upper bound will do justice to the
facts in a total-reduplication language like Bambara (Northwestern Mande)
where facts such as  \ling{wulu-o-wulu} `whichever dog',
\ling{wulunyinina-o-wulunyinina} `whichever dog searcher',
\ling{wulunyininafil\`{e}la-o-wulunyininafil\`{e}la} `whoever watches
dog searchers', etc.\/ can be arbitrarily extended, as \citeN{culy:85}'s careful
investigation shows. 
Positing separate models for the two types would seem to be highly
problematic, since bounded and unbounded reduplications may occur
in the same language. For example, Madurese,
whose total reduplication appeared in \ref{prob:madurese}, also 
has bounded final-syllable reduplication:
\ling{tr\textepsilon-\textepsilon str\textepsilon} `wives', 
\ling{bu-s\textepsilon mbu} `something increased', \ling{wa-buwa} `fruits'.

\subsection{Discontiguity}
It is a well-known fact that morphology is not always
concatenative. Rather, various patterns of morpheme ``overlap'' can be
observed, as shown in \ref{prob:overlap}.
\begin{Ex}
\item {\sc Patterns of Morpheme Overlap}\\[1ex]
\label{prob:overlap}
\begin{tabular}{@{}l|c@{}c|l@{}}
 Morphemes  & [overlap] &  [inclusion] & Language \\ \hline\hline
\setlength{\unitlength}{3947sp}%
\begingroup\makeatletter\ifx\SetFigFont\undefined%
\gdef\SetFigFont#1#2#3#4#5{%
  \reset@font\fontsize{#1}{#2pt}%
  \fontfamily{#3}\fontseries{#4}\fontshape{#5}%
  \selectfont}%
\fi\endgroup%
\begin{picture}(2454,282)(34,467)
\thinlines
\put( 46,524){\line( 1, 0){1125}}
\put(1171,569){\line( 0,-1){ 90}}
\put( 46,569){\line( 0,-1){ 90}}
\put(1351,524){\line( 1, 0){1125}}
\put(2476,569){\line( 0,-1){ 90}}
\put(1351,569){\line( 0,-1){ 90}}
\put(136,614){\makebox(0,0)[lb]{\smash{\SetFigFont{12}{14.4}{\rmdefault}{\mddefault}{\updefault}t   e   i   l}}}
\put(1441,614){\makebox(0,0)[lb]{\smash{\SetFigFont{12}{14.4}{\rmdefault}{\mddefault}{\updefault}b    a    r}}}
\end{picture}
& \bf -- & \bf -- & German \\ 
\dotfill & \dotfill & \dotfill & \dotfill \\
\setlength{\unitlength}{3947sp}%
\begingroup\makeatletter\ifx\SetFigFont\undefined%
\gdef\SetFigFont#1#2#3#4#5{%
  \reset@font\fontsize{#1}{#2pt}%
  \fontfamily{#3}\fontseries{#4}\fontshape{#5}%
  \selectfont}%
\fi\endgroup%
\begin{picture}(1734,552)(34,197)
\thinlines
\put( 46,254){\line( 1, 0){855}}
\put(901,299){\line( 0,-1){ 90}}
\put( 46,299){\line( 0,-1){ 90}}
\put(600,569){\line( 1, 0){1125}}
\put(1756,614){\line( 0,-1){ 90}}
\put(600,614){\line( 0,-1){ 90}}
\put(136,344){\makebox(0,0)[lb]{\smash{\SetFigFont{12}{14.4}{\rmdefault}{\mddefault}{\updefault}h  i}}}
\put(766,344){\makebox(0,0)[lb]{\smash{\SetFigFont{12}{14.4}{\rmdefault}{\mddefault}{\updefault}t}}}
\put(600,614){\makebox(0,0)[lb]{\smash{\SetFigFont{12}{14.4}{\rmdefault}{\mddefault}{\updefault}S}}}
\put(991,614){\makebox(0,0)[lb]{\smash{\SetFigFont{12}{14.4}{\rmdefault}{\mddefault}{\updefault}a       n       a}}}
\end{picture}
  & + & \bf -- & Mod. Hebrew \\ 
\dotfill & \dotfill & \dotfill & \dotfill \\
\setlength{\unitlength}{3947sp}%
\begingroup\makeatletter\ifx\SetFigFont\undefined%
\gdef\SetFigFont#1#2#3#4#5{%
  \reset@font\fontsize{#1}{#2pt}%
  \fontfamily{#3}\fontseries{#4}\fontshape{#5}%
  \selectfont}%
\fi\endgroup%
\begin{picture}(1419,507)(34,242)
\thinlines
\put(316,614){\line( 1, 0){405}}
\put(721,659){\line( 0,-1){ 90}}
\put(316,659){\line( 0,-1){ 90}}
\put( 46,299){\line( 1, 0){1395}}
\put(1441,344){\line( 0,-1){ 90}}
\put( 46,344){\line( 0,-1){ 90}}
\put(136,389){\makebox(0,0)[lb]{\smash{\SetFigFont{12}{14.4}{\rmdefault}{\mddefault}{\updefault}b}}}
\put(766,389){\makebox(0,0)[lb]{\smash{\SetFigFont{12}{14.4}{\rmdefault}{\mddefault}{\updefault} a    s   a}}}
\put(361,659){\makebox(0,0)[lb]{\smash{\SetFigFont{12}{14.4}{\rmdefault}{\mddefault}{\updefault}u   m}}}
\end{picture}
  & + & + & Tagalog 
\end{tabular}
\end{Ex}
In these diagrams, we depict the surface
extent of a morpheme as the temporal interval between its first and
last segment.%
\footnote{This assumes an understanding of morphemes (and
words) as totally ordered sets of segments. Suprasegmental morphemes
like `nasalize word till first  obstruent' (cf.\/ the  Arawakan
language Terena, \citeNP{bendor-samuel:60}) need
  a generalization that refers to observable phonological effects instead of
  segments. As far as I know, cases of improper inclusion or total
  overlap always  correspond to such suprasegmentals, too, and can be exemplified 
  by phenomena like nasalization, pharyngealization, tone marking etc.}
 Hence, overlap between two morpheme intervals implies
that at least one of the ordering relationships
between intramorphemic segments must be  weakened from immediate to 
{\em transitive} precedence: the hallmark of a discontiguous morpheme.

While German \ling{teil-bar} `divisible' is a typical
instance of purely concatenative arrangement of morphemes, the next
example from Modern Hebrew illustrates a first case of deviation from
the concatenative ideal, namely {\em morpheme-edge
metathesis}. Here the conjugation class prefix /hit-/ ({\em hitpael} binyan), which ends in
coronal /t/,  partially overlaps verbal stems whose first segment is
a coronal obstruent. Note that  {\em hit-} 
and coronal-initial verbal roots are only discontiguous if cooccurring
(!\ling{\ip{hiStana}} `changed (intr.)', *\ling{\ip{hit-Sana}}, but
\ling{\ip{me-Sane}} `change (pres.)' and \ling{hit-gala} `appear', *\ling{higtala}, etc.). 

In Tagalog {\em 
  infixation}, the deviation is more severe, as the actor-trigger
morpheme \ling{-um-} is totally overlapped by the 
verb stem in \ling{b-um-asa} `read'. In the dual case of {\em circumfixation}, also attested in
Tagalog (e.g.\/ \ling{ka-an}, as in \ling{ka-bukir-an} `fields'), affix and
stem  have simply changed roles in what amounts to the same pattern of total overlap.
It is worth pointing out that Tagalog still has its share of purely
concatenative morphology, even involving the same stems (e.g.\/
\ling{mak\'a-basa} `read to somebody (by chance)'). 

Even if a language allows discontiguity in regular morphology, it may
nevertheless exhibit exceptional morphemes that 
forbid intrusion into their own material. For example, in Ulwa construct-state
morphology \ref{intro:examples}.b there are noun stems  
like \ling{kililih} `cicada' where infixation *\ling{kili-ka-lih}
would be predicted, but only suffixation \ling{kililih-ka} is
possible. It must therefore be possible to control infixability in the 
Ulwa lexicon.

With so much focus on discontiguity, we should tackle a potential objection.
Is productive discontiguous morphology perhaps limited to so-called
`exotic' languages? English, thought
otherwise to be purely concatenative, allows us to give a negative
answer to that question. The language has a productive process of
expletive insertion which readily creates words like
\ling{Kalama-goddam-zoo, in-fuckin-stantiate, kanga-bloody-ro,
  in-fuckin-possible, guaran-friggin-tee} \cite[45]{katamba:93},
thereby breaking up stems that appear elsewhere as contiguous. 

The challenge of discontiguity then is to come up with a
generic formal solution that is ideally able to represent morphemes in a
uniform manner, regardless of whether discontiguity is prominent, rare
or nonexisting in a language. It should also capture the fact that
immediate precedence of intramorphemic segments appears to be the default, giving way to
transitive precedence only if immediate adjacency leads to
ungrammaticality. Finally, the solution should allow for cases where 
discontiguity is either grammatically or lexically forbidden. 
\subsection{Partiality}
Sometimes morphemes do not realize all their segmental material. We
have already seen that in Tonkawa and Modern
Hebrew, certain stem vowels are omitted by way of regular processes, depending on the
affixation pattern. However, in these and 
many similar cases the number of potentially zero-alternating segments
is strictly predictable. In Tonkawa, at most every stem vowel (and /h/, 
phonetically a voiceless vowel) can loose one mora 
(V, /h/ $\rightarrow$ \lingnull, VV $\rightarrow$ V), whereas in Modern
Hebrew there is a maximum of two alternating stem vowels per verb
form. Furthermore, since in these languages omittable vowels are
intercalated with stable consonants, the length of contiguous stretches of
deletable material is also bounded by a small constant, often 1.
Besides this type of {\em bounded partial realization}, however, there is a type of potentially
{\em unbounded partial realization}, a case of which we have seen in
productive German 
i-truncation \ref{intro:examples}.\/a. Here, the length of the deleted 
string suffix of a base noun is a linear function of its original length and therefore
 in principle unbounded (2, 3 and  5 segments in
 [\ip{pe:t}\sout{\ip{ra}}i], [\ip{Pand}\sout{\ip{Ke:a}}i], [\ip{gOKb}\sout{\ip{atSOf}}i]). While
it is true that Standard High German does not use truncation for ordinary %
grammatical processes such as pluralization,%
\footnote{However, some German
dialects do have limited truncation for plurals, e.g. Upper Hessian
\ling{\ip{hOnd}} `dog (sg.)' $\rightarrow$ \ling{\ip{hOn}} `(pl.)'
\cite{golston.wiese:96}.}
other languages like
Tohona O'odham, Alabama, Choctaw and Koasati 
employ truncation for exactly this purpose \cite[65f]{anderson:92}.

Summing up, some natural desiderata for a generic formalization of partiality
would be to allow for morpheme representations where a priori {\em no}
segmental position {\em must} be realized, to provide for flexible
control over actual  realization patterns, and to account easily for the
frequent %
case where no part of a morpheme is omitted.
\subsection{Floating Morphemes and Directionality}
The inherent assumption of the continuation-classes approach to
morphotactics \cite{koskenniemi:83} is that morphemes are always tied to a fixed position in the
usual chain of affixes and stems that make up a word. However, clear counter-cases of so-called
 {\em floating}  morphemes do exist, for example, in Huave (Huavean) and
Afar (East Cushitic). In Afar \ref{prob:afar}, the same affix may flip between prefixal and suffixal
position, depending on the phonology  of the stem \cite{noyer:93}.
\begin{Ex}
\item \label{prob:afar} {\sc Second Person Floating Affix in Afar}\\
\begin{tabular}[t]{llllll}
a. & {\bf t}-okm-\`e & b. & yab-{\bf t}-\`a & c. & ab-{\bf t}-\`e \\
    & 2-eat-perf & & speak-2-impf & & do-2-perf \\
    & `you ate' & & `you speak' & & `you did'
\end{tabular}
\end{Ex}
Descriptively, the second-person affix ``{\em t-} occurs as a prefix before a nonlow
[stem-initial, M.W.] vowel and elsewhere as a suffix''\cite{noyer:93}. 

The placement of variable-position infixes has been analyzed in OT
by attributing to them an inherent, directional `drift' towards the left or right
edge of some constituent such as the word. Analyses using this idea
typically employ additional affix-independent constraints expressing
e.g.\/ prosodic wellformedness to control the ultimate deviance 
from the desired position. OT's development
of constraint-based directionality was originally illustrated
with the Tagalog \ling{-um-} infixation case that we
briefly mentioned above. It led to the introduction of the EDGEMOST
constraint \cite{prince.smolensky:93} and later to a family of
generalized alignment constraints \cite[et seq]{mccarthy.prince:94}
that seek to minimize the distance to some designated edge. 

Adapting the gist of the idea for our present purposes, one could analyze Tagalog \ling{-um-}
by assuming leftward drift for the infix while simultaneously
imposing a specific prosodic wellformedness condition on surface
words: syllables must have onsets.  
One advantage of this analysis is that it explains why lexically
onsetless affixes cannot become prefixes: *\ling{.um.-ba.sa.}, 
!\ling{b-u.m-a.sa}.%
\footnote{This analysis deviates from \citeANP{prince.smolensky:93}'s original 
  treatment, which admitted onsetless \ling{um-} before roots like
  \ling{aral} `teach'. Their analysis is rejected in
  \citeN[198]{boersma:98}. He points out that orthographically
  vowel-initial  roots are actually pronounced
with a leading glottal stop (\ling{\ip{.P-u.m-a.ral.}}), while forms like
  \ling{\ip{.mag.-Pa.ral.}} `study', with a proper prefix, show that this glottal
  stop is better assumed to be part of lexical representation. To prevent intrusion
    of \ling{-um-} into complex-onset roots like \ling{gradwet}, we
    may either assume that infixal /m/ is prespecified to become a syllable onset
    or make sure that immediately adjacency in complex onset
    members is inviolable lexical information:
    *\ling{.g-um.-rad.wet.}, !\ling{.gr-u.m-ad.wet.} `graduate'. 
}

We may proceed similarly for the Afar case.
This time the leftward drift of \ling{(-)t-} needs
to be coupled with an affix-specific conditional constraint that demands a nonlow vowel to
the right if landing in word-initial position. In addition, 
some formal way of ruling out discontiguous stems is needed to prevent 
infixed \ling{-t-}. Note that if the second person affix cannot become
a surface prefix, it indeed lands 
on the {\em first} position after the stem, thereby
underscoring the leftward-drifting behaviour attributed to {\em t-}.

For motivation of rightward drift, let us look at some data from
Nakanai \ref{prob:nakanai}, an Austronesian language cited in
\citeN[213f]{hoeksema.janda:88}. 
\begin{Ex}
\item \label{prob:nakanai} {\sc Nakanai suffixing VC reduplication}\\
\begin{tabular}[t]{lll}
haro & har-{\bf ar}-o & `days' \\
velo & vel-{\bf el}-o & `bubbling forth' \\
baharu & bahar-{\bf ar}-u & `windows' \\
abi & ab-{\bf ab}-i & `getting' \\
kaiamo & kaiam-{\bf am}-o & `residents of K. village'
\end{tabular}
\end{Ex}
Besides technical devices to model total substring reduplication $XY
\, \rightarrow\, X_iY_j-X_iY_j$ and a \ling{VC}
requirement for the reduplicant's segmental content, a minimalist analysis 
only needs to add rightward-drifting behaviour for the reduplicant.%
\footnote{Interestingly, the symmetrical case of leftward-drifting VC
  reduplication with otherwise identical conditions can be found in
  the Salish language Lushootseed \cite[214]{hoeksema.janda:88}: 
  \ip{st\'ubS} $>$ \ip{st-{\bf \'ub}-ubS} `man', \ip{P\'ibac} $>$
  \ip{P-{\bf \'ib}-ibac} `grandchildren'.}
The reader will find it easy to verify that the forms in the second
column of \ref{prob:nakanai} indeed optimally satisfy both the drift
specification and the VC requirement, whereas any drift further to the right would
violate the latter requirement: *\ling{haro-{\bf  ro}} shows faithful copying and perfect suffixation,
but has a CV reduplicant.%
\footnote{One objection to the analysis just sketched might be that
  one could (perhaps generally) eliminate drift at the expense of prosodic
  subcategorization. In Nakanai -- at least for the data
  in \ref{prob:nakanai} -- this would involve no more than the reduplicant's requirement
  for adjacency to a word-final syllable nucleus. We will discuss the nature of the
  tradeoff involved in choosing between conventional versus drift-based analyses
  in more detail in \sref{ex}.}

A seemingly different kind of directionality is involved in choosing
grammatical vowel/\lingnull{} realization patterns in the bounded-partiality
morphologies of Modern Hebrew, Tonkawa and others. To account for the
kind of left-to-right preference found in these %
patterns, \citeN{walther:97} has proposed a so-called Incremental Optimization
Principle, ``Omit zero-alternating segments 
as early as possible''. \label{firstIOPdiscussion} The principle explains,
inter alia, why in Tonkawa \ling{\ip{we-pcen-oP}} is grammatical (cf. \ref{back:tonkawa})
but *\ling{\ip{we-picn-oP}},*\ling{\ip{we-picen-oP}} are not.  The
latter two represent a missed chance to leave  
out /i/, which  appears earlier in the speech stream than the second stem
vowel /e/. Prosodic wellformedness alone does not distinguish between these
forms since none of them violates the CV(C) syllable constraints. 

To sum up, floating morphemes and the different kinds of
directionality posit yet another challenge for
formalization. The question here is how to express drift in lexical
representations and whether it is possible to unify the seemingly
diverse kinds of directionality within a single formal mechanism.

Having provided linguistic motivation for certain abstract
requirements for formalization of prosodic morphology, we now turn to
our central proposals that meet these requirements in the context of a finite-state
model.
\section{\label{ext}Extending Finite-State Methods to \\ One-Level Prosodic Morphology}
In this section I will present the extensions to OLP that are deemed necessary to
formalize the range of prosodic morphology phenomena  described
above. A basic knowledge of formal languages, regular
expressions and automata will be helpful in what follows, perhaps at
the level of introductory textbooks on the subject (\citeNP{hopcroft:ullman:79},
\citeNP{partee.meulen.wall:90}). 
\subsection{Technical Preliminaries}
\citeN[\S 3]{bird.ellison:94} proposed state-labelled automata as
the formal basis for autosegmental phonology. In the present
work, however, I will return to more conventional arc-labelled
automata. The main reason for this choice is that it eases actual computer
implementation, since existing FSA toolkits all rest on the arc-label
assumption (\citeNP{vannoord:97}, \citeNP{mohri.pereira.riley:98}).
As \citeANP{bird.ellison:94} themselves note, the choice has no
theoretical consequences because the two automata
types are equivalent (Moore-Mealy machine equivalence).

With respect to the actual content of the arcs, however, I will follow
OLP by allowing {\bf sets as arc labels}. \citeN{eisner:97}, who
employs the same idea for an implementation of OT-type phonological
constraints, argues that this helps keep constraints small.
The reason why we may gain a more compact encoding of automata for
phonological and morphological purposes is because boolean combinations
of finitely-valued features can be stored as a set on just one arc, rather than being
multiplied out as a disjunctive collection of arcs. Again it must be
emphasized that the choice is not crucial from a theoretical point of
view, but simply convenient for actual grammar development. 

Of course, sets-as-arc-labels require modifications to the
implementation of various standard operations on finite-state
automata. Our representation for these sets is in the form of bit
vectors. For FSA {\em intersection} $A \cap B$ this implies that the identity
test ($A.arc_{i}=B.arc_{j}$) between two arc labels  must be replaced with a refined
notion of arc compatibility ($A.arc_i\cap B.arc_j \neq \emptyset$),
which is efficiently implementable with bitwise logical AND and
test-for-nonzero instructions. While FSA concatenation, union and reversal 
have operationalizations that are independent of the nature of arc symbols, forming the
{\em complement} of a FSA involves {\em determinization} and {\em completion}, two
operations which again require modification. Recall that a complete
automaton is one that has a transition from every state for each
element of the automaton alphabet. 

Completion can be implemented 
for set-labelled automata by creating a new nonfinal state $sink$
and adding for every state $s\neq sink$ an extra arc pointing at
$sink$. This arc is labelled with the universal alphabet set
$\Sigma$ minus the union of all sets that label the outgoing arcs of
state $s$. Again, bitwise operations 
for complement and logical OR can be used here. Whereas for completion a direct realization 
seems preferrable, one way to realize FSA determinization instead involves a
reduction to the conventional, identity-based version. It capitalizes 
on the insight that non-empty label intersection gives the same results
as label identity test iff all label sets bear the property of being either pairwise identical or 
disjoint. By replacing existing arcs with disjunctive arcs in such a
way  that this property is met
in the entire automaton, a new automaton can be created that is then
subject to conventional determinization, i.e.\/ with bit vectors reinterpreted as simple
integers. Hence, this scheme allows the reuse of existing, optimized
software. An optional post-determinization step could then fold multiple
disjunctive arcs connecting any two states back into a single arc bearing a disjunctive label.
Of course, nothing precludes more efficient schemes which 
would presumably require somewhat more extensive modification of classical
determinization algorithms to make the implementation efficient for
bit vector arc labels. 

Finally, FSA mimization is an important operation when it comes to
the compact presentation of analysis results and also when memory
efficiency is crucial.  However, because reversal and
determinization suffice to implement minimization
\cite{brzozowski:62}, no further modifications are 
necessary to support set-based arc labels.  
\subsection{Enriched Representations}\label{enrichments}
This section describes in detail some generic enrichments to conventional
finite-state representation in the OLPM model, motivating in each case why inclusion of
the proposed representational element is warranted for one-level analyses of prosodic
morphology.

In what follows we will abbreviate
set-based arc labels with mnemonic symbols for reasons of
readibility. Also, we will liberally apply set-theoretic notions
to symbols, speaking e.g.\/ of disjoint symbols when we really mean that 
the sets denoted by those symbols must be disjoint. 

As it is actually done in the implementation, we will furthermore conceive of
those symbols as types that are organized into a type hierarchy, allowing the grammar writer to
express both multiple type inheritance and type disjointness. However,
the syntactic details will be suppressed in the text; rather, we will
describe the essential parts of the type signature in prose for the sake of clarity. The semantics
of the type system assumed here is extremely simple: the denotation of a parent type in the 
directed acyclic graph that constitutes a type inheritance hierarchy is defined as the union of the
denotation of its children, whereas a type node without children denotes
a unique singleton set (cf.\/ \citeNP{ait-kaci.et.al:89}). Complex
type formulae are permitted, using the Boolean connectives \texttt{\&
  ;}$\bf\sim$ for logical AND (intersection), OR (union) and NOT (complement), respectively. As mentioned before, all
type formulae are ultimately represented by bit vectors. 

Let us proceed to the first enrichment, which is a preparatory step for {\bf reduplication}. 
It is a definining characteristic
of reduplication that it repeats some part of a string and it would be
nice if this property could somehow be encoded explicitly. Intuitively, in one sense
repetition is just about moving backwards in time during the process of
spelling out string symbols. (We momentarily disregard the second
aspect of repetition, that of ensuring proper identity of
symbols). Therefore, in an initial attempt to flesh out this intuition
one could add designated `backjump' arcs to an  automaton. Backjump arcs $S \stackrel{\text{\tiny backjump}}{\longrightarrow} P_i$ would
directly connect every state $S$ to all of its predecessor states $P_i$.
A predecessor state $P_i$ is defined as follows: $P_i$  lies on a path $p$ leading from a start
state to $S$ and there exists a non-empty proper subpath of $p$
beginning at $P_i$. The example in \ref{ext:backjumps} shows the automaton for a Bambara word \ling{malo} `rice'.
\begin{Ex}
\item \label{ext:backjumps} {\sc Backjumps: an initial attempt at repetition}\\
\epsfig{file=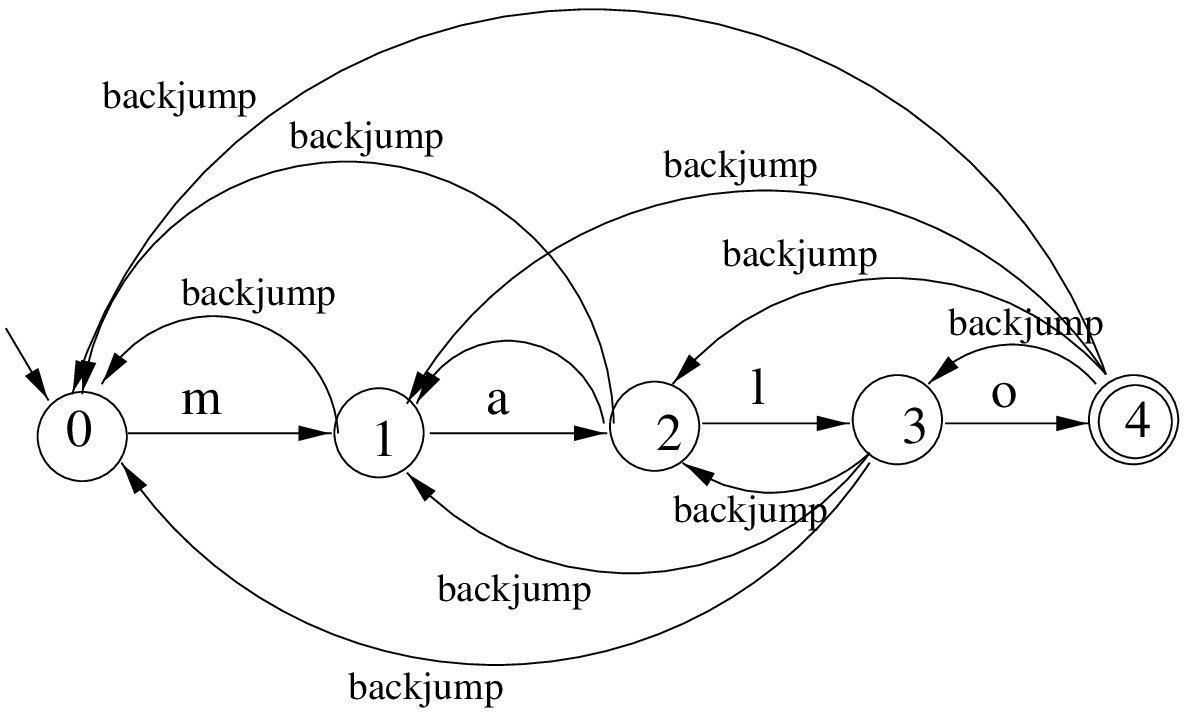,width=.9\linewidth}
\end{Ex}
However, this potential solution has several shortcomings. 
It is not linear in the length of the string, adding $n(n+1)/2$ arcs
for a string of length $n$. Given a
single backjump symbol -- which is desirable for reasons of uniform
`navigation' within a network --, it introduces too much nondeterminism,
because at string position $i, 0 < i \leq n$ there is a choice between
 $i$ possible backjumps. This nondeterminism in turn makes it cumbersome to 
specify a fixed-length backjump, a specification which would not be  unusual in
linguistic applications. Finally, we would require an unbounded memory of already
produced states if lazy incremental generation of string arcs was
desired for an efficient implementation.

As it turns out, however, we really need only a proper subset of the backjumps
anyway, because with the kleene-star operator we
already have sufficient means in our finite-state calculus to express
iterated concatenation, i.e.\/ nonlocal arc
traversal. A better solution therefore breaks down 
nonlocal backjumping into a chain of local single-symbol backjumps.
It consists of adding a reverse {\bf repeat arc} $j \rightarrow i$
labelled with a new technical symbol {\tt repeat} for every pair of states ${<}i,j{>}$
connected by at least one content arc $i \rightarrow j$. Content arcs are
defined as arcs labelled with segmental and other properly 
linguistic information. Furthermore, content symbols must be disjoint
from technical symbols. 
\ref{ext:repeat-arcs} has an example showing the Bambara word for
\ling{rice} under the new encoding. 
\begin{Ex}
\item \label{ext:repeat-arcs} {\sc Repeat arcs: the final version}\\
\epsfig{file=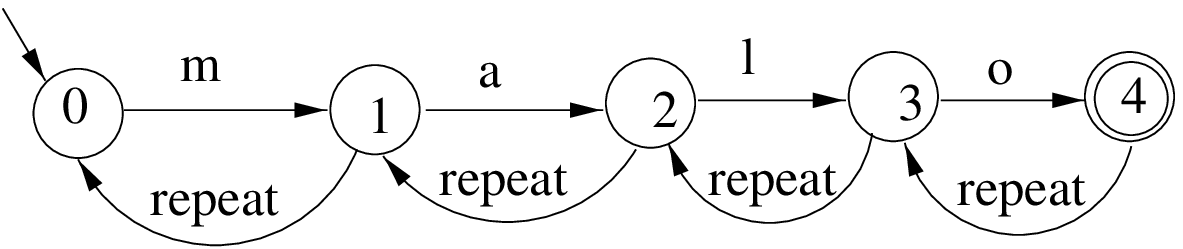,width=.9\linewidth}
\end{Ex}
To give a preview of how this can be used in reduplication: the
regular expression 
\[ segment^*\,repeat\, repeat\, repeat\, repeat\, segment^* \]
describes an
automaton which -- when intersected with \ref{ext:repeat-arcs} -- will 
yield a string that contains two instances of \ling{malo} separated by 
four repeat symbols. The type $segment$ used in the previous expression
abbreviates the union of all defined segmental content symbols.

Summing up then, the repeat-arc solution is better than the
previous one, because the additional amount 
of repeat arcs is linear in the length of the string, because there is no 
nondeterminacy in jumping backwards, because it is now trivial
to add a local repeat arc upon lazy generation of a content arc, and
because it has become easy to jump backwards a fixed amount of $k$
segmental positions by way of the regular expression $repeat^k$.

The reader will have noted, however, that repeat arcs
change the language recognized by the  
respective automaton, in particular by rendering it infinite through
the introduction of a cycle within each arc-connected state pair
${<}i,j{>}$. Also, by referring to {\em  states}, repeat-arc introduction
was defined as manipulation of a concrete automaton representation
rather than with respect to the language denoted by the  
automaton.%
\footnote{This way of constructing automata is sometimes discredited,
   with preference given to exclusively high-level algebraic
  characterizations of the underlying languages and relations
  \cite[376]{kaplan.kay:94}. However, in line with 
  \citeN{vannord.gerdemann:99} who argue convincingly for a more flexible
   overall approach, there are good reasons to deviate from
   Kaplan\&Kay's advice in our case. Concretely, while it is possible to define the appropriate
automaton  for all strings of length 1 through the regular expression
$ContentSymbol\, ( ContentSymbol\, repeat)^*$, concatenation  
of  $n$ such expressions for a string $s, |s| = n$ does {\em not} define the
same language as the repeat-enriched automaton corresponding to
$s$. I conjecture that only the automaton-based approach
can be compositional with respect to concatenation.}
Obviously, the minimization properties of automata
containing repeat arcs are rather different from those of repeat-free 
automata as well. I will take up these and other issues related to
repeat arcs later, when we have seen how the proposed enrichment 
interacts with intersection and synchronization marks to implement reduplication.

The next enrichment covers {\bf discontiguity}, with its characteristic
property of transitive precedence between content symbols. Again, to
be maximally generic we need to allow
for intervening material at every string position, i.e.\/ before and after
every content symbol. Also, intervening material may itself be
modified by other morphological processes, i.e.\/ it may contain
technical symbols such as repeat. We therefore add a {\bf self loop} to every
state $i$, i.e.\/ an arc $i \rightarrow i$ labelled with $\Sigma$, the union of
content and technical symbols. The example
this time displays the Tagalog word \ling{basa} \ref{ext:self-loops}.
\begin{Ex}
\item \label{ext:self-loops} {\sc Self loops: a representation for discontiguity}\\
\epsfig{file=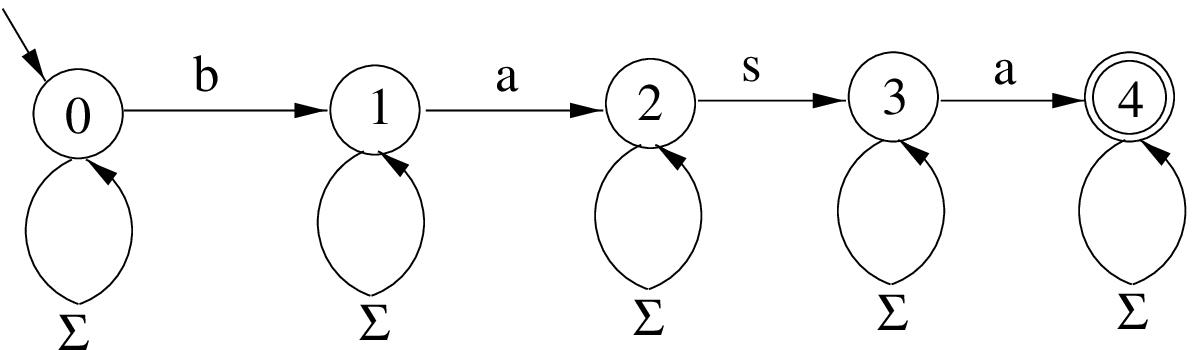,width=.9\linewidth}
\end{Ex}
To preview a later use of this representation: when figuring as
part of  \ling{b-um-asa}, the self loop pertaining to state
1 will absorb the infixal material \ling{um}.
The further issues and consequences that arise from this enrichment
step are mostly the same as for the previous one, apart from the
additional question of what to do with unused self loops like those
emerging from states 0,2,3,4 in the case of singly-infixed \ling{b-um-asa}. Again, it seems
best to treat those issues later on. 

The third enrichment deals with {\bf partiality} and is especially useful
for truncation. Here we want to be able to
exercise fine control over the amount of material that gets realized
or skipped over, and also have the option of leaving out an {\em a
  priori} unbounded amount of segmental content. Therefore, the
leading idea again is to use a local encoding, which consists of adding companion {\bf skip arcs} $S
\stackrel{\text{\tiny skip}}{\longrightarrow} T$ to all content arcs  $S
\stackrel{\text{\tiny ContentSymbol}}{\longrightarrow} T$.
 Like its repeat counterpart, $skip$ is defined to be a new
 technical symbol.  Example \ref{ext:skip-arcs} illustrates how the automaton representation of 
German \ling{Petra} `proper (first) name' looks like after the
enrichment.
\begin{Ex}
\item \label{ext:skip-arcs} {\sc Skip Arcs: a representation for unbounded truncation}\\
\epsfig{file=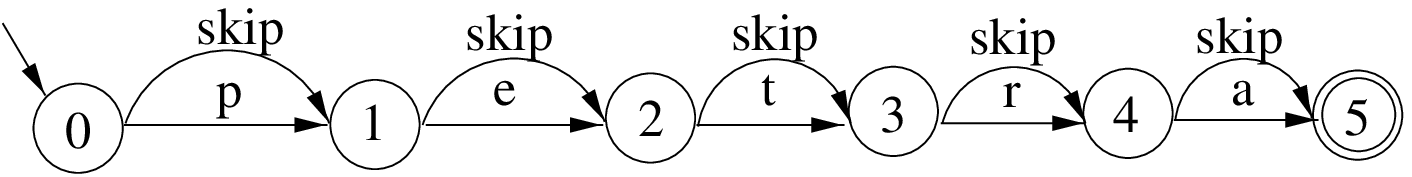,width=.99\linewidth}
\end{Ex}
To give a simple idea of how skip arcs might be used, consider the
following regular expression for picking out the base portion of the
hypocoristic form \ling{Pet-i}: \[  seg^*\, skip\, skip \] When
intersecting this expression with \ref{ext:skip-arcs},  
the string $p\,e\,t\,skip\,skip$ results. A more principled analysis
of German hyporistic formation follows in \ref{ex:i-formation}. 

Noting that in this solution technical and content arcs systematically share both source and
target states, we can actually merge those two arc types and avoid the additional complexity
introduced by separate skip arcs. Under a set-based labelling regime
this works as follows: every content arc will now be labelled with the
{\em union} of the old content symbol and the skip symbol (type formula:
$ContentSymbol ; skip$). Thus separate skip arcs would only be strictly
necessary when for some reason set labels are not available. 
\subsection{Resource Consciousness} \label{resource}
So far we have ignored a specific problem associated with self loops,
the mechanism proposed in \ref{ext:self-loops} to handle free
infixation. The problem is that the same self loops that were designed to absorb infixal content
may also absorb accompanying contextual constraints in
unexpected ways. It is best to illustrate this unwanted interaction by 
way of an actual example that is already familiar, Tagalog {\em -um-} infixation. Since the
infix lands just behind an obligatory word-initial stretch of syllable onsets, one 
particularly simple way of encoding this prosodic requirement is to attach it to the left
side of the infix itself. In doing so we assume that segments are tagged
with syllable role information by means of finite-state
syllabification (not shown here).%
\begin{Ex}
\item \label{ext:um} {\sc Automata for Tagalog} \ling{-um-} {\sc infixation}\\
\epsfig{file=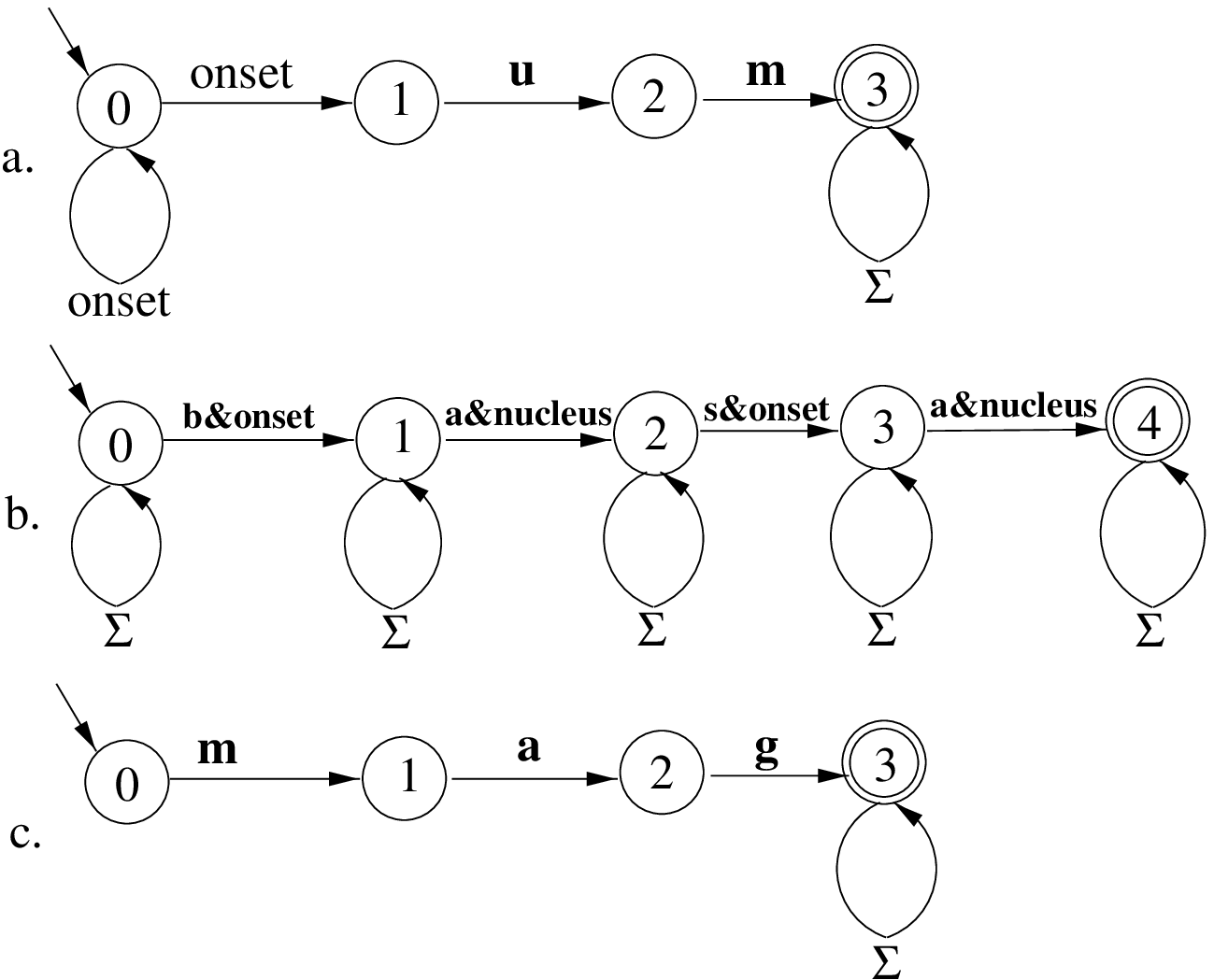,width=.9\linewidth}
\end{Ex}%
However, upon intersection of \ref{ext:um}.a%
\footnote{Note that 
  \ling{-um-} is specified as a contiguous morpheme  by leaving out
  morpheme-medial self loops. This reflects the fact that the infix itself never gets broken up.}
 with the self-loop-enriched 
representation of a stem like \ling{basa} \ref{ext:um}.b the resulting
complex automaton overgenerates. It contains at least one ungrammatical placement
of the affix-plus-left-context in absolute prefixal position:
$ onset^+ u\, m\, b\, a\, s\, a$ versus the correctly infixing $b\&onset\, u\, m\,  a\,
s\, a$. This is because the affix proper {\em   and} its prosodic
context are both consistent with \ling{basa}'s initial self loop $0
\stackrel{\Sigma}{\rightarrow} 0$, causing a kind of vacuous
self-fulfilment of the contextual constraint  which now hallucinates
onsets that were never provided by the stem itself. 

Note that
it is not possible to remove the initial self loop: ordinary
prefixation using e.g.\/ \ling{mag-} \ref{ext:um}.c. can apply to the same
stems that take \ling{-um-}. Therefore, unless we are willing to give up
the overall idea of morphemes-as-constraints that are uniformly
combined via intersection, we should use
intersection in this case as well, with the initial self loop now
playing host to the prefixal material.  The argument becomes
particularly compelling in the case of conditioned allomorphy. \citeN{ellison:93} has argued
that because such cases of contextually restricted morphemes
exist even in agglutinative languages like Turkish and because intersection
can  enforce restrictions and simulate concatenation but not 
vice versa, it should become the preferred method of morpheme combination in
a constraint-based setting. 

What then is the cause of our problems in \ref{ext:um}?
My diagnosis is that we lack an essential distinction between {\bf   producers
and consumers of information}. Contextual constraints should only be
satisfiable or `consumable' when proper lexical material has been
 provided or `produced' by {\em independent} grammatical resources. Now this
 notion itself is already familiar from other areas of computational
 linguistics, going back at least to LFG's distinction between
 constraining and constraint equations \cite{bresnan.kaplan:82}. Since then
 the general idea has gained some popularity  under the heading of
 resource-conscious logics.%
\footnote{See e.g.\/ the `glue logic' approach to the
 syntax-semantics interface in LFG \cite{dalrymple:99}; asymmetric
 agreement under coordination \cite{bayer.johnson:95};
 \citeN{johnson:97}'s development of a more thoroughly resource-based
 R-LFG version; \citeN{dahl.tarau.li:97}'s Assumption Grammar
 formalism; \citeN{abrusci.fouquere.vauzeilles:99} logical formalization of TAGs.} 

Here I propose to introduce resource consciousness into 
automata as well. Suppose we tag symbols with a separate {\em producer/consumer bit}
(P/C bit) to formally distinguish the two kinds of information. P/C =
1 defines a producer, P/C = 0 its consumer counterpart. By convention,
let us mark producers by bold print in both regular expressions and
automata; see \ref{ext:um} for illustration. Suppose 
furthermore that we distinguish two modes of interpretation within our
formal system. During {\bf open interpretation} the intersection of
two compatible arcs produces a result arc whose P/C bit is the logical OR
of its argument bits, whereas in {\bf closed interpretation} mode the result 
arc receives a P/C bit that is the logical AND of its argument
bits. Drawing the analogy to LFG, open interpretation mode is somewhat
akin to unification-based feature constraint combination during LFG parsing,
while closed interpretation mode would correspond to checking the
satisfiability of constraining equations against minimal models at the end of the parse.
In open interpretation, producers are dominant in intersective
combination, so that the only consumer arcs surviving after a chain of
constraint intersections are those that never combined with at least
one producer arc. This is similar to the behaviour found in intuitionistic resource-sensitive
logics, where a resource can be multiply consumed, but must have been produced 
at least once.

Open versus closed interpretation imposes a natural two-phase evaluation
structure on grammatical computation. After ordinary intersective
constraint combination using open interpretation mode obtains in  phase
I, phase II intersects the resulting automaton  with 
$\boldsymbol{\Sigma}^*$ -- the universal producer language -- in closed
interpretation mode. Because as a result only producer arcs survive,
the second step effectively prunes away all unsatisfied
constraints, as desired.%
\footnote{Because pruning in closed interpretation mode is based on
  examination of an arc's content, specifically its P/C bit, our proposal
  differs from \citeN{bird.ellison:92}'s purely structural $prune(A)$ 
  operation, which indiscriminately removes all self loops from a state-labelled automaton $A$.}
 Observe that all the real work happens in phase I, which is
fully declarative. Phase II on the other hand is automatic and not under control of the
grammar.%
\footnote{This is to be understood as a conceptual statement. For practical
  experimentation we have devised a macro
  \code{closed}\code{\_interpretation} whose use {\em is} visible in formal
  grammars. It follows that open interpretation mode is the default
  setting in our implementation.}
 To indicate that declarativity has only slightly
been sacrificed, we will call the resulting grammatical framework that obeys two-phase
evaluation and the open/closed interpretation distinction {\bf
quasi-declarative},  and speak of Quasi-Declarative Phonology etc.

It is now easy to see that our introductory problem of getting
\ling{b-um-asa} right has been solved: in the illformed alternative
corresponding to word-initial position the consumer-only onset
arcs that constitute \mbox{\ling{-um-}'s} contextual constraint meet a
consumer-only $0 \stackrel{\Sigma}{\rightarrow} 0$ self
loop. Since consumers intersecting with each other in phase I remain
just what they are, they are immediately eliminated when intersecting
with the universal {\em producer} language in phase II of grammatical
evaluation. In later sections we will encounter further examples 
that underscore the utility of a resource-conscious style of grammatical analysis
and demonstrate its wide applicability in prosodic morphology. %

\subsection{Copying as Intersection}
In this section I will explain a generic method to describe
reduplicative copying using finite-state operations. The germ of the
idea already appeared in \citeN[48]{bird.ellison:92}, where the authors 
noted that the product of automata, i.e.\/ FSA intersection, 
is itself a non-regular operation with at least
indexed-grammar power. In illustrating their claim they
drew attention to the fact that odd-length strings of indefinite
length like the one described by the regular expression $(a\, b\, c\,
d\, e\, f\, g)^+$ can be repeated by intersecting them with an
automaton accepting only strings of even length, yielding  $(a\, b\, c\,
d\, e\, f\, g\, a\, b\, c\, d\, e\, f\, g)^+$ in the example at hand.

Since we already know now that with intersection we have a promising
operation in our hands to implement reduplication, let us work out the 
details. First, we will show how to get total reduplication in a way
that makes use of neither the odd-length assumption of
\citeANP{bird.ellison:92}'s toy example nor of {\em a priori} knowledge about
the length of the string as in our preview of a possible application
of repeat arcs \ref{ext:repeat-arcs}.
An initial step that improves on that previous account for reduplicating the repeat-encoded
\ling{malo} string dispenses with the length knowledge by using the
following regular expression:
 \[ m\, seg^*\, o\,\, repeat^*\, m\, seg^* \] 

Although we have now replaced 4 ($=|malo|$) consecutive repeats by
$repeat^*$, an expression which encodes jumping back an indefinite
amount of time, it is clear that something else is required to  
generalize beyond this example. In comparison to words like \ling{wulu}
`dog', \ling{malo} is special in that $m$ and $o$ are distinct
symbols that occur only once in the string. Thus
they are able to serve two functions at the same time, acting as ordinary content
symbols {\em and} as markers of left and right edges of the
reduplicant. 

That observation points to the crucial issue at hand, which is how to
identify the edges of a reduplicant in a generic 
fashion. In \ling{malo} it just happened %
that the edges were already self-identifying. For the general case we may borrow
freely from \citeN[68f]{bird.ellison:94}'s solution to an analogous
problem arising in autosegmental phonology, that of modelling
the synchronizing behaviour of association lines. Just like in their
solution, let us assume a distinct {\bf synchronization bit}. Here it will
be added to all content symbols, being set to 1 for the edges we want to
identify and receiving the value 0 elsewhere.%
\footnote{\citeN{ellison:93} contains an earlier application of
  synchronization symbols to the problem of translating concatenation
  into intersection. Ellison's comment that only a finite alphabet
  (\{0,1\}) is needed in the translation carries over into the present
  setting: synchronisation symbols need not be multiplied for
  triplication, quadruplication, etc.} 
 Adopting
\citeANP{bird.ellison:94}'s notation for combining content and
synchronization information, we can draw the automaton for
\ling{wulu} as follows.%
\footnote{In our typed setting, we actually use a new type $synced$,
 writing $ContentSymbol \& synced$ for $ContentSymbol{:}1$ and
 $ContentSymbol\&{\sim}synced$ for $ContentSymbol{:}0$. 
 Type declarations ensure that each instance of $ContentSymbol$ is
 itself underspecified with respect to  synchronization.}   
\begin{Ex}
\item \label{ext:repeat-sync} {\sc Repeat Arcs with Synchronization Bits}\\
\epsfig{file=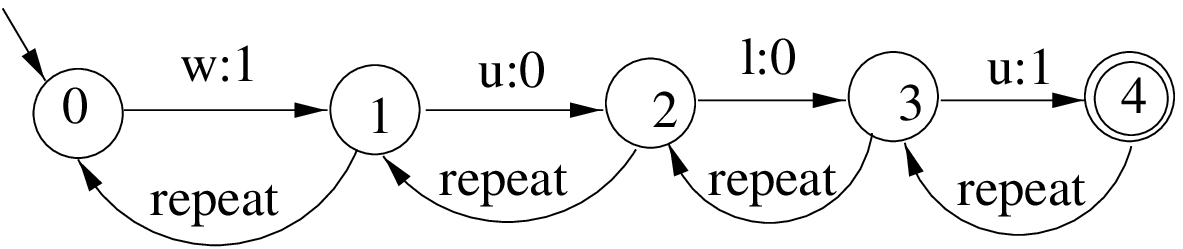,width=.9\linewidth}
\end{Ex}
Now the regular expression that describes total reduplication can be
liberated from mentioning any particular segmental content: 
\[ seg{:}1\, seg{:}0^*\, seg{:}1\, repeat^*\, seg{:}1\, seg{:}0^*\, seg{:}1 \] 
Figuratively speaking we start at a synchronized symbol
representing the left edge, move right through a possibly empty series
of unsynchronized segments to another synchronized symbol representing
the right edge, then go back through a series of repeat arcs until
we encounter a synchronized symbol again, which must be the left
edge. Note how  the same subexpression is used twice
to identify `original' and `copied' occurrence of the reduplicative
constituent. With more instances of $seg{:}1\, seg{:}0^*\, seg{:}1$,
triplication, quadruplication etc. would all be feasible using the
same approach. Also, it is interesting to see how {\em one} bit  
actually suffices in this scheme to  identify {\em two} kinds of edges
in all strings of length $>$ 1, exploiting the fact that
concatenation  is associative but not commutative.

To handle actual Bambara's \ling{Noun-o-Noun} reduplication, we need to
combine the enrichments of section \ref{enrichments} and
\ref{resource}, in particular using one of the self loops to provide
space for the intervening /o/. Here then is the full encoding of both
\ling{wulu} `dog'  and the reduplicative construction itself: 
\begin{Ex}
\item \label{ext:wulu} {\ling{wulu} \sc and reduplication automaton}\\
\epsfig{file=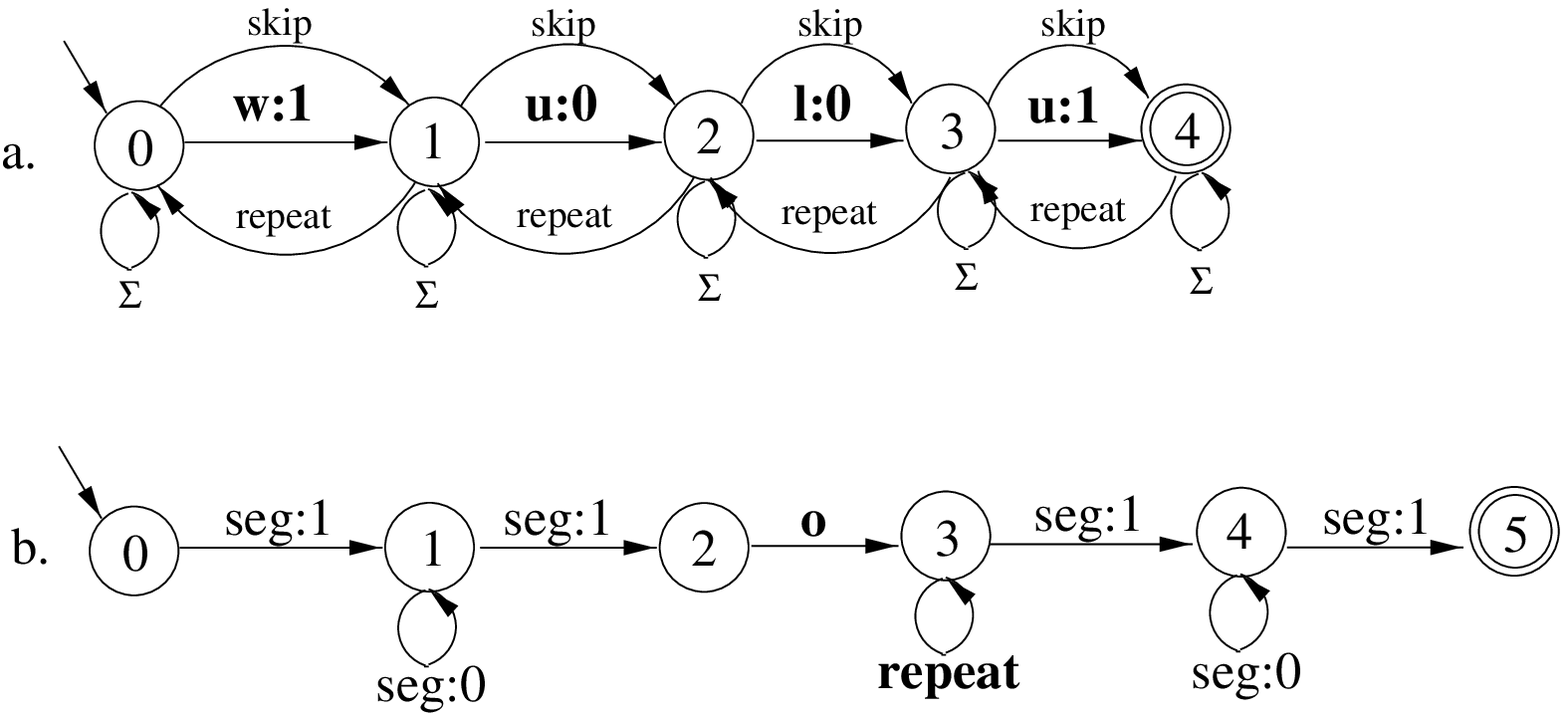,width=.99\linewidth}
\end{Ex}
After intersection \ref{ext:wulu}.a $\cap$ \ref{ext:wulu}.b and
pruning away of consumer-only arcs we get an automaton which is
equivalent to \[  w{:}1\, u{:}0\, l{:}0\, u{:}1\, o\, repeat^4\, repeat^*\,
w{:}1\, u{:}0\, l{:}0\, u{:}1 \]

There was nothing special in intersecting with a singleton lexical
set, hence it is trivially possible to extend the Bambara lexicon beyond the
single example we have shown. %
All that is needed is to
represent new lexemes as FSAs in the same vein as \ling{wulu}, and %
define the lexicon 
as the union of those FSA representations.

Also, it is easy to see that we can construct a variety of other reduplication
automata that use different sychronization and repeat patterns or
employ additional skip arcs. These would encode the various types of
partial reduplications that we saw in section
\ref{prob:reduplication}. I will indeed present %
one such case in section \ref{ex}, but %
leave the others as an exercise to the reader for reasons of space.
\subsection{Bounded Local Optimization}%
This section defines a final
operation over enriched automata called Bounded Local Optimization
(BLO), which can be understood as a restricted, non-global search for
least-cost arcs in weighted automata. The operation will be shown to permit implementation of 
the Incremental Optimization Principle (IOP, p.\/\pageref{firstIOPdiscussion}), while later sections
illustrate that it can also be used for morpheme drift and longest-match behaviour.

To prepare the ground for such an operation it is best to return to %
the simple
Tonkawa example \ling{we-pcen-o\textglotstop} (cf. \ref{back:tonkawa}) that was discussed 
previously in connection with the IOP. Given a stem representation $p(i)c(e)n(a)$
that contains three zero-alternating vowels, the addition of two
nonalternating affixes \ling{we-} and \ling{-o\textglotstop} does not
enlarge the resulting set of eight ($2^3$) word  
forms. Intersection of this set with some simplified prosodic constraints 
*CCC and *VV that ban sequences of at least three consecutive
consonants and two adjacent vowels still leaves us with three remaining
forms. Here the IOP steps in, preferring \ling{we-pcen-o\textglotstop} 
over *\ling{we-picen-o\textglotstop}, *\ling{we-picn-o\textglotstop}
because only the first form lacks the \ling{i} that
constitutes the earliest omittable vowel.

To implement this kind of behaviour, the first idea is to extend the
FSA model once again, namely towards the inclusion of {\bf
local weights} on arcs. The weights are taken from a totally ordered
set and represent the costliness of a particular choice. In our
example, the realization of an alternating vowel  
should be more costly than its omission, e.g.\/ by receiving a greater
weight. Now this move in itself fits in with the recent gain in popularity that
weighted automata and transducers have enjoyed, finding application in areas such as speech
recognition, speech synthesis, optimality theory and others
(\citeNP{pereira.riley:96}, \citeNP{sproat:96}, \citeNP{ellison:94}). 
Usually, however, the theoretical assumption has been that the minimal weighted
unit is the string, not the individual symbol. Taking advantage of this
assumption, \citeN{mohri:97} is able to both move and modify individual weights
between arcs in an operation called `pushing' in order to prepare a
weighted automaton for minimization.  In our application, though,
weights represent localized linguistic information which should not be 
altered. Therefore, I will at present pursue the alternative of
weighting the symbols themselves. Also, for 
purposes of this paper it will actually suffice to introduce a small finite number
of different weights.%
\footnote{Interestingly, \citeN{kiraz:99} reports that actual
  grammars for Bell Labs text-to-speech applications also obey this restriction, e.g.\/ the German 
  module has 33 weights and the French module 12.}
 Hence the weights used here can equally
well be formalized as part of the type hierarchy that structures label 
sets, and that is indeed what the current implementation does.
While it would be definitely be worth trying to take the other route
as well and make the transition to the general case of unrestricted, possibly numeric,
weighting schemes that build upon  the weighted-string assumption, 
the choice seems premature right now. Rather, it seems best to wait
until the analysis of enough decisive 
phenomena has been carried out in the present framework and then evaluate what
the ultimate consequences of each assumption are.

Here then is one automaton representation of the \{%
\ling{we-pcen-o\textglotstop}, \ling{we-picen-o\textglotstop},
\ling{we-picn-o\textglotstop}\} set, enriched with only two weights representing the {\em marked}
 realization case (depicted as /1) and the {\em
  unmarked} elsewhere case (depicted as /0).  
The ordering that will be assumed is $marked > unmarked$.
\begin{Ex}
\item \label{ext:weights} {\sc Weighted automaton for Tonkawa example} \\
\epsfig{file=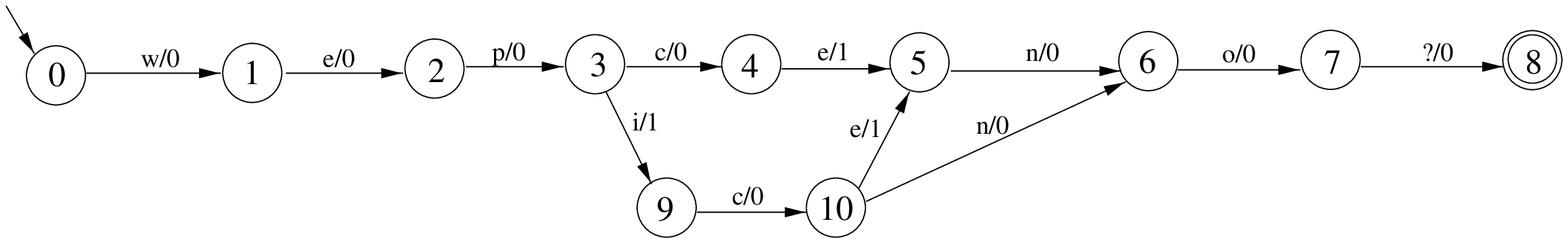,width=.99\linewidth}
\end{Ex}
Note that in particular affix vowels are correctly represented as unmarked,
because they are nonalternating, just like all members of the consonantal 
`root' $p-c-n$.

Now, how does one use the weights to implement the IOP? The crucial
observation is that IOP's pruning of costlier alternatives translates
into local inspection of the arcs emanating from
states like  3 and 10 in
\ref{ext:weights}. In each of the choices $ 3 \rightarrow 4\, or\,
9,\, 10 \rightarrow 5\, or\, 6$, there is an {\em alternative} between a marked
arc and an unmarked one. Let us call such an arc $A \in
State.arcs$  a {\em choice arc} whenever  $|State.arcs| > 1$.
By cutting away marked, i.e.\/ non-optimal choice arcs%
, we arrive at an automaton which
recognizes the single string \ling{wepceno\textglotstop}, as
desired. Since it is desirable to abstract away from the specifics of this example,
we will in the following develop a new operation called Bounded Local
Optimization (BLO) which will encapsulate the
locally-determined pruning of non-optimal
arcs. To make it widely applicable, we introduce two generalizations
into our example procedure. 

The first generalization is to prune only those arcs
from the set $S.arcs$ whose weight is
greater than the {\em minimum} weight over this entire set. For
example, $3.arcs = \{3 \stackrel{c/0}{\rightarrow} 4, 3
\stackrel{i/1}{\rightarrow} 9\}$ has the associated minimum weight $0$. As a consequence,
preservation of non-choice arcs like $4 \stackrel{e/1}{\rightarrow} 5$
is automatic, since the minimum over singleton weight sets is
independent of the only element's weight value. Also, the generalization means that
even multiple choice arcs will survive pruning iff they are all weighted with 
the minimum cost, thus providing a way to maintain alternatives
beyond the pruning step, e.g.\/ to implement free variation.%
\footnote{\label{neg_weights} One slight modification that is not pursued here (but will
  be assumed in \sref{ex:tagalog_sect} would add a
  mechanism to make designated arcs inert to minimum-based
  pruning. The rationale behind this move is that there are scenarios
  where e.g. technical arcs would be compared to content arc alternatives of
  various weight in a nonsensical way. A simple way to signal
  inertness is by negative weights:  an arc is pruned whenever the
  length-$k$ alternatives yield a {\em positive} lower summed
  weight. With the help of a special value $-\infty$ we can even prevent
  arc pruning {\em independent} of $k$ and the actual weight
  distribution in the alternatives.}

The second
generalization is to parametrize the operation under development with
a fixed look-ahead of $k$ arcs, summing up weights over each individual $k$-length path
extending from a given state. In our running example, $k=1$ was sufficient
to detect gradient wellformedness differences in a maximally local
fashion. In general, though, one might need to examine a greater %
number of consecutive arcs to discover the (non-)optimality of an
alternative path. For example, if the arc labels of $3 \rightarrow 9,
9 \rightarrow 10$ switched places and arc $10 \rightarrow 6$ was
eliminated, we would need $k=3$ to see that $3 \rightarrow 9$
initiates a costlier path, hence should be optimized away ($\sum{ 3
\rightarrow 9 \rightarrow 10 \rightarrow 5} = 2,\sum{ 3
\rightarrow 4 \rightarrow 5 \rightarrow 6} = 1$).

To sum up,
calling BLO a kind of optimization is justified because non-minimally-%
weighted choice arcs are pruned; the optimization is boundedly local because no
path of length greater $k$ influences the decision of which arcs to
prune. In contrast, the (N-best) shortest-path algorithms
(\citeNP{dijkstra:59}, \citeNP{tarjan:83}) %
used in most  other applications of weighted automata constitute global optimization procedures
which will use information from the entire automaton to determine
their result. Although the latter optimization procedures could presumably
 be used as part of the aforementioned general exploration of the
weighted-string alternative, pursuing the more restricted 
local variant is preferred here, because it incorporates an interesting
hypothesis about what formal power is actually needed in processing
prosodic morphology.%
\footnote{Interestingly, \citeN[96ff]{mohri.riley.sproat:96} also explore
  (different) incomplete optimization methods that visit only a subset of the
  states. However, their motivation is to solve a
  practical problem in speech recognition, namely that the enormous
  number of states prohibits plain application of  
  single-source shortest-path algorithms on current hardware.}
 Another consequence of adopting BLO is that it
makes the question of what maximal look-ahead is (perhaps universally)
required a topic of promising empiricial research, which could shed
further light onto the resource-conscious structure of natural language patterns.

Let us now proceed from informal sketches to a more precise definition 
of BLO. First, I will give a different characterization of BLO as a 
function that maps between weighted formal languages, in order to ensure its
representation-independent definability. Such a characterization is
desirable, since for any fixed $k$ and suitable locally-weighted language
$L$ one can construct `illbehaved' automaton representations, e.g.\/ by
inserting epsilon transitions%
, so that non-optimal paths cannot be detected within a window of
length  $k$. Therefore, in a second step I will briefly consider the
automaton-theoretic implementation of the BLO again, 
focussing in particular on the question of which automaton
representation acts as wellbehaved input to it. 

In the language-theoretic characterization of BLO, then, the  idea is to sum
over (the local weights of) $k$-length substrings,  the equivalent of
$k$-length arc paths. We discard a string $w$ when there is at least one 
position whose associated $k$-length sum exceeds the
minimum sum for that position obtained through evaluation of other 
comparable strings, i.e.\/ those which share a common prefix. For the
case where only a substring of 
length $j < k$ exists, we simply define that its 
weight sum is the sum of the existing weights up to position $j$, or
equivalently that non-existing positions have implicit weight 0. Here then is the formal 
version of BLO:
\begin{examples}
\item \label{ext:blodef} Given an alphabet $\Sigma$ as a finite, nonempty set
  of symbols and a set of positive, real-valued weights $\mathbb{R}_+$,
  a {\em locally weighted language} $L$ is defined as follows: $L \subseteq
  (\Sigma\times\mathbb{R}_+)^*$.  We will also speak of a {\em locally
    weighted string} $w$ whenever $w
  \in L$ for some locally weighted language $L$. 
  Finally, let $w[i]$ pick out the 
  $i$-th pair in $w$ for $0 \leq i < |w|$, let $w[0\dots i]$ denote
  the length-$i$ prefix of $w$ and let $\pi_2$ be the projection of the
  second element of a pair.\\[1ex]

\noindent Then {\em Bounded Local Optimization} $BLO: L \times \mathbb{N}\setminus \{0\} \mapsto
L$ is defined as %
\[ \begin{split} 
BLO(L, k) =  \{ & 
w \in L | \nexists pos \in \mathbb{N},\, 0 < pos + k
\leq |w|,\, \nexists v \in L{:}\, %
\\ & v[0\dots pos] = w[0\dots pos], \\ & 
weight\_sum(pos, k, v) < weight\_sum(pos, k, w)
\},\end{split} \] 

\noindent and, with $k,pos \in \mathbb{N}$ and $w$ a locally weighted 
string, %
\[ weight\_sum(pos,k,w)\, =\, 
     \begin{cases}
0, & k = 1 \wedge pos \geq |w| \\
\pi_2(w[pos]), & k = 1 \wedge pos < |w| \\
\begin{split} & weight\_sum(pos, 1, w) + \\ & weight\_sum(pos,k-1,w),\end{split} & k > 1
\end{cases} \]
\end{examples}
To illustrate the operation just defined, suppose 
\begin{align*} & L=\{w_1,w_2\}, \text{with } w_1 = {<}a,0{>} {<}b,1{>}, w_2 =
{<}c,1{>}{<}d,0{>}. \text{ Then } BLO(L,1) = \{w_1\}, \text{because} \\
& w_1[0\dots 0] = w_2[0\dots 0] = \epsilon \text{ and } \\
& weight\_sum(0,1,w_2) = \pi_2({<}c,1{>}) = 1 > 0 =
\pi_2({<}a,0{>}) = weight\_sum(0,1,w_1),\\ &\text{together with the
  fact that examination of position 1 leaves the optimality of } w_1
 \text{ unchallenged,} \\ &\text{since no common prefix exists: } 
w_1[0\dots 1] = {<}a,0{>} \neq w_2[0\dots 1] = {<}c,1{>}.
\end{align*}
Thus, the example illustrates that BLO is in fact a `greedy', 
directional type of optimization: weights are evaluated by position
such that a string cannot compensate for an initial costly string
portion by some cheaper suffix  if an alternative with cheaper prefix exists. This is 
a welcome result since one application of BLO
is to model the IOP, whose original formulation ``Omit zero-alternating segments 
{\em as early as possible}'' was intentionally defined in directional terms.
Although BLO's directionality is perhaps not immediately apparent from
the declarative definition, it does in fact follow from the local
examination of weights embodied in $weight\_sum$ together with the
common prefix requirement.   
Note also that
$BLO(L,k){=}L$ for $k>1$, because by
$weight\_sum(0,k,w_{1/2}) {=} 1$ both strings are kept. Hence, we see that optimization results
are not necessarily monotonic with respect to parameter $k$. In particular,
care must be taken to avoid look-ahead windows that are too big,
because -- as we have just seen -- overstretching the bounds of locality
can sometimes blur distinctions expressed by the weights.  

There is a last peculiarity worth noting, which has to do with the entity over
which BLO, or any other grammar-defining optimization approach, for
that matter, should be applied. Suppose that in our illustrative example $L$
the strings $w_1$ and $w_2$ actually represented different lexical items rather
than realizational alternatives of a single item. The result $BLO(L,1) 
{=} w_1$ then means that with $w_2$ unfortunately a lexical item itself has been pruned, or in
other words, that optimization cannot be safely applied over the entire
lexicon. Rather, BLO must be applied on a per-item basis, at least
conceptually. There are various ways to actually implement this
requirement. One way would be to prefix each lexical item with
string-encoded semantic and morphological
information that is weighted with the same item-independent weight. 
The prefixes make item beginnings unique, so they will be preserved
even in a minimized FSA version of the lexicon, and the uniform
weights ensure that no item will be prematurely discarded, thus
banning harmful interaction. However, one drawback of such a scheme
would be that FSA minimization would have less chances of reducing the
size of automata, as compared to the usual encoding of grammatical
information at the {\em end} of phonological strings \cite{karttunen.kaplan.zaenen:92}. The latter
encoding could be preserved in a second scheme where during generation one first intersected
the lexicon automaton with $\Sigma^*{<}${\em semantic/morphological
  features of desired word form}${>}$ and then applied BLO to the
result. Finally, one might devise an algorithm to predict which arcs potentially
participate in harmful interaction and prefix only these with
disambiguating information, in what might be seen as an attempt to use the
first method on a demand-driven basis for improved minimization behaviour.

Returning to the question of what machine representation $\alpha$ of a locally
weighted language $L$ is sound input
to an automaton-based implementation of BLO, I conjecture that a sufficient
condition for soundness is that $\alpha$ takes the form of the {\em minimal deterministic
automaton for $L$ having a single start state}. To see the plausibility 
of this conjecture recall that
while a minimal automaton is defined to have the minimal number of
states, it is also minimal in number of transitions \cite[Corollary
1]{mohri:97}. Being minimal and deterministic, 
string prefixes are shared wherever possible and there are neither
epsilon transitions nor is there useless nondeterminism. Thus the
remaining choice arcs must encode non-reducible local choices. While the BLO
algorithm considers each of the choice arcs emanating from a given
state for pruning, it suffices to examine the summed weights of
length-$k$ paths starting with those arcs because the condition of
common prefixes in definition \ref{ext:blodef} is already guaranteed
through sharing. Because $\alpha$ has a single start state, it follows that
the condition is also met for the case of the empty prefix. 

Given this clarification about the nature of BLO input, we are 
now in a position to present the algorithm in pseudocode for the
automaton-based implementation in \ref{ext:bloalg}. 
\begin{Ex}
\item \label{ext:bloalg} {\sc BLO Algorithm}\\[-1ex]
\begin{tabbing}
10 \= {\bf do} \= {\bf if} \= {\bf if} \kill
BoundedLocalOptimization($\alpha,\, k$) \\
1 \> $\beta .trans \leftarrow visited \leftarrow \emptyset$ \\
2 \> $states \leftarrow \beta .start \leftarrow \alpha .start$ \\
3 \> {\bf while}  $states \neq \emptyset$ {\bf do}\\
4 \>  \> $q \leftarrow$ {\sc Dequeue}$(states)$ \\
5 \> \> $visited \leftarrow visited \cup \{q\}$\\
6 \> \> {\bf if} $|q.arcs| > 0$ \\
7 \> \> {\bf then}  $(nextstates, nextarcs) \leftarrow \text{NextArcsOnMinimalPaths}(q,k)$\\
8 \> \> \> {\sc Enqueue}$(states, nextstates - visited)$\\
9 \> \> \> $\beta . trans \leftarrow \beta .trans \cup nextarcs$\\
10 \> \>  {\bf if} $q \in \alpha . final$ \\
11 \> \>  {\bf then} $\beta .final \leftarrow \beta .final \cup \{q\}$ \\
12 \> {\bf return} $\beta$
\end{tabbing}
\end{Ex}
The algorithm takes a locally weighted automaton $\alpha$ and the
look-ahead constant $k$ as input.
Line 1 initializes the set of transitions of the result automaton
$\beta$ and the set of already $visited$ states to empty, while line 2
copies the input start states to both the result start states and the
set of unprocessed $states$. While there are $states$ to process (line 
3), we remove a current state $q$ from this set and mark it as $visited$
(line 4--5). If that state has outgoing arcs (line 6), we examine
all paths of maximum length $k$ that originate at $q$ and return those 
$nextstates \subseteq \{ dest | q \stackrel{label}{\longrightarrow}
dest\}$ and $nextarcs \subseteq q.arcs$ that were found to lie on one
of the paths with minimal weight sum (line 7).%
\footnote{Note a slight complication that arises when type-based
  disjunctive arc labels are allowed: sometimes even a single arc like $ i
  \overset{b\&1;a\&0}{\longrightarrow} j$ with type-encoded weights
  $0$ and $1$ must not be pruned alltogether but rather have 
  non-minimal disjuncts removed, as in $ i
  \overset{a\&0}{\longrightarrow} j$. The necessary generalization of
  $\text{NextArcsOnMinimalPaths}(q,k)$ is easy: one simply collects the
  {\em set} of (summed) weights $W$ for any given arc (path) using $n$ type subsumption
  checks that test containment of each of the $n$ weight types, and
  then {\em intersects}  the arcs $q.arcs$ with
  $0 \overset{min(W)}{\longrightarrow} 1$ to effect pruning. However,
  for expository purposes we will stick with the conventional
  one-arc-per-disjunct assumption,
 at least in connection with BLO.} 
 We then add the new, i.e.\/ non-visited
`minimal' states found in this way to the set of unprocessed states (line 8). Note that 
this step is responsible for the local, incomplete exploration of the state set
of $\alpha$: `non-minimal' states will not be considered in further iterations
(unless, of course, other arcs happen to refer back to them). In the next step we 
add $nextarcs$ -- the pruned subset  of $q$'s outgoing arcs -- to the
transition set of the result automaton (line 10). Finally, if the
current state was final in the input, then so must it
be in the result  (line 11). After the loop over all
`minimal' states has been completed, we return the finished result
automaton $\beta$ in line 12.

It is easy to see that the algorithm in \ref{ext:bloalg} must always
terminate. The set of $states$ that controls the only existing loop is
initialized to a finite %
set of start states at the
beginning. While line 8 from the loop body 
increases $states$ by some amount which is bounded by $\underset{q\in
  \alpha.states}{max}(|q.arcs|)$, it simultaneously ensures that no state
will be added more than once due to the subtraction of $visited$
states ($visited$ itself grows monotonically, line 5). Because by definition
$|\alpha.states| < \infty$, the total increase must be finite as
well. Since each iteration unconditionally removes one element from $states$, the
nonemptiness condition in line 3 will evaluate to $false$ after a
finite number of iterations, as required for the proof of termination. 

Similar ideas have been explored under the heading of locality in {\em
  violable} constraint evaluation by \citeN{tesar:95} and in particular
\citeANP{trommer:98} (\citeyearNP{trommer:98}, \citeyearNP{trommer:99}). \citeN[p.30,fn.12]{trommer:98} explicitly
acknowledges the intellectual debt to the Incremental Optimization
Principle of \citeN{walther:97}, which is also the precursor of BLO;
both of Trommers papers apply the local evaluation  
concept in an interesting way to Mende tone data. However, while the clearest exposition of
his local optimization algorithm is in \citeN{trommer:99}, there are a 
number of differences and problems. Trommer's {\em Optimize(T)} is
defined as an algorithm over transducers only, there is neither a characterization in terms
of regular relations comparable to our language-theoretic definition
of BLO nor a discussion of the dependency on a suitable normal
form for automata. His algorithm definition is both somewhat erroneous (lines 8,9) and not 
formulated as an incomplete search that directly exploits the
computational advantage of locality. Finally, the generalization to a
look-ahead $k > 1$ is missing, and {\em Optimize(T)} is used at each
step of a constraint cascade, in contrast to the restricted use of BLO 
as a one-step filter on the final result of autonomous automata intersections.
\subsection{Flat representation of prosodic constituency}
So far we have avoided to take any stand on the issue of which set of
prosodic categories to assume, how to  conceive of the relationships
between such categories and how to represent these in a finite-state
framework. For concreteness, we will now briefly consider the
topic in a bit more detail. However, perhaps somewhat suprisingly,
prosodic constituency above the level of sonority will
 not be strictly necessary in any of the three case studies under \sref{ex}. Thus, the reader
may skip this subsection on a first reading, resting assured on the
principal result developed below, namely that the present framework
freely allows for conventional prosodic constituency, albeit in a new
representational format, whenever the empirical facts or different 
styles of analysis seem to warrant its inclusion.

Since at least \citeN{selkirk:80} many authors have assumed that phonology
above the segmental level is organised in a fashion much similar to
syntax, employing hierarchical structure to represent
prosodic constituency. In Selkirk's work, for example, the
categories of syllable $\sigma$, foot $\Sigma$ and prosodic word
$\omega$ are proposed, together with subscripted s(trong)/w(eak)
modifications to mark up subcategories and superscripted primes to tag 
supercategories. Hence, a word like English \ling{sensational}
receives the following prosodic representation \ref{ext:sensational}. %
\begin{Ex}
\item \label{ext:sensational} {\sc English} sensational {\sc according
    to \citeN[601]{selkirk:80}}\\
\attop{\xymatrix@-10pt{
 & &  \omega \ar@{-}[dll] \ar@{-}[dr] \\
\Sigma_w \ar@{-}[dd] & & & {\Sigma_s^{'}} \ar@{-}[dl] \ar@{-}[ddr]\\
& & \Sigma_s \ar@{-}[dl] \ar@{-}[dr]\\
\sigma & \sigma_s & & \sigma_w & \sigma_w \\
\text{\bf sen} & \text{\bf sa} & & \text{\bf tio} & \text{\bf nal}
                        }
         }
\end{Ex}
Though not depicted by Selkirk, one might proceed similarly below the
syllable level with syllabic roles that ultimately connect to segments 
\ref{ext:sensational_2}. %
\begin{Ex}
\item \label{ext:sensational_2} {\sc English} sensational: {\sc
    possible syllabic structure}\\
\attop{\xymatrix@-10pt{
 & \sigma  \ar@{-}[dl] \ar@{-}[d] \ar@{-}[dr] & & &
 \sigma_s  \ar@{-}[dl] \ar@{-}[d] \ar@{-}[dr] & & &
 \sigma_w\ar@{-}[dl] \ar@{-}[d] \ar@{-}[dr] & & 
 \sigma_w  \ar@{-}[dl] \ar@{-}[d] \ar@{-}[dr]  \\
O  \ar@{-}[d] & N \ar@{-}[d] & C \ar@{-}[d] & O \ar@{-}[d] & N \ar@{-}[d] &
C \ar@{-}[d] & O \ar@{-}[d] & N \ar@{-}[d] & CO \ar@{-}[d] & N \ar@{-}[d] &
C \ar@{-}[d] \\
\text{\bf s} & \text{\bf\textepsilon} & \text{\bf n} & \text{\bf s} &
\text{\bf\textepsilon} & \text{\bf j} & \text{\bf \textesh} &
\text{\bf\textschwa} & \text{\bf n} & \text{\bf\textschwa} & \text{\bf 
  l}
                                    }
         }
\end{Ex}
In our case we have used the four roles O, N, C, CO for onset,
nucleus, coda and codaonset, the latter being a representation for
ambisyllabic segments and geminates (cf. \citeNP[ch.3]{walther:97}). Of
course there are many competing proposals as to which categories to
adopt and how to make best use of the dominance relationships. Yet all 
of these proposals share the common assumption of a finite category
set; hence for formal purposes the examples just given suffice to illustrate our main point.

This main point is how to linearize such graph-structured
representations  in conventional finite-state 
models. In particular, a perspicuous lossless encoding of both dominance and immediate
precedence relationships is needed. Moreover, we would ideally want a
kind of distributed representation where the properties denoted by
categories can be locally inspected rather than, say, demanding a
nonlocal reference to some distant boundary symbol in a traditional bracketed notation.

Towards this goal, our leading idea will be to {\bf reinterpret the transitive
dominance relation as monotonic inheritance}. Now
\citeN[44f]{bouma.nerbonne:94} have pointed out that 
one of the restrictions of the inheritance relation, as it is usually
defined, is {\em idempotency}: inheritance cannot distinguish between
structures that differ only in recursion level
(e.g.\/ \ling{anti-anti-missile} $\neq$ \ling{anti-missile}). Fortunately, 
the above diagrams -- and most others in the phonological literature -- contain
no such recursion in the literal sense. Although a super-foot category
$\Sigma^{'}_s$ dominates an ordinary foot $\Sigma_s$ in \ref{ext:sensational}, it is
distinguished by a prime. To proceed with inheritance, we therefore
demand that occurrences of any such pseudo-recursive categories must be pairwise
distinct, which can be achieved by means of e.g. X-bar levels or
primes that form part of the symbol at hand.  

A second restriction imposed specifically by {\em monotonic} inheritance is {\em
  commutativity under associativity}: if $C$ inherits from $B$ and $B$
inherits from $A$, then the result is the same as if $C$ inherits from 
$A$ and $A$ inherits from $B$. This equivalence implies that for
purposes of simulation-by-inheritance the order of dominance must not
matter. One way to ensure this is to fix a particular order in
advance. We therefore demand that the dominance relation of any
constituent structure must be consistent with an {\em a priori} given total
order $>_{dom}$ over the set of categories: $\forall x,y:
category(x)\wedge category(y) \wedge dominates(x,y) \rightarrow x
>_{dom} y$.
 In our examples, we would have $\omega >_{dom} \Sigma^{'}_{s,w} >_{dom} 
\Sigma_{s,w} >_{dom} \sigma_{(s,w)} >_{dom} O,N,C,CO$. Hence, a diagram
where e.g.\/ $\sigma$ dominated $\Sigma_w$ would be ruled out as inconsistent.

With only a finite set of prosodic categories left that enter into
formally non-recursive structures and 
moreover respect the dominance precedence relation $>_{dom}$, the recipe
for flattening a given structure is now quite simple to formulate \ref{ext:flattening}.
\begin{examples}
\item \label{ext:flattening}\begin{examples}
\item To prepare classification of occurrences of category $X$, set up the following
  type hierarchy for each nonterminal $X\in CategorySet \stackrel{def}{=} \{c|
  \exists x: c <_{dom} x \vee x <_{dom} c\}$: 

\noindent\attop{\xymatrix@-10pt{
& & X \ar@{-}[dl] \ar@{-}[d] \ar@{-}[dr] \\
& {[}X \ar@{-}[dl] \ar@{-}[dr] & \_X\_ & X{]} \ar@{-}[dl] \ar@{-}[dr]\\
{[}X\_ & & {[}X{]} & & \_X{]}
                                    }
         }

\noindent The intuition behind this is that category $X$ is best
modelled as a phonological event \cite{bird.klein:90}, i.e. a temporal
interval bearing the property $X$. On the standard assumption that
there are terminal categories whose concatenation
forms the `terminal yield' of a category-as-temporal-interval, we
then will be able to tag each of those terminal elements for their relative position within
the interval (cf.\/ also \citeNP{eisner:97}). In actual phonological practice, terminals will 
frequently be segments, but could also be features etc. Left or right
brackets in  boundary subtypes signal interval beginnings or endings
whereas the underscore symbol as left or right part of a subtype stands for nonempty
context, i.e. there is at least one terminal to the left or right.
\item Whenever terminal type $T$, transitively dominated by category $X\in 
  CategorySet$, is
  found in initial or medial or final position of the terminal yield
  of $X$, add a conjunct
  $[X$ or $\_X\_$ or $X]$ to $T$. The `or' is `inclusive OR', in particular to cover
  length-1 terminal yields.
\item Whenever terminal type $T$ is not transitively dominated by category 
  $X\in CategorySet$, add a conjunct $\neg X$ to $T$.
\item Add a conjunct $\neg {[}X{]}$ to all terminals whose type formula
  is not maximally specific with respect to category $X$. This step ensures
  full specification for  boundary occurrences in those intervals
  which either contain more than one atom or are not multiply dominated.
\end{examples}         
\end{examples}         
To exemplify: \ref{ext:sensational_tagged} shows a flat representation 
of the joint diagrams of \ref{ext:sensational} and
\ref{ext:sensational_2}. (Note that for abbreviatory purposes we 
assume here that $\Sigma^{(')}$ is the supertype of
$\Sigma^{(')}_w$ and $\Sigma^{(')}_s$). The reader may verify herself that we can
indeed recover the graph-structured version provided that the
dominance precedence relation $>_{dom}$ is known. 
\begin{Ex}
\item \label{ext:sensational_tagged} {\sc Flat representation
    of} sensational\\
\begin{multicols}{2}
$\text{\bf s}\wedge [O]\wedge[\sigma\_\wedge[\Sigma_w\_\wedge\neg\Sigma^{'}\wedge[\omega\_\newline
 \text{\bf\textepsilon}\wedge [N]\wedge\_\sigma\_\wedge\_\Sigma_w\_\wedge\neg\Sigma^{'}\wedge\_\omega\_\newline
 \text{\bf n}\wedge
 [C]\wedge\_\sigma{]}\wedge\_\Sigma_w{]}\wedge\neg\Sigma^{'}\wedge\_\omega\_\newline
 \text{\bf s}\wedge
 [O]\wedge{[}\sigma_s\_\wedge{[}\Sigma_s\_\wedge{[}\Sigma^{'}_s\_\wedge\_\omega\_\newline
 \text{\bf \textepsilon}\wedge
 [N]\wedge{\_}\sigma_s\_\wedge{\_}\Sigma_s\_\wedge{\_}\Sigma^{'}_s\_\wedge\_\omega\_\newline
 \text{\bf j}\wedge
 [C]\wedge{\_}\sigma_s{]}\wedge{\_}\Sigma_s\_\wedge{\_}\Sigma^{'}_s\_\wedge\_\omega\_\newline
 \text{\bf \textesh}\wedge
 [O]\wedge{[}\sigma_w\_\wedge{\_}\Sigma_s\_\wedge{\_}\Sigma^{'}_s\_\wedge\_\omega\_\newline
 \text{\bf \textschwa}\wedge
 [N]\wedge{\_}\sigma_w\_\wedge{\_}\Sigma_s\_\wedge{\_}\Sigma^{'}_s\_\wedge\_\omega\_\newline
 \text{\bf n}\wedge
 [CO]\wedge{[}\sigma_w{]}\wedge{\_}\Sigma_s{]}\wedge{\_}\Sigma^{'}_s\_\wedge\_\omega\_\newline
 \text{\bf \textschwa}\wedge
 [N]\wedge{\_}\sigma_w{\_}\wedge\neg\Sigma\wedge{\_}\Sigma^{'}_s\_\wedge\_\omega\_\newline
 \text{\bf l}\wedge
 [C]\wedge{\_}\sigma_w{]}\wedge\neg\Sigma\wedge{\_}\Sigma^{'}_s{]}\wedge\_\omega{]}$
\end{multicols}
\end{Ex}

The impact of having a linearized distributed representation of
(prosodic) constituency is twofold. First, we can now refine prosodically
underspecified segmental strings with suitable constraints that 
specialize for each layer of the prosodic hierarchy. For example, given a
finite-state version of declarative syllabification
(\citeNP{walther:92}, \citeNP{walther:95}) for predicting syllabic
roles from segmental information (itself using an intermediate layer of sonority 
difference information), the next layer would use local
syllable role configurations
 to demarcate syllable boundaries
${[}\sigma,\sigma{]}$ and syllable interior $\_\sigma\_$%
\footnote{For example,\/  by way of the following monotonic rules: $O\vee N
  \rightarrow {[}\sigma\, /\, \neg O\, \underline{\phantom{xx}}$, $N\vee C
  \rightarrow \sigma{]}\, /\, \underline{\phantom{xx}}\, \neg C$, $O
  \rightarrow \_\sigma\_ \, / \, O \vee CO \underline{\phantom{xx}}$,
  \phantom{x}$N \vee C \rightarrow \_\sigma\_ \, / \, \underline{\phantom{xx}} C
  \vee CO$. On automata representations of monotonic rules, see
  \citeN[34f]{bird.ellison:92}. 
  The disjunctions in the preceding rules can even be eliminated with a suitable featural
  decomposition of syllable roles using the features [$\pm$onset] and
  [$\pm$coda] \cite[\S 3.4.3]{walther:97}.}
, and so forth. 

Second, we can freely use this locally encoded prosodic information 
to condition both generic and construction-specific constraints that
must reflect some dependency on a given prosodic
configuration. \citeN{eisner:97} illustrates, from the
perspective of his Primitive Optimality Theory, just how appealing
such local encodings can be for purposes of compact constraint
formulation. Because in his results the emphasis is on locality in
representational formats rather than on violability, one can be confident that their advantages
will be preserved in the present framework. 
\subsection{Parsing}
So far we have described OLPM from the perspective of generation
only. Because reversibility is usually held to be an important
property in practical applications of finite-state networks, we will
now briefly consider how to do parsing under the new framework.

Disregarding optimization at first, parsing seems next to trivial. The 
central mechanism for constraint combination is automaton
intersection, an associative operation that supports
reversibility. Under this view one would simply intersect the string
to be parsed with the FSA constituting grammar and lexicon; a nonempty 
result would then signal successful recognition. In the face of a
structure-building grammar that adds e.g.\/ syllable role information
or other prosodic annotations, we of course should represent the parse
string symbols as prosodically underspecified segmental types to allow for
their subsequent specialization in the process of intersection.

However, an immediate complication is that in OLPM the technical arcs 
{\em skip, repeat} would prevent literal matching of even structurally 
underspecified surface strings with the grammar. If e.g.\/ some reduplicated form is to be
recognized, the grammar will assign several repeat arcs as part of the
`surface' string, and this decorated surface form will then fail to
intersect with the plain, undecorated parse string. The solution is to 
employ a trivial preprocessing step at the interface between phonetics 
and phonology: enrich the automaton corresponding to an undecorated
parse string with consumer self loops that tolerate exactly the set of
technical symbols. Note that automata enriched in this way are still
rather different from transducers, even granting a simulation
of composite arc symbols $x^{\underrightarrow{input{:}output}}z$ as consecutive arcs
$x^{\underrightarrow{input}}y^{\underrightarrow{output}}z$ in conventional
automata (cf.\/ fig.\/ 9 in US patent 5,625,554 granted to Xerox on
April 29, 1997). This is evident from   the fact that, in contrast to
their behaviour in the simulation, odd arcs do not    consistently
play the role of inputs and neither can even arcs    be seen as corresponding outputs. 

A second issue is that in parsing one would normally want a little more than mere 
recognition of grammatical forms (and rejection of ungrammatical
ones), namely categorial information in the form of morphological,
syntactic and semantic properties or features. Although we have been
silent on this issue up to now, it is actually simple to represent the 
required annotations in grammar and lexicon by reserving one or more 
final arcs at the end of automata for appropriate category labels, just like in the
transducer-based proposals of
\citeN{karttunen.kaplan.zaenen:92}. 
(Categorial information is again supposed to be pairwise disjoint from
technical and segmental type information.) However, unlike in the FST
version, where one can map underlying categorial information to the
empty string $\epsilon$ on the surface, in our one-level version this information would again
be visible in the surface string. As a consequence, the above
preprocessing step needs to be slightly modified to tolerate
categorial information in those self loops that are attached to final
states. Finally, the parse string itself constitutes an unconfirmed hypothesis that 
needs verification by independently produced grammatical and lexical
resources. This means that -- at least if self-loop enrichments are present in the
grammar -- it is necessary to formally mark each segment of the parse
as a {\em consumer}. Only when parse-segments-as-consumer-hypotheses 
intersect with matching producer segments from the lexicon, will they
survive phase II of the two-stage evaluation procedure outlined in \sref{resource}.

The following definition of a \code{parse} operator in \ref{ext:parse} 
accurately reflects the preceding
discussion. Because it makes use of the notational format of the
Prolog-based FSA toolbox that will only be introduced later in \sref{ex},
the reader is urged to come back to this section on a second
reading. Note in particular the interspersed self loops, defined via
the Kleene star operator \code{*} and the intersection \code{\&} of the
\code{preprocessed(ParseString)} with \code{grammar\_and\_lexicon}. 
\begin{Ex}
\item \label{ext:parse}{\sc Parsing (in the absence of
    optimization)}\\[-3ex]
\startpiece
\begin{verbatim}
preprocessed([SurfaceSegment|RestSegments]) := 
                preprocess(RestSegments, SurfaceSegment).

preprocess([], LastSegment) := 
   [consumer(LastSegment & segment),
    consumer((technical_symbols ; categorial_information)) *].

preprocess([CurrentSegment|RestSegments], PreviousSegment) := 
   [consumer(technical_symbols) *,
    consumer(PreviousSegment & segment) |
    preprocess(RestSegment, CurrentSegment)].       

parse(ParseString) := 
     closed_interpretation(preprocessed(ParseString) & 
                                      cache(grammar_and_lexicon)).
\end{verbatim}
\stoppiece
\end{Ex}

As is to be expected, extending the OLPM parsing task to cover optimization adds new
complications. We can no longer be sure that a nonzero intersection with
grammar and lexicon signals grammaticality; such a result merely means 
that the parse string is consistent with one member of the set of
alternatives to be optimized over. To be sure, this is still a welcome 
improvement over the parsing problem that would obtain in an
all-default framework like OT, where the notion of consistency plays
no role at all. However, it means that a second step must be added to
the \code{parse} step from \ref{ext:parse}. 

That step consists first of the extraction of categorial information from
the annotated parse string and then using that information to {\em
  generate} an optimal surface result via application of Bounded Local 
Optimization. If this optimal result and the preprocessed parse string 
intersect, fine; if not, it means that the parse string is
ungrammatical. 
Extraction itself can be performed by composing (\code{o}) the annotated parse
string with a simple transducer that maps segmental symbols
to their maximally underspecified representatives and preserves the
identity of all other symbols. With some caching of intermediate
results we can prevent doing double work in our optimizing
parser. Note also the use of Bounded Local Optimization \code{blo}
which needs to know its \code{LookaheadConstant}.
Here then again comes a code fragment that shows
\code{optimizing\_parse} in all its glory:
\begin{examples}
\item \label{ext:optparse}{\sc Parsing in the presence of
    optimization}\\[-3ex]
\startpiece
\begin{verbatim}
extraction := [ { identity(consumer(repeat)),      % type identity
                        identity(consumer(skip)),  % ... ditto
                        identity(consumer(segment))% ... ditto
                      } *,
                      $@:$@ *   % token identity 
                                % elsewhere, i.e.
                                % in mapping
                                % categorial info!
                    ].

optimizing_parse(String, LookaheadConstant) := 
    blo(  ( cache(parse(String))
                  o
              extraction
          )
          & grammar_and_lexicon,
          LookaheadConstant)
    & parse(String).
\end{verbatim}
\stoppiece
\end{examples}
We could call the preceding proposal a kind of analysis-by-synthesis approach
(cf.\/ \citeNP{walther:98} for further
discussion in a feature-logical setting). Given these initial results, 
there clearly is a need for further research into parsing under
optimization. In particular, one should investigate its efficiency in
realistic cases and conduct a careful implementation that makes use of lazy
automaton intersection \cite{mohri.pereira.riley:98}.  
\section{Implemented Case Studies}\label{ex}
In this section I will discuss worked examples from three languages
that illustrate the interplay of the various enrichments and
mechanisms proposed above. To be maximally concrete,  snippets from
the actual implementation are provided. The notational format is that
of the FSA Utilities toolbox \cite{vannoord:97}, a subset of which is depicted
in \ref{ex:regex}.
\begin{Ex}
\item \label{ex:regex} {\sc Format of regular expression operators}\\
\begin{tabular}[t]{cl}
\tt [] & empty string \\
\tt \{\} & empty language \\
\tt Lower:Upper & pair \\
\tt [E1,E2,\dots,En]    & concatenation of \tt E1,E2\dots,En \\
\tt \verb+{+E1,E2,\dots,En\verb+}+    & union of \tt E1,E2,\dots,En\\
\tt E*                  & Kleene closure\\
\tt E+                  & Kleene plus (\tt [E,E*]) \\
\tt E\verb+^+                  & optionality\\
\tt E1 \verb+&+ E2      & intersection\\
\tt RelA o RelB & composition \\
\tt identity(E) & identity transduction \\
\end{tabular}
\end{Ex}
Significantly for our purposes, FSA Utilities  offers the possibility
to define new regular expression operators. Departing from the
original \code{macro(Head,Body)}
notation I use the infix expression \code{Head := Body} -- to be read
as ``\code{Head} is substituted   by \code{Body}'' -- for reasons of better readibility.
Macro definitions may be parametrized with the help of Prolog variables in order to
define new regular expression operators in terms of existing
ones. Also, Prolog hooks in the form of definite-clause attachments
are provided to help construct more complicated expressions which
would be too cumbersome to build using the above facilities
alone. Finally, it is sometimes of importance that regular expression
operators are alternatively definable through direct manipulation of the
underlying automata. Again, here the toolbox provides abstract data
types that support access to alphabets, states, transitions etc. 

I will take liberty in sometimes suppressing macro definitions whose details are not
essential to the discussion at hand, resorting to descriptions in
prose instead. Also, for the sake of brevity the type hierarchy that structures the alphabet
will not be displayed separately, which can be justified on the ground 
of mnemonic type names that make it obvious what the hierarchy would
be like.
\subsection{Ulwa construct state infixation}
Ulwa is an endangered Misumalpan language spoken in Eastern 
Nicaragua. %
The purpose of this section is to analyze the placement of possessive
infixes in nouns, since ``Ulwa serves as a nice example of a language
in which infixation is clearly sensitive to prosodic structure'' \citeN[49]{sproat:92}.
While Ulwa data have been discussed in the literature for some time
(e.g. \citeNP{mccarthy.prince:93} and \citeNP{sproat:92}), an
up-to-date descriptive reference has only recently become available
\cite{green:99}. Green shows that Ulwa nouns can participate in a
syntactic construction called {\em construct state}, ``a cover term for
an entire paradigm of genitive agreement inflection'' (ibid., 78) where the
head noun is marked morphologically by affixation. The affix shows
inflection for person and number \ref{ex:cns}. The primary semantics
expressed by the construct is possession. 
\begin{Ex}
\item \label{ex:cns} {\sc Forms of the Construct-state Affix}\\
\begin{tabular}[t]{l|ll}
{\em Person} & {\em sg.} & {\em pl.} \\ \hline
1st & -ki- & \begin{tabular}[t]{@{}ll}  -ki-na & {\scriptsize exclusive}\\
  -ni- & {\scriptsize inclusive}\end{tabular} \\
2nd & -ma- & -ma-na \\
3rd & -ka- & -ka-na-
\end{tabular}
\end{Ex}
\ref{ex:ulwa} shows some data for the third person singular affix \ling{(-)ka-},
collected from \citeN[105]{mccarthy.prince:93} and
\citeN[49]{sproat:92} and checked against the dictionary in appendix B 
of \citeN{green:99}.%
\footnote{Sproat additionally cites \ling{gaad, gaad-ka} `God', while Green's
  dictionary completely lacks g-initial words; this is because he concludes that
  -- given only a single native counterexample, \ling{aaguguh} `song,sing' -- /g/ is not a
  phoneme of Ulwa. There is no contradiction here because Green acknowledges that
  /g/ exists in a few obvious loan words.
 McCarthy \& Prince erroneously cite 
 the form \ling{kulu-ka-luk} from pseudo-reduplicative \ling{kululuk}
  `lineated woodpecker', which \citeN[54f]{green:99} marks as
ungrammatical since ``speakers seem to recognize them as reduplicative 
in form, making these stems resist the infixation process.''}
Long vowels are represented as \ling{VV}, which both simplifies the statement of heaviness
and eases the actual analysis.
\begin{Ex}
\item \label{ex:ulwa} {\sc Ulwa construct state suffixation/infixation}\\
\begin{tabular}[t]{lll}
\underline{\phantom{/}N\phantom{/}} & \underline{his/her/its N} \\[1ex]
bas & bas-ka & `hair' \\
kii & kii-ka & `stone, rock' \\
taim & taim-ka & `time' ({\em preferred: aakatka})\\
sapaa & sapaa-ka & `forehead' \\
suulu & suu-ka-lu & `dog' \\
asna & as-ka-na & `clothes, dress' \\[1ex]
paumak & pau-ka-mak & `tomato' \\
waiku & wai-ka-ku & `moon, month' \\
siwanak & siwa-ka-nak & `root' \\
arakbus & arak-ka-bus & `rifle,gun' ({\em Spanish: arquebus}) \\
\end{tabular}
\end{Ex}
According to the descriptions of \citeANP{mccarthy.prince:93} and \citeANP{sproat:92},
primary stress in Ulwa falls on the first syllable, if it is heavy, otherwise 
on the second syllable from the left. In Ulwa, the core syllable
template is (C)V(V)(C) with a small set of exceptions that exhibit
complex onsets or codas. Syllables count as
heavy iff they are either closed off by at least one consonant
(\ling{bas}) or contain more than one vowel 
(\ling{kii,pau,taim}). Monosyllabic words are always heavy. 

From this description and the data in \ref{ex:ulwa} alone it would
follow that the possessive affix is invariably located after the
stressed syllable, emerging as a suffix 
after heavy monosyllables and as an infix otherwise. Note that the
affix itself is never stressed. The immediate goals of the
analysis to be developed below will then be to formalize both this stress distribution and the 
morphemes involved.

Before we can do that, however, we should note that the fuller picture 
that \citeN{green:99}  presents both for the infixation/suffixation
behaviour and the stress facts does add some complications. There are
many cases of free variation between suffixation and infixation
(\ling{kubalamh-ki $\sim$ kuba-ki-lamh} `butterfly'). Stems show
two-way exceptions, some taking suffixes exclusively although
infixation should {\em a priori} be allowed (\ling{tiwiliski-ka,
  *tiwi-ka-liski} `sandpiper'), a small set also tolerating
infixation although it should {\em a priori} be ruled out
(\ling{ta-ka-pas} `mouth'). The same goes for stress which can
sometimes oscillate (\ling{\textprimstress baka, ba\textprimstress
  kaa} `child'), while on other occasions preceding
(\ling{\textprimstress\ip{sariN}} `avocado') or following
(\ling{tas\textprimstress laawan} `needlefish') the locus predicted
above. Interestingly, construct state formation may involve
accentuation of the affix in a few exceptional cases
(\ling{ma-\textprimstress ka-lnak} `payment'),
can even disambiguate alternating stress (\ling{ba\textprimstress kaa-ka},
*\ling{\textprimstress baka-ka} `child') and cause stress shift in a number of
pseudo-reduplicative root shapes (\ling{ki\textprimstress liilih}
\ling{kili\textprimstress lih-ka} `cicada'). We refer the reader to Green's
extensive discussion for further study, concentrating on the core
cases in the analysis to follow.

In a first step, the original, disjunctive formulation of the stress generalization
can be simplified by stating that {\em the syllable containing the
second mora from the left must be stressed}. According to moraic theory \cite{hayes:95}, a
mora $\mu$ is an abstract unit of syllabic weight which figures prominently in 
the analysis of stress systems of many of the world's languages. Thus,
reference to moras  is wellfounded in our context. To exemplify:
$ba_\mu s_\mu$, $ta_\mu i_\mu m_\mu$, $a_\mu s_\mu .na_\mu$ all receive 
stress on the first syllable, while $sa_\mu .pa_\mu a_\mu$, $ku_\mu
.lu_\mu .lu_\mu k_\mu$ are accented on the second syllable.

To facilitate identification of moras, we will take an intermediate
step by tagging each segment with the {\em relative difference in sonority}.
That is, we will mark whether sonority is rising, falling or level
when comparing each segment  with its right neighbour. Recall that sonority is an abstract  
measure of intrinsic prominence for speech sounds. While it is
customary to employ sonority for determining full syllable
structure, which in turn then serves as the input to foot structure
and stress computation, it is possible to bypass all higher-level structure in the
case at hand.%
\footnote{As an aside, note that evidence for higher-level prosodic structure
  above the syllable role level is often surprisingly weak, in stark
  contrast to the wholesale adoption of the entire prosodic hierarchy
  \cite{mccarthy.prince:91} throughout most of the generative literature. In the case of Ulwa, for example,
  \citeN[64]{green:99} admits that iterativity in
  noun stress -- usually held to be a basic reflex of foot formation -- rests on
  inconclusive evidence from three forms only. 

Also, for a full phonological grammar encompassing syllabification, computation
of relative sonority differences is independently needed,
since it forms  an essential first step in the declarative syllabification schemes of  
\citeN{walther:93}, \citeN{walther:97}. While these were originally
couched in feature logic, a non-weighted finite-state version is both easy to implement
and attractive due to its conceptual simplicity, as there is no need for a
simulation of the Maximum Onset Principle (which complicated
\citeN[139]{mohri.riley.sproat:96}'s weighted finite-state syllabification for
Spanish). However, the details are beyond the scope of this paper.} 
 Also, for current
purposes we can conflate most of the distinctions of
\citeN[211]{blevins:95}'s nine-positional sonority scale, keeping only
$consonant \ll vowel$. Given that scale, each segment is tagged with one of $\{up,
down\}$, where the tag depends on the sonority value of its right
neighbour: if the right segment's prominence is higher, $up$ is used, while $down$ is 
assigned in the case of lower or same sonority. The final segment,
which has no natural right neighbour, is marked with $down$. To
exemplify: the Ulwa word for `clothes' will be tagged
$a_{down}s_{down}n_{up}a_{down}$ and `stone' is marked as
$k_{up}i_{down}i_{down}$. The crucial observation now is that moraic
segments are exactly those that are tagged with $down$. 

This observation has obvious repercussions on the formalization of Ulwa stress below:
\startpiece
\begin{verbatim}
material_is(Spec) := [consumer(Spec) *].
stress := sonority_differences &
         [ pre_main_stress, main_stress +, post_main_stress ^ ].
pre_main_stress := [non_moraic *, mora, non_moraic *] & 
                             material_is(unstressed).
post_main_stress := [non_moraic, material_is(anything)] & 
                              material_is(unstressed).
non_moraic := consumer(up).
mora(Spec) := consumer(down & Spec).
mora := mora(anything).
main_stress  := mora(stressed).
\end{verbatim}
\stoppiece
Unsurprisingly, \code{stress} is built upon computation of \code{sonority\_differences}.
Note next how the stress pattern itself is initially decomposed into zero or more unstressed 
non-moraic onset segments followed by the first mora followed by more
non-moraic material (\code{pre\_main\_stress}). Thereafter comes an
obligatory stretch of stressed moraic 
material delimitated by an optional block of post-stress 
segments whose start is signalled by a non-moraic segment. Observe
that there will be multiple adjacent stress marks if the syllable hosting the second
mora is not monomoraic. In other words: the whole rime of 
the accentuated syllable is formally marked as stressed
(e.g. $b_{unstressed}a_{stressed}s_{stressed}$), in what 
amounts to the explicit  linear equivalent of the stress feature percolation  or
structural referral to $\sigma_s$  that is implicit in more traditional
approaches. Also, stress is contingent  
on the availability of independently introduced lexical material, hence everything
is encoded as \code{consumer}-type information.

With stress assignment already given, it is now fairly easy to
define the affix itself.
\startpiece
\begin{verbatim}
possessive_third_singular :=
    add_repeats(contiguous([consumer(stressed), 
    producer(k & unstressed), producer(a & unstressed)])).
\end{verbatim}
\stoppiece
Its segmental content \ling{ka} is simultaneously marked as
unstressed, in accordance with the surface facts (modulo the small 
number of exceptional words of type \ling{ta-\textprimstress kaa-pas}
  mentioned above, for which an allomorph would have to be set up). Of
  course, this is \code{producer} 
information.  Additionally, in this analysis the affix receives a prosodic
subcategorization frame: its left context restriction mentions an
immediately adjacent stressed segment. As a
contextual requirement, it must be encoded using the \code{consumer} macro.
The whole tripositional sequence is wrapped with two more macros: \code{contiguous}
introduces edge-only self loops  to permit infixal behaviour on the
one hand while disallowing internal breakup of \ling{-ka-} on the
other hand. Ob top of this \code{add\_repeats} modifies the
 automaton as described in \sref{ext} to allow for possible uses in reduplication. 
According to \citeN{green:99}, Ulwa does indeed exhibit reduplicative
constructions, although the details are beyond the scope of this section.

Encoding of stems is not complicated, yet contains some points
worth noting:

\noindent\begin{minipage}[t]{\linewidth}
\startpiece
\begin{verbatim}
discontiguous_lexeme(L) := 
  add_repeats(discontiguous(lexeme(L) & stress)) .
hair := discontiguous_lexeme("bas").
forehead := discontiguous_lexeme("sapaa").
root := discontiguous_lexeme("siwanak").
gun := discontiguous_lexeme("arakbus").
\end{verbatim}
\stoppiece
\end{minipage}
The first point is that, of course, stems  must tolerate discontiguity in order to
host infixes: \code{discontiguous} therefore adds self loops at all
positions, not only at the edges.%
\footnote{Recall that this refers to the core cases; exceptions that ban infixation would require 
a different treatment. One way would be to parametrize {\tt
  discontiguous} for  the actual content of  self loops -- currently
the entire alphabet $\Sigma$ -- which then could be restricted to
technical symbols that are incompatible with the segmental content of affixes.}

 The innermost \code{lexeme} macro converts a
Prolog string into a concatenation of producer-type segmental
positions, working off the assumption that all symbols represent defined
segmental types. 

Now the most interesting second aspect is that the
undecorated string automaton corresponding to the lexeme itself is
 intersected with the \code{stress} constraint. This is nothing but the
 constraint-based equivalent of a lexical rule application for stress
 asignment, applied to stems in isolation. The reason for 
assuming lexically stressed stems is that we want to rule out coalescence, i.e. 
amalgamation of segmental material, between affix and stem
segments. As the affix contains the segments \ling{k} and \ling{a}, it could in
principle overlap the \ling{k} in `gun' (*\ling{ara-\underline{k}a-bus}) or the
(pen)ultimate \ling{a} in `forehead' (*\ling{sap-k\underline{a}-a}, *\ling{sapa-k\underline{a}}). 
Note that such overlapping placement would still satisfy the affix's
prosodic requirement, as the immediately preceding segment in these
examples is indeed the second mora from the left and therefore would
be surface-stressed.
To be sure, coalescence as such is attested in other languages (e.g. in
Tigrinya, \citeNP{walther:97}). However, it must be forbidden in our
Ulwa construction. The solution is now easy to understand: by lexically stressing
the stems, the left context of the infix in the ungrammatical coalescent
realizations of `gun' and `forehead' is fixed to
\code{unstressed}, hence will properly {\em conflict} with the \code{stressed}
requirement of the possessive affix and lead to the elimination of the 
illformed disjuncts. Note that another, morphological solution to the 
coalescence problem would have been to tag affix and stems
contrastively, e.g. as ${-}k_aa_a{-}$ versus $s_sa_sp_sa_sa_s$
 (cf. \citeNP{ellison:93}). Interestingly, at least in this case such purely
technical diacritics seem not be required; rather, we can profit from a
clean phonological solution.

There is one last aspect which needs our attention, and that concerns banning
discontiguous main stress. Even with lexical stress assignment, a
heavy stem syllable such as \ling{\'a\'s} receives two formal stress marks, 
and -- being interruptible -- could therefore satisfy the prosodic requirements of the
possessive in two ways: *\ling{\'a-ka{-}\'s},
!\ling{\'a\'s-ka{-}}. Continuing with our phonological solution,
the cure is again immediate: surface and lexical stress must not
conflict! Since the formulation of \code{stress} ensures
contiguous main stress by way of the kleene plus operator, and because it is
descriptively true that  lexical stem stress %
coincides with the word-based surface stress pattern (again modulo the
exceptions noted above), we can simply  impose the stress 
constraint once again on the whole infixed word to ensure full
wellformedness of the Ulwa possessive construction: 
\startpiece
\begin{verbatim}
word(Stem) := Stem & possessive_third_singular & stress.
stems := {hair, forehead, root, gun}.
possessive_nouns := closed_interpretation(word(stems)).
\end{verbatim}
\stoppiece
Note that it is the intersection of all constraints that defines a
\code{word}. After expanding that macro in the body of \code{possessive\_nouns} 
with (the disjunction of) defined stems as actual argument together with pruning away of
unsatisfied consumer arcs through \code{closed\_interpretation}, we
arrive at an automaton which contains exactly the desired grammatical surface
strings \ling{\{baska, sapaaka, siwakanak, arakkabus\}}, of course
enriched with stress and sonority information. 

At this point an additional  remark seems appropriate. 
Thomas Green's reference work summarizes the underlying cause of the
construct-state phenomenon in derivational parlance as follows: ``\dots the construct
morphology itself does not receive stress, and does not cause shifts
in the stresses of the material which follows it \dots it is as if the infixation takes place at a
point in the derivation {\em after} the metrical structure [i.e.,
stress, M.W.] of the word has been determined'' \cite[64f]{green:99}.
I take it to be quite satisfying that the iterative process of formal 
grammar development, while seemingly being driven by technical
problems very unlike those of a descriptive
grammarian, nevertheless led to the same fundamental conclusions.

The observant reader will have noted that the analysis presented so
far is not based on the notion of drift, hence does not need to make any use 
of optimization and the concomitant representational extensions. As
promised in \sref{ext}, we will now show what the drift-based
alternative looks like.%
As it turns out, the
experience is highly instructive and sheds new light on the pros and
cons of optimization.

The first step in such an optimization-based analysis of the same
facts of Ulwa construct-state infixation consists in distributing the
appropriate weights to both the affix and the stem, which in turn is
contingent on the direction of drift we would like to see in the data.
Contrary to what \citeN[107]{mccarthy.prince:93} suggested with their
use of the {\sc Rightmostness} OT constraint, I would argue that the
proper direction is leftward. The reasoning is as follows: with
lexical stem stress given as before, a leftward-drifting affix will
correctly `float' towards the accented position.  To prevent the affix 
from floating {\em past} that position -- which would immediately
cause disruption of the lexical stress pattern -- we again simply impose the same stress
constraint on the surface word form. In order to formalize leftward
drift, the affix material must be weighted cheaper than the stem, hence
we will use two weights-as-types \code{unmarked} $\ll$ \code{marked} to that effect.

Here then is the exchanged portion of the stem-defining macros:
\startpiece
\begin{verbatim}
discontiguous_lexeme(L) :=
  add_repeats(discontiguous(lexeme(L) & 
  material_is(marked) & stress)) .
\end{verbatim}
\stoppiece

The crucial difference to the previous analysis now is that the affix does not need to be
prosodically subcategorized! Here is the new definition:
\startpiece
\begin{verbatim}
possessive_third_singular :=
    add_repeats(contiguous(lexeme("ka") & 
    material_is(unmarked & unstressed))).
\end{verbatim}
\stoppiece
The \code{unmarked} tagging of affixal tagging is the hallmark of
leftward drift, as expected; no further mention of stress is needed.
The definition of \code{word} does not need any changes, since as
already noted the same need of avoiding surface discontiguities in main stress arises in 
the present analysis. The only remaining difference is, of course,
that bounded local optimization (\code{blo}, with look-ahead 1) needs 
to be formally applied on the resulting, minimized  (\code{mb}) automaton:
\startpiece
\begin{verbatim}
optimized_possessive_noun(Stem) := 
   blo(mb(closed_interpretation(word(Stem))),1)).
\end{verbatim}
\stoppiece
With  stems like \code{hair}, \code{gun} etc. as actual parameters, the same
results obtain as in the previous analysis. It is instructive to see
what the automaton for Ulwa `gun' -- encoding \mbox{\{%
\ling{*arakbuska}},\ling{*arakbukas}, \mbox{\ling{arakkabus}\}} -- looks like before
optimization \ref{ex:gunaut}.
\begin{Ex}
\item \label{ex:gunaut} {\sc Weighted automaton for Ulwa `gun'} \\
\epsfig{file=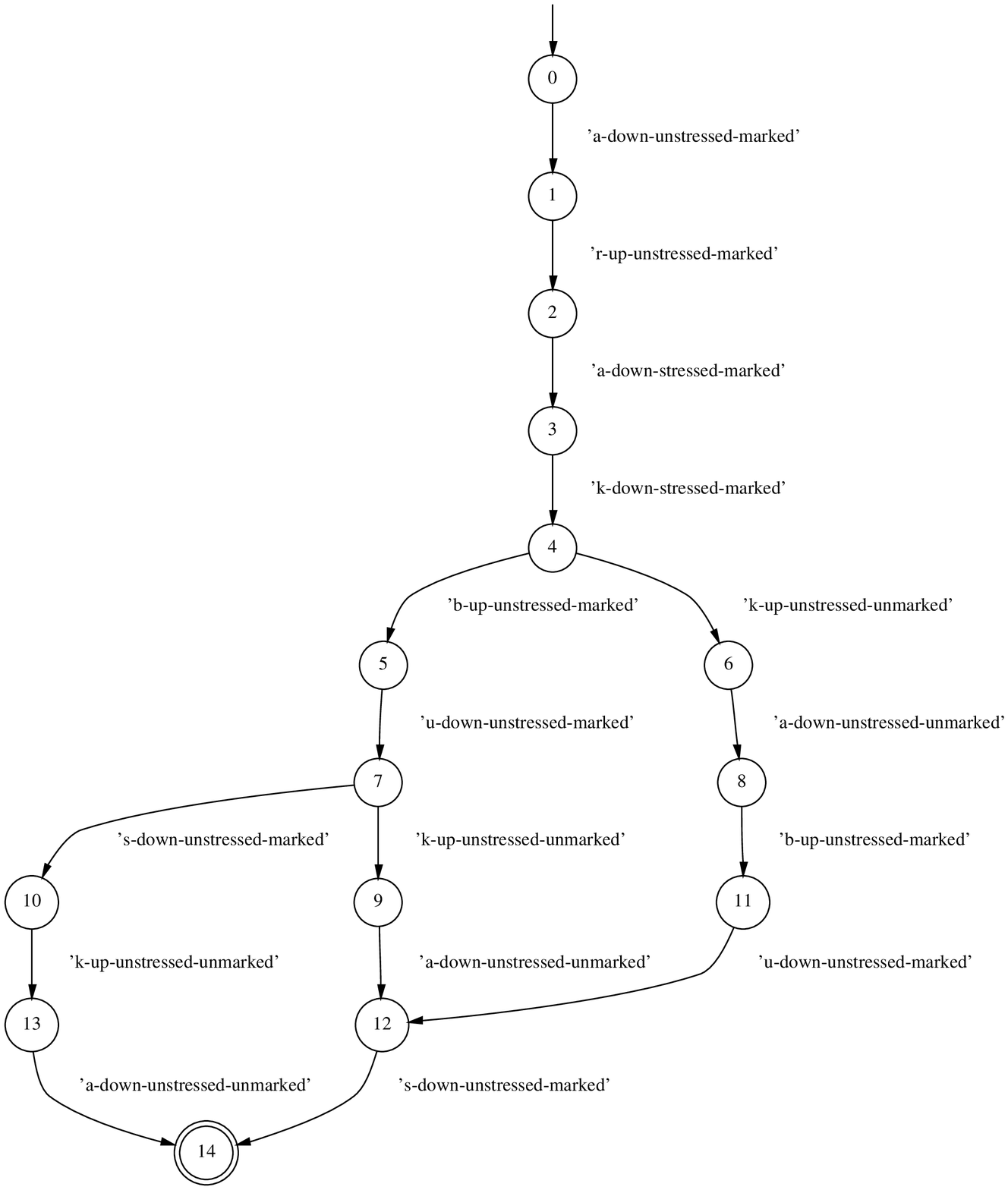,width=.99\linewidth}
\end{Ex}
Note how at each bifurcation in the graph the immediate alternative is between
an \code{unmarked} affix versus a \code{marked} stem segment; hence
look-ahead $k=1$  indeed suffices here. Also, as a byproduct of pruning arc $4
\rightarrow 5$ first, the algorithm in \ref{ext:bloalg} will never explore
the set of states $\{5,7,9,10,13\}$; thus the incomplete search
embodied in BLO does indeed bear fruit in practical cases.

With both analyses in place, let us sum up. The drift-based optimizing
analysis probably holds some special appeal to theoretical linguists
because one can dispense with affix-specific prosodic subcategorization and cling to the `lean
lexicon' view that is popular in much of generative linguistics. 
Moreover, it could be argued to be slightly more explanatory in that it sees
nondisruption of lexical stress, a global desideratum for words, as
the sole prosodic factor which drives the placement of Ulwa construct-state affixes.%
\footnote{With only slight changes to the formulation of the stress constraint, one
  could go even further and claim that the unstressedness of the
  construct-state inflectional affixes itself is derivable as well: having less
  than two moras, CV syllables like \ling{ki,ka,ma} will not receive primary
  stress when viewed in isolation, e.g. in their lexical entry form!}
One disadvantage of this analysis from a formal point of view is that it
requires a canonical, i.e. minimized automaton format for application of BLO.
BLO itself must be counted as an additional ingredient in the analysis which 
furthermore precludes simple computation of a whole lexicon, as
noted above. 

The non-optimizing alternative, on the other hand,
avoids all the problems associated with optimization, but at the
expense of greater representational cost: the affix specification
requires explicit mention of the left-adjacent stress peak. This in
turn makes it harder to justify why exactly this subcategorization
happens to be crucial in Ulwa morphology, and why 
e.g. unstressedness on the third segment to the right would not be an equally 
plausible candidate for a possible prosodic subcategorization. On the
positive side one should realize that local, surface-detectable
properties such as adjacency to a stress peak would probably be rather easily learnable for
both child and machine. If proven to be feasible in future
research, such a learnability result would have an important impact on 
the ongoing debate about richness of lexical representations versus the
need for optimization. %
\subsection{\label{ex:i-formation} German hypocoristic truncation}
We briefly mentioned in \ref{intro:examples} that German provides a
productive form of truncation for hypocoristic forms of proper
names. Taking up the subject again, let us look at a representative
sample of the data in \ref{ex:hypo}.
\begin{Ex}
\item \label{ex:hypo} {\sc German hypocoristic i-truncation} \\
\begin{tabular}[t]{llll}
a. & Pet{\bf ra} /\ip{"pe:tKa}/ $>$ Pet-i & b. &  And{\bf reas}
/\ip{an"dKe:as}/$>$ And-i \\
c. & Gab{\bf riele} /\ip{gabKi"e:l@}/ $>$ Gab-i & d.  & Pat{\bf rizia} /\ip{pat"Ki:tsia}/$>$ Patt-i \\
e. & Gorb{\bf atschow} /\ip{"gOKbatSOf}/ $>$   & f. &
Chruschtsch{\bf ow} /\ip{"kK\;UStSOf}/ $>$ \\
 & Gorb-i & & Chruschtsch-i \\
g. & Imk{\bf e} /\ip{"Imk@}/ $>$ Imk-i & h. & Hans /\ip{"hans}/$>$ Hans-i
\end{tabular}
\end{Ex}
All derived forms in \ref{ex:hypo} end in \ling{-i}, and all
polysyllabic ones truncate some portion of their base (truncated part shown in boldface).

Previous analyses so far have sought to establish a connection between the cutoff
point in truncation and syllable
structure. \citeN{neef:96} and
\citeN{werner:96} proposed that the initial portion of the base preceding the
\ling{-i} must form a `potential maximal syllable'. The qualification
`potential' is significant here, because -- as \ling{Gab.ri.e.le} `female first name' vs.\/ 
\ling{Gor.bat.schow} `Gorbatchev' show -- reference to actual base syllabifications
would make wrong predictions (*\ling{Gor-i}, !\ling{Gab-i}, because
\ling{.Gorb.} is a maximal syllable, but *\ling{.Gabr.} with reversed
consonantal cluster is not). However, \ref{ex:hypo}.f,g show that the maximal syllable
approach misses some crucial data: *\ling{.Chruschtsch.}, and
*\ling{.Imk.} are illformed as syllables of German, yet their
\ling{i}-suffixed versions are the correct hypocoristic
forms. \ref{ex:hypo}.f also rules out another proposal
\cite{fery:97}, namely that the relevant criterion should
instead be one of `simple second syllable onset':
\ling{Chrusch.tsch-i} has a complex two-member onset /\ip{tS}/.

In contrast to these proposals I claim that the simplest correct
analysis is again one which makes direct use of the subsyllabic
concept of sonority. To see the plausibility of this claim, let us
first assume that the sonority scale for German is as follows
\cite{wiese:95}: 
\begin{quote}\scriptsize obstruents $\ll$ nasals $\ll$ laterals $\ll$ rhotics $\ll$ high
  vowels $\ll$ nonhigh vowels
\end{quote}
Graphing sonority 
over time for the critical example \ling{Chruschtschow}
/\ip{kK\;UStSOf}/ in \ref{ex:chruschtschow},
\begin{Ex}
\item \attop{\setlength{\unitlength}{3947sp}%
\begingroup\makeatletter\ifx\SetFigFont\undefined%
\gdef\SetFigFont#1#2#3#4#5{%
  \reset@font\fontsize{#1}{#2pt}%
  \fontfamily{#3}\fontseries{#4}\fontshape{#5}%
  \selectfont}%
\fi\endgroup%
\begin{picture}(4197,2232)(196,-1561)
\thinlines
\put(916,-841){\vector( 1, 0){3465}}
\put(1216,-271){\line( 1, 1){570}}
\put(1801,284){\line( 4, 1){448.235}}
\put(2251,389){\line( 1,-2){336}}
\put(2581,-286){\line( 1, 0){285}}
\put(2866,-286){\line( 1, 0){330}}
\put(3196,-286){\line( 2, 5){371.379}}
\put(3601,629){\line( 2,-5){367.241}}
\put(3976,-286){\line( 1, 0){ 15}}
\put(3211,-1321){\vector( 0, 1){375}}
\put(1126,-961){\vector( 0, 1){1620}}
\put(196,509){\makebox(0,0)[lb]{\smash{\SetFigFont{12}{14.4}{\rmdefault}{\bfdefault}{\updefault}sonority}}}
\put(4336,-1141){\makebox(0,0)[lb]{\smash{\SetFigFont{12}{14.4}{\rmdefault}{\bfdefault}{\updefault}time}}}
\put(3001,-1561){\makebox(0,0)[lb]{\smash{\SetFigFont{12}{14.4}{\rmdefault}{\bfdefault}{\updefault}minimum}}}
\put(1201,-721){\makebox(0,0)[lb]{\smash{\SetFigFont{12}{14.4}{\rmdefault}{\bfdefault}{\updefault}k}}}
\put(1756,-721){\makebox(0,0)[lb]{\smash{\SetFigFont{12}{14.4}{\rmdefault}{\bfdefault}{\updefault}\textinvscr}}}
\put(2191,-736){\makebox(0,0)[lb]{\smash{\SetFigFont{12}{14.4}{\rmdefault}{\bfdefault}{\updefault}\textscu}}}
\put(2521,-736){\makebox(0,0)[lb]{\smash{\SetFigFont{12}{14.4}{\rmdefault}{\bfdefault}{\updefault}\textesh}}}
\put(2866,-751){\makebox(0,0)[lb]{\smash{\SetFigFont{12}{14.4}{\rmdefault}{\bfdefault}{\updefault}t}}}
\put(3130,-736){\makebox(0,0)[lb]{\smash{\SetFigFont{12}{14.4}{\rmdefault}{\bfdefault}{\updefault}\textesh}}}
\put(3500,-736){\makebox(0,0)[lb]{\smash{\SetFigFont{12}{14.4}{\rmdefault}{\bfdefault}{\updefault}\textopeno}}}
\put(3910,-736){\makebox(0,0)[lb]{\smash{\SetFigFont{12}{14.4}{\rmdefault}{\bfdefault}{\updefault}f}}}
\end{picture}
} \label{ex:chruschtschow}
\end{Ex}
\noindent we note next that the last
base segment to be retained in the hypocoristic form is the second
/\ip{S}/, which is located at the {\em first sonority
minimum}. A sonority minimum is defined as a
segmental position at which sonority goes upwards to the right while it does
not rise from the left. Observe that the leftmostness expressed by
`{\em first}'  is not redundant for bases with at least two syllables, because suffixing the
characteristic ending \ling{-i} to the unaltered base would create
another minimum. Inspection of the other forms in \ref{ex:hypo} reveals that the
sonority-minimum criterion in fact forms a surface-true generalization
over German hypocoristic truncations.%
\footnote{There are, however, three kinds of exceptions: 
  (i) base portion is not string prefix of base word: \ling{Birgit $>$ Bigg-i, *Birg-i
  (C1V1C2C3\dots $>$ C1V1C3\dots) E\textprimstress lisabeth $>$
  Liss-i, Barbara $>$ Babs-i} (ii) base portion extends past
  sonority minimum: \ling{Depressiver $>$ Depr-i (?Depp-i); As\textprimstress phalt-i,
  Bank\textprimstress rott-i, Be\textprimstress deut-i, Alterna\textprimstress tiv-i, Kom\textprimstress
  post-i, Ele\textprimstress gant-i, \textprimstress Erstsemest-i}
(iii) base portion stops before
  minimum: \ling{West/Ostdeutscher $>$   Wess-i/Oss-i, *West-i/Ost-i,
  Hunderter $>$ Hunn-i}. As far as I can tell, these exceptions are a problem for
  all other published analyses as well, which is to be expected given the intimate connection
  between sonority graphs and syllable structure.
  Neglected factors playing a role here seem to include the 
  influence of morphological structure in the form of
  (pseudo)\-compounding and avoidance of homonymy, the position of main
  stress, details of phonetic  realization and recognizability of the
  base. Therefore the proposed analysis should be viewed as the core
  component of a more complete account that integrates those additional
  factors.}

Given our sonority-based generalization, we can now put the pieces
together to create a formal analysis. 
Here are high-level definitions for our example stem and the
hypocoristic constraint:
\startpiece
\begin{verbatim}
chruschtschow := contiguous_lexeme("kRUStSOf").
hypocoristic :=  sonority_differences & 
                        [sonority_based_cutoff_point, 
                         truncated_part,
                         characteristic_ending].
\end{verbatim}
\stoppiece
Although all we really ask of a truncatable representation is
tolerance of skipping, \code{chruschtschow}  is  
defined as a \code{contiguous\_lexeme} for better reuse of existing
macros.
Stems will be intersected with a \code{hypocoristic} constraint which
-- apart from tagging segments with
\code{sonority\_differences} --  dissects a word into an initial portion 
up to a \code{sonority\_based\_cutoff\_point}, followed by a possibly
empty (!\ling{Hans-i}) truncated part  and finally an obligatory
\code{characteristic\_ending}. These macros are in turn defined as follows:
\startpiece
\begin{verbatim}
sonority_based_cutoff_point := first_(sonority_minimum).
truncated_part := [producer(skip) *].
characteristic_ending := producer(i).

sonority_minimum := [consumer(segment & ~ up), consumer(up)].
first_(X) := [not_contains(X), X].
\end{verbatim}
\stoppiece
Of these, only the definition of \code{first\_} is not entirely
straightforward. The idea here is to establish leftmostness by excluding via 
\code{not\_contains} any occurrences of a particular set of strings \code{X} in
the -- possibly empty -- material  {\em preceding} the realization of \code{X}.

\begin{figure}[H]
\begin{examples}
\item \label{ex:chru_aut} {\sc Unoptimized Automaton for hypocoristic} \ling{Chruschtschow} \\
\epsfig{file=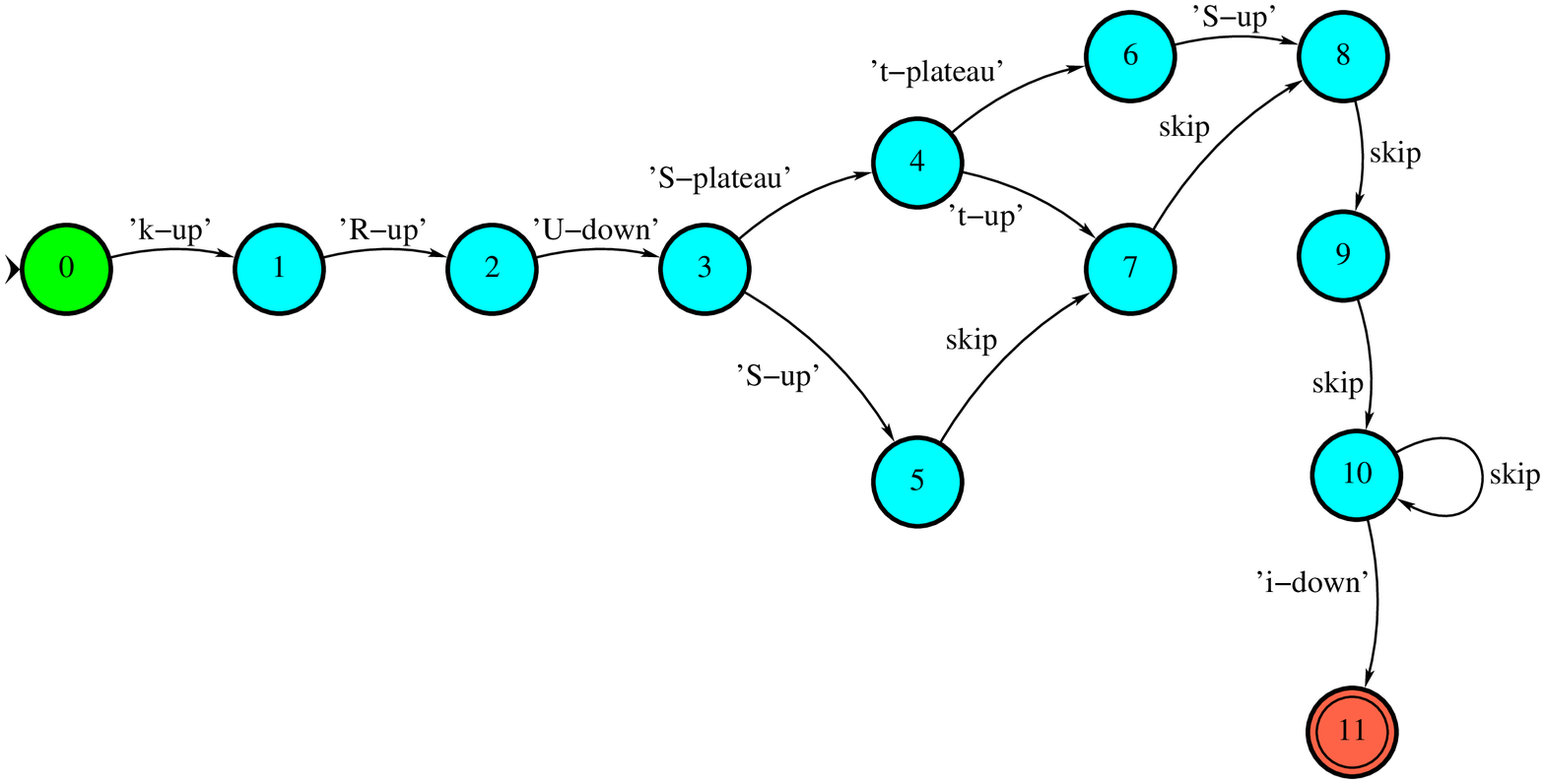,width=.99\linewidth}
\end{examples}
\end{figure}

Testing the definitions so far reveals an
interesting deficiency of the analysis as it stands: it contains an
unformalized hidden assumption of longest-match behaviour! To see
this, consider in \ref{ex:chru_aut} the automaton that
\code{closed\_interpretation(chruschtschow \&  hypocoristic)} evaluates to.
Besides the correct realization, two alternative paths starting at nodes 3 and 4 mark unwanted
realizations that truncate earlier within the medial consonant cluster 
/\ip{StS}/. These alternatives occur in the first place because (i)
sonority differences are computed over the {\em truncated} surface
form, (ii) the amount of truncation is left indeterminate by
\code{truncated\_part}, and (iii) a stem-internal $(\neg up)^n\,up$ configuration 
like our
\ling{\textscu$_{down}$\textesh$_{plateau}$t$_{plateau}$\textesh$_{up}$} 
provides $n$ possible truncation points satisfying
\code{sonority\_based\_cutoff\_point}, with no preference given to
anyone of them. Now \citeN{karttunen:96} reports an interesting
simulation of longest-match behaviour for rewrite rules using only
finite-state machinery. Unfortunately, it appears that his results are
crucially dependent on the ability of finite-state transducers to
describe two-level correspondences, precluding a transfer of his
results to the present monostral setting. Fortunately, we already have another formal device 
to express preferences, namely Bounded Local
Optimization. Observe
that the unwanted alternatives are distinguished from their grammatical
counterpart in that their length-2 subpaths $3\rightarrow 5\rightarrow 
7$, $4\rightarrow 7\rightarrow 8$ all contain a \code{skip}
-- the hallmark of truncation -- whereas the corresponding grammatical 
paths $3\rightarrow 4\rightarrow 
6$, $4\rightarrow 6\rightarrow 8$ are `better' insofar as they contain
only  proper segmental material. Therefore a weight scale $segment <
skip$ which feeds into look-ahead-2 application of Bounded Local
Optimization \code{blo} solves our problem by
effectively preferring late truncation. This leads us to the final
formulation of \code{i\_formation}: 
\startpiece
\begin{verbatim}
i_formation := 
 blo(mb(closed_interpretation(chruschtschow & hypocoristic)),2).
\end{verbatim}
\stoppiece

\noindent At this point the alert reader may start to wonder whether German
hypocoristic truncation admits a non-optimizing alternative analysis
similar in spirit to Ulwa. The short answer is yes, but with some
extra subtleties. Again, such an alternative involves lexical
constraint application, this time pertaining to the prespecification
of sonority differences in stems. 

The crucial observation leading to this is
that too-short truncations necessarily entail a conflict between the
sonority curves of isolated stems and their truncated remainder in the 
\ling{i-}suffixed word forms: the vocalic suffix provides a new right
context which affects the sonority difference of its 
consonantal left neighbour. For example, in
\ling{k$_{up}$\textinvscr$_{up}$\textscu$_{down}$\textesh$_{plateau}$t$_{plateau}$\textesh$_{up}$\textopeno$_{down}$f$_{down}$}
  the first /\textesh/ is on a sonority $plateau$ because of
  neighbouring /t/ in the stem, but will be tagged with the conflicting value $up$ when
  higher-sonority /i/ forms the new right context: 
  *\ling{k$_{up}$\textinvscr$_{up}$\textscu$_{down}$\textesh$_{plateau\pmb{\neq} up}$-i$_{down}$}.

 However, lexical tagging of sonority differences must be applied
 cautiously in our surface-true setting because of monosyllabic words
 like \ling{Hans} $\sim$ \ling{Hans-i}: since computation of
 individual sonority differences requires examination of the right
 context segment, the sonority difference of the last stem segment --
 which lacks a natural context --
 must be lexically underspecified to avoid a conflict $down\neq up$
 whenever \ling{-i} adds some further word-level context. Fortunately,
 this demand corresponds well to a modular version of
 \code{sonority\_differences}, which separates the
 \code{boundary\_condition} responsible for tagging a word's last
 segment with \code{down} from
 the \code{plain\_sonority\_differences} that take care of the
 rest. Also, we must make sure that lexical tagging does see a
 non-truncatable, skip-free representation of the stem in order to
 avoid inadvertent tagging of the set of
 all possible truncations. The last concern is adressed by
 devising a separate \code{add\_skips} regular expression operator which is
 applied on the result of lexical tagging of skip-free stem
 material. This much being said, we can now present the alternative analysis:
\startpiece
\begin{verbatim}
technical_symbols := material_is((skip;repeat)).

sonority_differences :=
  ignore(plain_sonority_differences & boundary_conditions, 
        technical_symbols).

stringToSegments([],[]).
stringToSegments([ASCII|Codes],[producer(Segment)|Segments]) :-
        name(Segment,[ASCII]),
        stringToSegments(Codes,Segments).

(stem(String) := add_repeats(contiguous(add_skips(
                 Segments & plain_sonority_differences)))) :-  
   stringToSegments(String, Segments).

lexicon := { stem("kRUStSOf"), stem("hans") }.

non_optimizing_i_formation := 
  closed_interpretation(lexicon & hypocoristic).
\end{verbatim}
\stoppiece
In order to be surface-true despite occasional occurrences of
interspersed \code{technical\_symbols}, in particular \code{skip},
\code{sonority\_differences} jumps over sequences of such symbols with
the help of the built-in \code{ignore} operator.
Note also that the \code{stem} macro makes use of the Prolog hook
facilities (\code{:- stringToSegments(...)}) to synthesize a regular expression denoting the
concatenation of $N$ producer-type segments from a more convenient
\code{String} description of length $N$, transferring the result via another
Prolog variable \code{Segments}. Finally, observe that the definition of
\code{hypocoristic} has been left unchanged, hence in particular imposing its
word-level view of the \code{sonority\_differences} to fully specify
even the last segment.

The pros and cons of the two alternative treatments of German
hypocoristic truncation are pretty much the same as in the case of
Ulwa. In particular, as indicated in the code fragment above, we can
again compute the results of hypocoristic formation over the entire
lexicon in one fell swoop. Like in Ulwa, careful lexical enrichment
was the key to achieve the non-optimizing analysis for
German. Although a full conclusion seems premature at this point, the
need for much more careful argumentation is only too \finalforcedpage apparent when it
comes to defending an all-optimization framework for phonological and
morphological analysis.\finalpagebreak

\subsection{\label{ex:tagalog_sect}Tagalog overapplying reduplication}
Tagalog, an Austronesian language with 14,850,000 first language
speakers, is the national language of the Philippines
\cite{ethnologue:96}. The language is of particular interest to
reduplication theorists because of its wide use of reduplication in
productive word formation, coupled with interesting instances of {\em overapplication} of nasal
 assimilation and coalescence processes on a par with {\em normal
  application} of an intervocalic flapping rule. While the specific Tagalog instance
was already noted by \citeN[221f]{bloomfield:33}, it was
\citeN{wilbur:73} who coined the generic term {\em overapplication} in
her pioneering thesis. There she described a class of reduplicative
processes interacting with phonological rule applications, where a
particular change to the base effected by a phonological rule is
mirrored in the reduplicant and vice versa, although only one of 
these constituents actually provides the context required by the
rule. In  derivational terms the rule is therefore said to {\em overapply} even where its
context is not met, whereas with {\em normal application} this
behaviour would be ruled out. (The converse case of {\em underapplication}
exists as well, even in Tagalog, but will not be discussed further due to lack of
space). With the advent of constraint-based theories the
mechanisms for analyzing overapplicational reduplication have of
course changed, but the terminology is still widely used for
convenience. 

For information on Tagalog reduplication in the context of various generative analyses, see
in particular \citeN{carrier:79}, \citeN{marantz:82}, \citeN[ch.4]{lieber:90}, \citeN{mccarthy.prince:95a}.
According to these sources, Tagalog has three major reduplication
patterns, termed RA, R1, R2 in \citeN{carrier:79}. 
Type RA prefixes the initial CV portion of the base, accompanied by
lengthening of the reduplicant vowel, while R1 insists on a short vowel in the
same CV reduplicant shape. Type R2 copies longer portions of the base, 
sometimes the entire stem. These three patterns function in a variety
of word formation processes, and their semantic contribution can only
be evaluated with reference to the accompanying affixes and other
aspects of morphological structure. For example, while any Tagalog verb can
undergo RA reduplication to receive aspectual marking, RA is interpreted as
causative aspect only together with the prefix \ling{na-ka-}, while
conveying future aspect in conjunction with
certain subject-topic markers such as \ling{mag-, -um-}. %
Note that Tagalog
verb forms always require at least one topic marker. The actual array of 
permissible topic markers must be lexically specified for each verbal stem.

Because RA reduplication -- due to
lengthening -- shows extra material not present  in the base, and
because it can exhibit both overapplication and normal  
application effects, it seems to have just the right amount of
`real-life' complexity  for an illustrative implementation. Hence we
will focus on this type in what follows. 

In \ref{ex:tagalog} we provide a relevant sample of the data (ST/DOT
abbreviate Subject/Direct Object Topic markers).
\begin{Ex}
\item \label{ex:tagalog} {\sc Tagalog} CV\textlengthmark{} {\sc reduplication} \\
\begin{tabular}[t]{llll}
a. &  \ip{mag-linis} \gl{ST-clean} & b. &  \ip{mag-li:-linis} \gl{ST-will clean} \\
c. &  \ip{mag-bukas} \gl{open!} & d. &  \ip{mag-bu:-bukas} \gl{will open} \\
e. &  \ip{nag-bukas} \gl{opened} & f. &  \ip{nag-bu:-bukas} \gl{is/was opening} \\[1ex]
g. &  *\ip{na-ka-Pantok} &  h. &  \ip{na-ka:-ka-Pantok} \gl{causing sleepiness}\\
i. &  \ip{Pi-pa:-pag-bilih} \gl{DOT-will sell} \\
 & \ip{Pi-pag-bi:-bilih}  & k. &  \ip{p-um-i:lit} \gl{one who compelled}\\
j. &  \ip{ma-Pi:-Pi-pag-linis}  \gl{will manage to clean for}\\
& \ip{ma-Pi-pa:-pag-linis}   & l. &  \ip{nag-pu:-p-um-i:lit} `{\it one who makes} \\
 & \ip{ma-Pi-pag-li:-linis}  & &  \phantom{\ip{nag-pu:-p-um-i:lit} `}{\it extreme effort}'
\end{tabular}
\end{Ex}
We can see in \ref{ex:tagalog}.\/a-f that RA reduplication has copied
stem material and that its meaning contribution varies as a function
of the segmental prefix. However, in \ref{ex:tagalog}.\/h-l we find that 
the same reduplicative process can also repeat affixal material,
once again underscoring the phonological character of reduplicative
copying. Most interestingly, with more than one prefix we get free
variation as to which part is reduplicated \ref{ex:tagalog}.\/i,j, with only the leftmost
affix being exempted from reduplication. To complete the case against
a putative morphological characterization of RA reduplication,
\ref{ex:tagalog}.\/l shows an example where the reduplicant freely combines stem and infix segments.%
\footnote{\citeN[221f]{bloomfield:33} cites the contrasting case of [\ip{ta:wa}]
  \gl{a laugh}, with [\ip{ta:ta:wa}] \gl{one who will laugh} turning
  into [\ip{tuma:ta:wa}] \gl{one who is laughing}. Although here it
  appears as if RA reduplication would not always copy right-adjacent
  material -- skipping over infixed \ling{-um-} in this case -- we
  will in fact assume that [\ip{ta:ta:wa}] and similar cases are lexicalized
  reduplications, i.e. new stems, to which \ling{-um-} infixation
  regularly applies. If this assumption turned out to be wrong,
  one whould have to make the RA reduplicant  discontiguous
 and condition the contrastive behaviour of pairs like
  [\ip{nag-pu:-p-um-i:lit}] vs. [\ip{t-um-a:-ta:wa}] with the help of
  suitable morphological features.}
As mentioned before, Tagalog also has an
alternation known in derivational terms as Nasal Substitution \ref{ex:nas_sub}.
\begin{Ex}
\item {\sc Nasal Substitution and Assimilation} \label{ex:nas_sub} \\
\begin{tabular}[t]{ll}
a. /\ip{maN-bilih}/ $\rightarrow$ \ip{mamilih} \gl{ST-shop} &
b. /\ip{maN-dikit}/ $\rightarrow$ \ip{manikit} \\
 & \gl{ST-get thoroughly  stuck} \\

c. /\ip{maN-basah}/ $\rightarrow$ \ip{mambasah} \gl{ST-read} &
d. /\ip{maN-dukut}/ $\rightarrow$ \ip{mandukut} \\
& \gl{ST-pick pockets} \\

e. /\ip{mag-kaN-dikit}/ $\rightarrow$ \ip{mag-kan-dikit} \\
\gl{ST-get stuck   accidently as a result of} \\
\end{tabular}
\end{Ex}
The first entries \ref{ex:nas_sub}.\/a,b show an assimilation of the place
features of the prefix-final nasal to the following stop, coupled with 
a coalescence of the two segmental positions into just one (the substitution). However,
the coalescence part of the alternation is subject to two-way
exceptions: not only do certain bases resist coalescence in the
presence of coalesceable prefixes like \ling{\ip{maN-}} 
  \ref{ex:nas_sub}.\/c,d, but also there are other prefixes like \ling{mag-\ip{kaN}-} which -- although they share the final velar nasal --  do not coalesce under
  concatenation with the right bases \ref{ex:nas_sub}\/.e. We will have
  to take care of this lexical conditioning in the analysis to follow.

The interaction of Nasal Substitution and RA reduplication now
produces the overapplication effects mentioned above \ref{ex:overapp}.
\begin{Ex}
\item {\sc Reduplication and Overapplying Nasal Substitution} \label{ex:overapp} \\
\begin{tabular}[t]{ll}
a. /\ip{paN-pu:tul}/ $\rightarrow$ & \ip{pa-mu-mu:tul} \gl{that used for cutting} \\
& *\ip{pa-mu-pu:tul} \cite[221f]{bloomfield:33}\\
b. /\ip{maN-kaPilanan}/ $\rightarrow$ & \ip{ma-Na:-NaPilanan} \gl{ST-will
  need} \\
& *\ip{ma-Na:-kaPilanan} \\
c. /\ip{maN-pulah}/ $\rightarrow$ & \ip{ma-mu:-mulah} \gl{will turn red} \\
& *\ip{ma-mu:-pulah} 
\end{tabular}
\end{Ex}
Although only the leftmost instance of the triggering plosive is local 
to the prefix-final nasal, its second occurrence must be realized as a 
place-assimilated nasal as well: nasal substitution overapplies.
To complicate matters, however, it must be
noted that this kind of long-distance dependency between base and reduplicant segments 
is restricted to certain segmental classes: like in the rest of morphology
\ref{ex:flap}.\/a,b, two occurrences of dental stops created by
one of the reduplication patterns can become dissimilated through intervocalic flapping, a
{\em normally} applying phonological `rule' \ref{ex:flap}.\/c-e.
\begin{Ex}
\item d $\sim$ \ip{R}: {\sc  Normal Application of Intervocalic Flapping} \label{ex:flap} \\
\begin{tabular}[t]{l@{}l}
a. \ip{da:mot} \gl{stinginess} & b. \ip{ma-Ra:mot} \gl{stingy} \\
c. \ip{man-da:-RamboN} \gl{bandit} & d. \ip{sunud-sunuR-in} \gl{no gloss} \\
e. \ip{d-um-a:-RatiN-datiN} \gl{attends now and then}
\end{tabular}
\end{Ex}
With this last piece of evidence on Tagalog reduplication we
have now assembled enough data and generalizations to proceed to the
analysis itself.  

Because our aim is to model RA reduplication, let us start with
various macro specifications which help implement the necessary
synchronisation of morpheme edges. The idea here is that all
non-floating, i.e. more or less concatenative morphemes have their left edge synchronized (type
\code{synced}) and their interior material unsynchronized
(\code{$\sim$synced}), while only the right edge of the stem is
synchronized as well; bound morphemes remain unsynchronized. Floating morphemes
like Tagalog's famous \ling{-um-} would be special in that they would
be underspecified for synchronization, to be compatible with whatever
landing site their prosodic conditions demand; however, we disregard
the additional complexity for the fragment under development. Here is
the code portion for synchronisation (applications of it will appear in later code snippets):
\startpiece
\begin{verbatim}
synced_position(Spec) := consumer(synced & Spec).
synced_position := synced_position(anything).
synced_producer(Spec) := producer(synced & Spec).

unsynced_position(Spec) := consumer(~ synced).
unsynced_position := unsynced_position(anything).

unsynced_portion := [unsynced_position *].
left_synced_portion := [synced_position, unsynced_portion].

synced_constituent := [left_synced_portion, synced_position].
\end{verbatim}
\stoppiece
Next we need to turn to the representational needs of stems. They
should of course be reduplicatable, internally discontiguous%
\footnote{Note that \code{internally\_discontiguous} is a variant of the
  \code{discontiguous} operator introduced in earlier analyses that
  spares both start and final states from the self-loop enrichment. It 
  follows that we can safely intersect such a representation with additional
  constraints like our \code{synchronized\_constituent} without fear
  of inadvertent constraining effects on peripheral material introduced outside of the
  morphemic domain under construction. In particular, here we want the 
  edges of a {\em morpheme} to be synchronized, not the edges of the whole
  word. Upon wrapping the result with
  the previously defined \code{contiguous} operator, which
  introduces self loops to the start and final states only, we then have an
  appropriately conditioned partial description of an entire word
  with the required tolerance of other morphemes that may be present.}
 (for later tolerance of
floating morphemes), and synchronized at both edges. While
non-stop-initial stems like \ling{linis} need only basic segmental
specifications in addition to these needs to complete their definition,
stems like \ling{pulah, dikit, ka\textglotstop ilanan} require
additional means to implement the alternating behaviour of their
initial plosives. Recall that in Declarative Phonology destructive
processes like the one hinted at by `nasal {\em substitution}' are impossible
to represent literally, hence must give way to alternative representational treatments. The 
solution here is to underspecify the manner features signalling
obstruenthood and (non-)voicing in one alternant of the initial
segment's specification, with a fully specified stop constituting the
other alternant. In a cooperative fashion, coalesceable \ip{N}-final prefixes will then supply
the missing nasal manner feature to guarantee full specification after
the intersection of all participating morphemes-as-constraints has
been done. Finally, to  model the default stand-alone realization as a 
plain stop, the fully specified alternant is encoded as a producer,
whereas the `cooperative', underspecified alternative is specified as a consumer. With
these definitions in place, a wide range of stems can now receive their
representations with ease:
\startpiece
\begin{verbatim}
discontiguous_stem(Segments) :=
  add_repeats(contiguous(internally_discontiguous(Segments) & 
  synced_constituent)).
                                              
underspecified(BaseSpec,AddedForFullSpec) :=
                    { producer(BaseSpec & AddedForFullSpec), 
                      consumer(BaseSpec) }.

bilih :=
 discontiguous_stem([underspecified(labial,obstruent & voiced), 
 stringToSegments("ilih")]).

pulah := discontiguous_stem(
        [underspecified(labial,obstruent & ~ voiced), 
 stringToSegments("ulah")]).

dikit := discontiguous_stem(
        [underspecified(dental,obstruent & voiced), 
  stringToSegments("ikit")]).

kaqilanan := discontiguous_stem(
             [underspecified(dorsal,obstruent & ~ voiced), 
  stringToSegments("aqilanan")]).

basah := discontiguous_stem(stringToSegments("basah")).

dukut := discontiguous_stem(stringToSegments("dukut")).

guloh := discontiguous_stem(stringToSegments("guloh")).

laakad := discontiguous_stem(stringToSegments("laakad")).

linis := discontiguous_stem(stringToSegments("linis")).

dambong := discontiguous_stem([producer(voiced & dental), 
                  stringToSegments("amboN")]).
\end{verbatim}
\stoppiece
With the previous remarks on underspecification as a suitable strategy 
for (many) seemingly destructive alternations, the reader will have no 
difficulty to understand the definition for \code{dambong}, not mentioned before;
here again missing manner features will be supplied from later constraints to 
guarantee full specification of its initial voiced dental, fleshing it 
out either as a stop or as a flap.

With the definitions for stems in place, let us concentrate next on interesting
prefixes, the foremost of which are the \ip{n}-final ones. They
present a good case for \citeN{ellison:93}'s claim that intersective
morpheme combination is often to be preferred over concatenative one,
since part of their definition is the nasalhood imputed on suitable
stem-initial consonants (\code{producer(nasal)}), i.e., an
underspecified contextual restriction that must be realized outside of their own
morphemic interval.  Because they are also ordinary segmental
prefixes, however, these morphemes are specified as
\code{contiguous} \code{synced\_constituent}s. What makes them special 
is that -- even when coalescence of nasal and stop features is not
possible, like in \ling{\ip{mam-basah}}, *\ling{ma-masah} -- the nasal
  must assimilate in place to the following obstruental stop, if any (otherwise
  receiving a default place, namely dorsal); hence the other
  prefix-final alternant labelled
  \code{assimilated\_nasal\_obstruent\_sequence}.%
\footnote{A subtle detail needs our attention here: the conjunction of
  \code{synced\_constituent} with the two-way alternating segmental
  specification correctly marks the rightmost segment as
  \code{synced}, irrespective of the ultimate length of the prefix
  (length 2 for the coalescent case, length 3 for the pure
  assimilation case). This rightmost segment is nothing else but the
  beginning of the stem, hence correct intermorphemic alignment can be ensured with the
  help of one additional constraint (to be detailed later) which
  governs the correct distribution of synchronisation marks.}
A non-coalescing prefix like \code{magkang} differs minimally in dispensing with the
underspecified nasal alternant, but retains the assimilation part, as
demanded by the data in \ref{ex:nas_sub}:
\startpiece
\begin{verbatim}
ng_final_prefix(String, SpecifiedAlternant) :=
      contiguous(synced_constituent &
                [stringToSegments(String),
                 { SpecifiedAlternant, 
                   assimilated_nasal_obstruent_sequence }]).

coalescent_nasal := producer(nasal).
assimilation_only := {}.  % empty language = discard alternant!

mang := ng_final_prefix("ma", coalescent_nasal).

pang := ng_final_prefix("pa", coalescent_nasal).

magkang := ng_final_prefix("magka", assimilation_only).
\end{verbatim}
\stoppiece
The place-assimilating behaviour itself must be simulated in a piecewise
fashion for all the possible place categories involved
(\code{place([labial,dental,dorsal])}), because of the well-known
lack of token identity in regular description languages. We can regain
some expressivity, though, by making use of a recursive macro
\code{assimilation\_for} to off-line-synthesize the iterated
disjunction from the list of places of articulation. The trick is to
exploit the fact that Prolog is the host language of our finite-state
toolkit, which means that we can use token identity at the description level, though not at 
the object level. Employing the power of logical variables we therefore distribute the variable
\code{SharedCategory} to each disjunct (which will of course become
bound at compile time).  

Putting together the pieces in \code{assimilated\_nasal\_obstruent\_sequence}
again requires a judicious use of producer- and consumer-type
information to distinguish the lexical contribution of the morpheme
itself -- which is the nasal part -- from the contextual requirement
participating in assimilation -- which is the obstruent part. Finally, 
a \code{default} disjunct encodes the dorsal realization that is
observed whenever a suitable obstruent stop is lacking:
\startpiece
\begin{verbatim}
place([labial,dental,dorsal]). % Prolog fact holding 
                                          % list of categories

(place_assimilation := assimilation_for(Place)) :- place(Place).

assimilation_for([]) := '{}'.
assimilation_for([SharedCategory|Categories]) :=
  { [consumer(SharedCategory), consumer(SharedCategory)], 
    assimilation_for(Categories) }.

assimilated_nasal_obstruent_sequence :=
{ [producer(nasal), consumer(obstruent)] & place_assimilation,
  default }.

default := [producer(nasal & dorsal),consumer(~ obstruent)].
\end{verbatim}
\stoppiece
Before we proceed to add the reduplicative part to our growing Tagalog 
fragment, let us test the definitions so far to obtain some highly
instructive intermediate results. With a little additional code shown
below to ensure wellformed synchronization, we can in fact form
meaningful intersections like \code{word(mang \&  bilih)}.\finalforcedpage
\startpiece
\begin{verbatim}
matching_synchronisation := 
  not_contains2(two_synced_segments_in_a_row).
two_synced_segments_in_a_row := 
  [consumer(synced),consumer(synced)].

generic_word_constraints := matching_synchronisation.
word(Expr) := closed_interpretation(Expr & generic_word_constraints).
\end{verbatim}
\stoppiece\finalpagebreak

\noindent However, the surprising result is that -- with coalesceable stems and
alternating \ip{N}-final prefixes -- we actually get two results
\ref{ex:two_res}.
\begin{Ex}
\item \label{ex:two_res} /\ip{maN}-bilih/: {\sc Nasal Coalescence Problem} \\
\epsfig{file=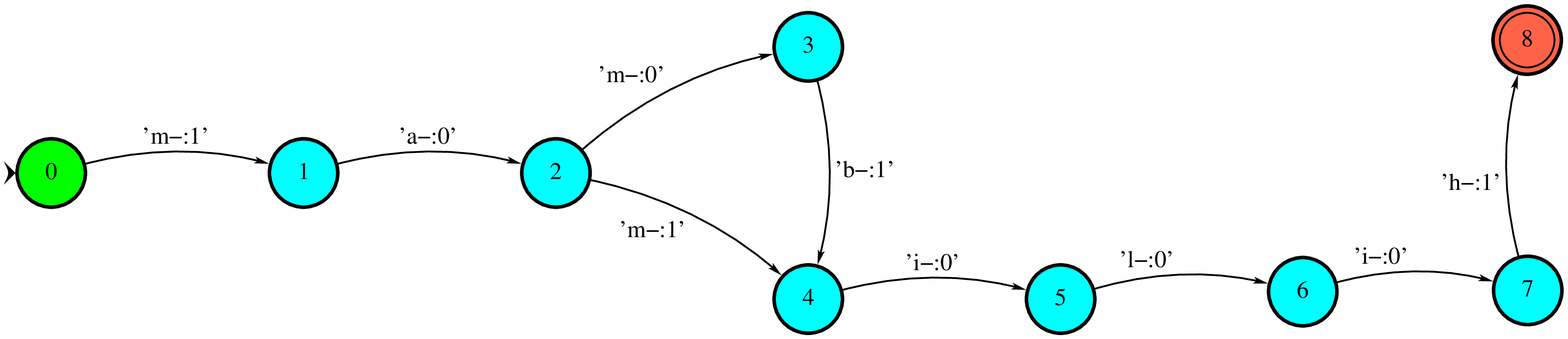,width=1.05\linewidth}
\end{Ex}
Besides the desired coalescing alternant \ling{m} ($2 \rightarrow 4$) we get an unwanted
alternative \ling{mb} ($2\rightarrow 3 \rightarrow 4$) that does not merge the two
segmental positions. Some reflection reveals that this behaviour is
unavoidable given our modelling assumptions about independent and
compositional specification of morphemes. Since the prefix must
combine with both the nasal-substituting \ling{bilih} and the
non-substituting \ling{basah}, the sequence \ling{mb} cannot be
ruled out categorically, hence both alternants must remain in the
denotation of the affix. Likewise, an alternating stem like
\ling{bilih} needs to tolerate non-nasal realizations of its first
segment in addition to the coalescent nasal case, to cater for isolated
pronunciation or other prefixal options. If we devise an abstract
feature [$\pm$ coalesce] to characterize both stems and prefixes, and
relate the alternating case to an underspecified feature value [0 coalesce], we
can illustrate the scenario with the following paradigm \ref{ex:para}.
\begin{Ex}
\item \label{ex:para}{\sc Prefix-stem Combination Paradigm}\\
\begin{tabular}[t]{|c||c|c|}
\hline {\em Prefix} $\rightarrow$ & [-- coalesce] & [0 coalesce] \\
{\em Stem} $\downarrow$ & & \\ \hline\hline
[-- coalesce] & magkam-basah & mam-basah \\
                     & [-- coalesce] &  [-- coalesce] \\ \hline
[0 coalesce] & magkam-bilih & ma-milih \\
                   & [-- coalesce]    & \bf [+ coalesce] \\ \hline
\end{tabular}
\end{Ex}
The paradigm reveals that we actually demand full specification from the
(intersective) combination of two underspecified 
features \mbox{[0 coalesce]}, an impossibility in a monotonic setting where
underspecification is equivalent to a (systematic) disjunction of fully specified disjuncts.
This state of affairs has been noticed before: replacing `coalesce' with $\delta$, the
featural part of the paradigm turns out to be a verbatim copy of the one discussed
in \cite[p.31, ex. (15)]{ellison:94a} under the general rubric ``paradigm[s that] might be 
decomposable into morphemes only with the use of defaults.''! With 
Tagalog Nasal Substitution we have merely uncovered a first concrete
instance of the abstract case predicted by Ellison. The default in our
case would be \mbox{[+ coalesce]}. 

Fortunately, optimization again comes to our rescue to implement the
required default behaviour. Note that the default preference for
coalescence amounts to an eager choice of synchronized morpheme
beginnings in our representational setting. This is  illustrated well in the
critical choice between unsychronized $2 \stackrel{m:0}{\rightarrow}
3$  and synchronized $2 \stackrel{m:1}{\rightarrow}4$ in
\ref{ex:two_res}. Therefore, Bounded Local Optimization parametrized
with look-ahead 1 and a weight scale $\neg synced > synced$
completely solves our little problem.

With Nasal Substitution handled on its own, let us move on towards a
declarative formalization of type RA reduplication with overapplication.
Because we again see
reduplication as partial specification of an entire word, we construct 
it as follows. Before the copy we need to expect at
least one (prefixal) morpheme, possibly more \ref{ex:tagalog}. After it we must
allow for the rest of the base and also tolerate whatever morphemic material
finishes off the word (especially in the case where one of the
prefixes is targeted for reduplication and the stem must still be incorporated).
Within the reduplicant-base compound we need to enforce a kind of long-distance
agreement between the quality of the first reduplicated segment and
its base counterpart to model `overapplication'. However, as we have
seen in the discussion of flapping, in a surface-true perspective this kind 
of agreement must be carefully constructed to hold only over certain
subcategories of the segmental makeup and spare the dimensions
involved in (non-)flapping.

On the face of it, the \code{type\_RA\_reduplicant} itself copies the first two base
segments -- which happen to form a CV sequence -- and lengthens the
V. To do so formally, we can identify the initial base segment as a
\code{synced\_position}, hence its successor is of course an
\code{unsynced\_position}. For lengthening of the vowel, we have chosen a
formalization which makes use of the \code{repeat} arcs to step back
one position and then proceed forwards again by providing for an
abstract \code{vowel} position, which can only be the vowel we have
targeted already before. After that, a kleene-plus-wrapped
\code{producer(repeat)} steps all the way back to the beginning of the 
base, which again is characterizable as a synchronized
segment. However, to make sure that consumer-type underspecified
alternants as contained in the first segment of \ling{bilih} will survive
\code{closed\_interpretation}, we have been careful here to specify
this beginning of the base proper with a matching
\code{synced\_producer}. Like in the nasal assimilation instance, the 
segmental positions agreeing in an overapplicational manner are again
abstracted out via the logical variable
\code{AgreeingFeatures}. Enforcing the agreement in a reasonably
perspicuous way is done via the Prolog hook mechanism of FSA Utilities; the respective predicate
\code{enforce\_agreement\_in\_/3} constructs an iterated
disjunction abbreviating the repeated realization of the abstract reduplicant
parametrized for each of the five values of the overapplicational
categories. 

This extensionalization of agreement is indeed the price to pay for a framework that does not
handle (possibly overridable) true token identity at the object level, as demonstrated by
\cite{beesley:98}. That paper is relevant here because it contains a
careful study of the problem of long-distance morphosyntactic dependencies in a
finite-state framework together with potential remedies; one could
indeed apply some of the solutions proposed there to our present task
of modelling nonlocal {\em phonological} dependencies. In  
particular, the use of weakly non-finite-state enhancements like global 
registers that can be set (for initializing an agreeing feature) and
tested against (an existing feature value) should be a profitable
amendment, once residual problems with automaton transformations like
minimization in the face of enhanced automata have been settled. 

Here then is the central code portion responsible for Tagalog RA reduplication:
\startpiece
\begin{verbatim}
some_morpheme := left_synced_portion.
pre_base_morphemes := [some_morpheme +].
rest_of_base := unsynced_portion.
rest_of_word := [some_morpheme *, synced_position].

( ra_reduplicated_word :=  [pre_base_morphemes, Reduplicant,
                  rest_of_base, rest_of_word]) :-
        enforce_agreement_in_(type_RA_reduplicant,
         [  nasal,  % 1: /m,n,N/
           ~ nasal & ~ voiced & labial,   % 2: e.g. /p/
           ~ nasal & ~ voiced & ~ labial, % 3: e.g. /t/
           ~ nasal & voiced & labial,  % 4: e.g. /b/
           ~ nasal & voiced & ~ labial], % 5: e.g. /d/
        Reduplicant).

type_RA_reduplicant(AgreeingFeatures) :=
 [ synced_position(AgreeingFeatures),
   unsynced_position,producer(repeat),
   unsynced_position(vowel), producer(repeat) +,
   synced_producer(AgreeingFeatures) ].

enforce_agreement_in_(_MacroName, [],  '{}').
enforce_agreement_in_(MacroName, [Cat|Cats], 
              { MacroInstance, MoreMacroInstances }) :-
 MacroInstance =.. [MacroName,Cat],
 enforce_agreement_in_(MacroName, Cats, MoreMacroInstances).

optimal_word(Expr) := 
  blo(mb(closed_interpretation(Expr & word)),1).
\end{verbatim}
\stoppiece
We are now in a position to show in \ref{ex:mamiimilih} the actual outcome of an automaton
specification such as \code{optimal\_word(mang \& bilih \& ra\_reduplicated\_word)}
\begin{Ex}
\item \label{ex:mamiimilih} /\ip{maN}-RA-bilih/ $\rightarrow$ [\ip{mami:milih}]: {\sc
      RA, overapplying} \\
\epsfig{file=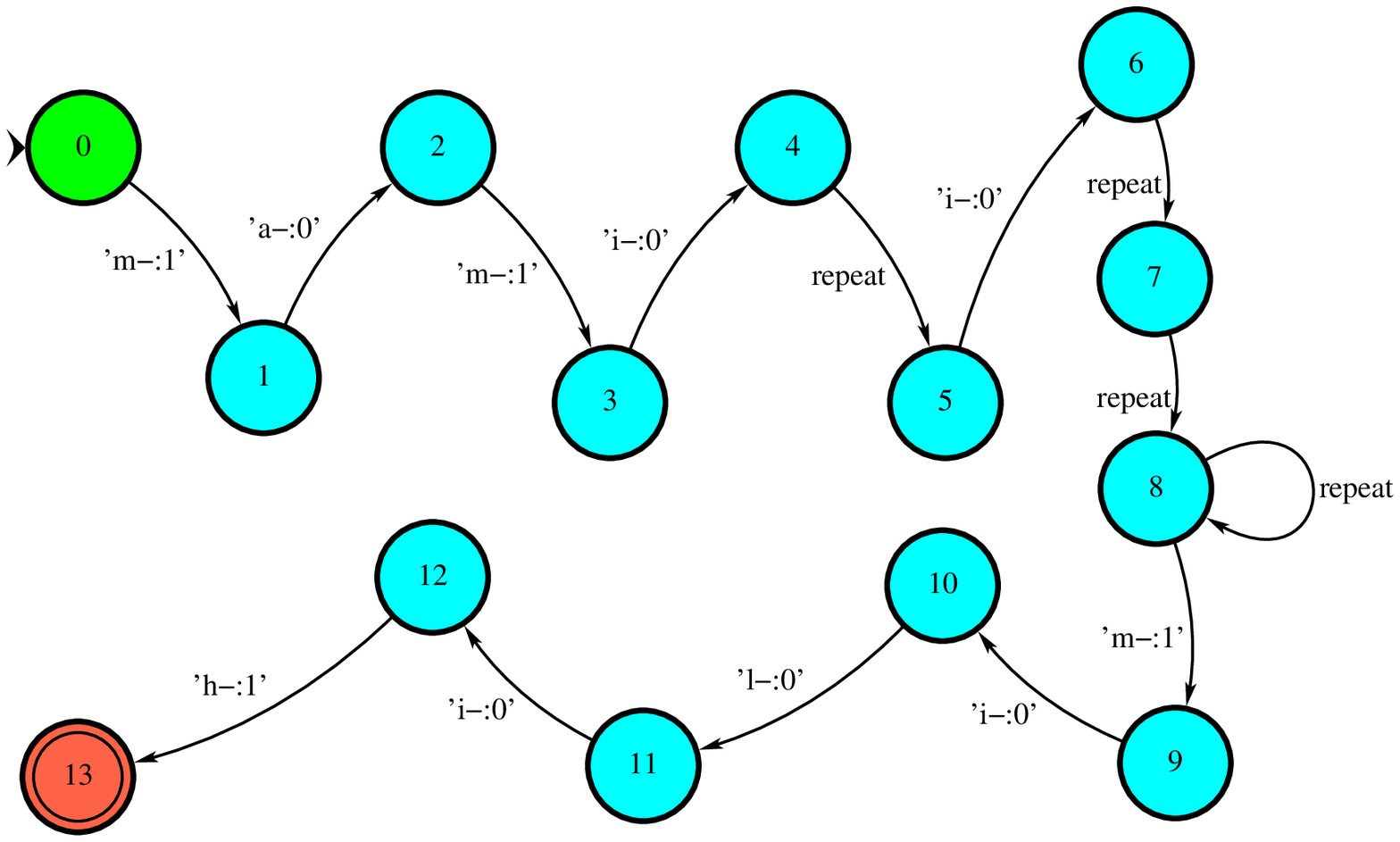,width=.7\linewidth}
\end{Ex}
With the addition of some ordinary \code{prefix}es we can even model the free
variation in reduplication arising from the presence of multiple
prefixes. For ease of exposition these will be
straightforwardly concatenated \code{[mag,qi,pag,\dots]} rather
than being intersected. The intersective modelling variant would make use of
the method proposed in \citeN{ellison:93} to simulate
concatenation with the help of suitable tagged representations.
\startpiece
\begin{verbatim}
prefix(String) :=
  add_repeats([stringToSegments(String) & some_morpheme, 
  material_is(anything)]).

qi := prefix("qi"). % `q' denotes glottal stop
pag := prefix("pag").
ma := prefix("ma").
\end{verbatim}
\stoppiece
The result of \code{word([ma,qi,pag,linis] \& ra\_reduplicated\_word)}
is depicted in \ref{ex:freevar}.
\begin{Ex}
\item \label{ex:freevar} {\sc Free variation in RA reduplication} \\
\epsfig{file=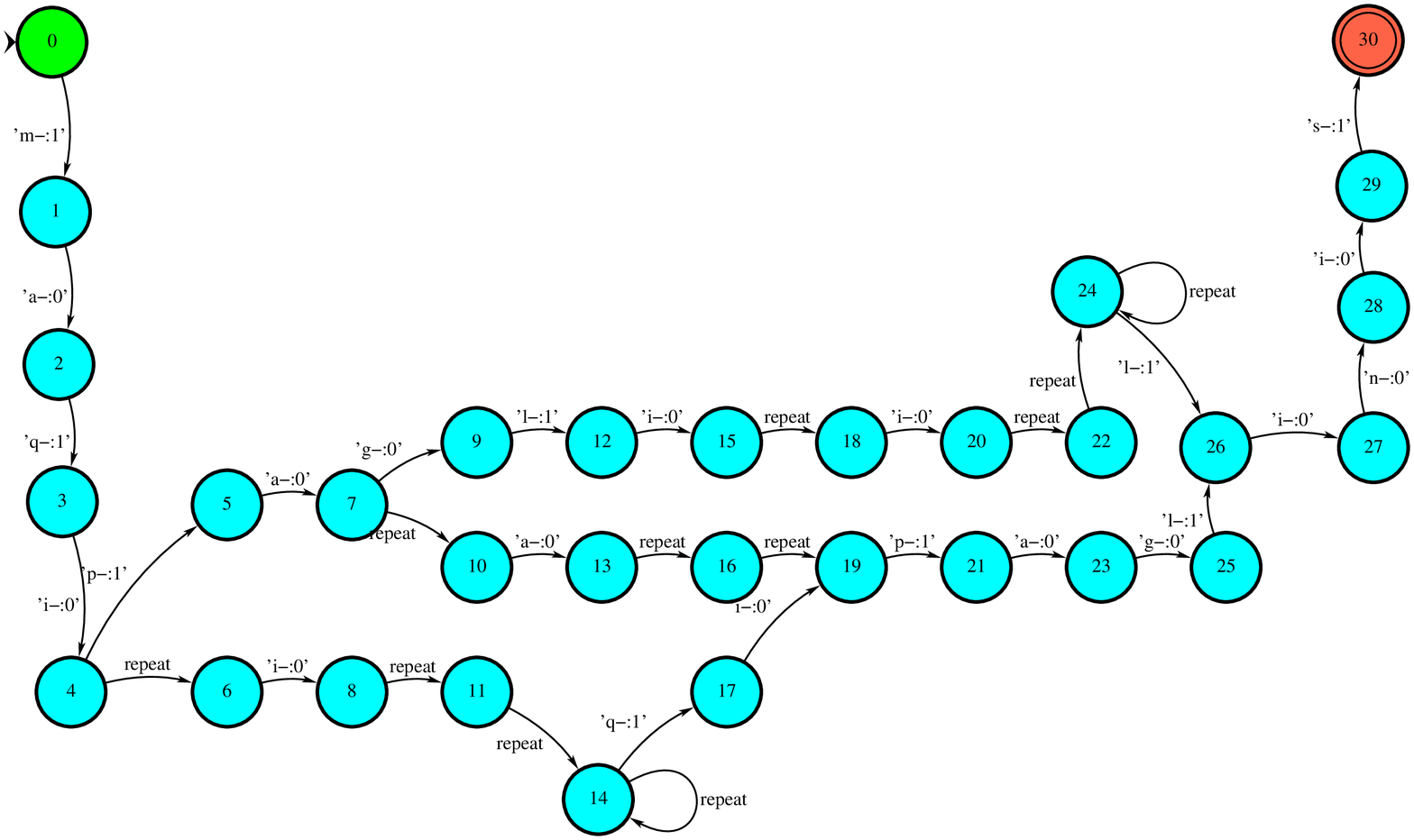,width=.99\linewidth} %
\end{Ex}
It turns out that we need to do something else to preserve this result 
under BLO application (\code{optimal\_word(\dots)}). The reason is
that we will otherwise loose one of the alternants due to a
nonsensical comparison to a repeat-initated alternative path, no matter what
weight we assign to repeat:  it cooccurs with both the larger ($4
\rightarrow 5$) and the smaller weight ($7 \rightarrow 9$) in the
automaton of \ref{ex:freevar}. This then is a perfect case for
applying the slight modification to weight computation proposed on
page \pageref{neg_weights},
fn. \ref{neg_weights}: because we want repeat arcs to behave as
inert with respect to BLO, we assign them infinite {\em negative} weight, thereby exempting them
from ever getting pruned. Remember that this welcome result follows because the essence
of the modification is that weight-sum comparison now only looks for
better {\em positive} alternatives.  

The final piece of the analysis handles the normal application of the
flapping `rule': only intervocalic voiced dentals are flapped
[\ip{R}], elsewhere they are pronounced as a stop [d]. In the context
of reduplication the rule applies uniformly to both base and
reduplicant, which translates into simple intersection of a
surface-true constraint \code{flap\_distribution} with the  
word. Here it is actually helpful that our framework is limited to
type identity, since the repetition of previous material will not copy
particular allophonic choices in either base or reduplicant, allowing the
flap constraint to rule freely. This constraint is defined below:
\startpiece
\begin{verbatim}
nonflapped_intervocalic_dental_stop := 
  [consumer(vowel), consumer(d), consumer(vowel)].

preCflap := [consumer(flap), consumer(~ vowel & segment)]. 
postCflap := [consumer(~ vowel & segment), consumer(flap)].

no_peripheral_occurence_of(Segment) := 
    ([consumer(segment & ~ Segment),  [material_is(segment),
      consumer(segment & ~ Segment)] ^] ^).

ignore_intervening_technical_symbols(Expr) := 
    ignore(Expr,material_is(misc)).

flap_distribution := 
      ignore_intervening_technical_symbols(
       not_contains3(nonflapped_intervocalic_dental_stop) &
       not_contains2(preCflap) &
       not_contains2(postCflap) &
       no_peripheral_occurence_of(flap)).
\end{verbatim}
\stoppiece
Observe that the constraint is build up (in an admittedly somewhat
roundabout way) from negative subconstraints banning intuitively
sensible subconfigurations where a flap must not occur. This is done by way of
parametrized constraints \code{not\_contains2/3} that ban occurrences
of its length-2/3 string arguments (details suppressed for the sake of 
readability). After that has been done, the resulting automaton is
modified to \code{ignore\_intervening\_technical\_symbols}, which will 
occur especially in reduplication.

With flap distribution in place, we have removed the last barrier to a 
fairly complete account of type RA reduplication, as we can now
demonstrate even the correct outcome of 
\begin{quote}\code{optimal\_word(mang \&
  dambong \& ra\_reduplicated\_word \& flap\_distribution)},\end{quote} which in
compact regular expression notation is
\begin{quote}\code{[m,a,n,d,a,repeat,a,repeat,repeat,repeat *,
  'D',a,m,b,o,'N']}.\end{quote} Note both the nonflapped \code{d} that 
starts off the reduplicant after the consonant-final prefix and the
flapping of \code{'D'} in the base, which correctly happens even
across intervening \code{repeat} symbols, because vowels 
\code{a\underline{\phantom{X}}a} form its ultimate segmental context.
\section{Discussion}\label{disc}
As we have seen in previous sections, the one-level approach presented
in this paper offers solutions to a wide range of analytical problems
that arise in modelling prosodic phenomena in morphology. Enriched
representations allow for repetition, skipping and insertion of
segmental material. The non-finite-state operation of reduplicative
copying is mapped to automata intersection, an operation whose formal
power is also beyond finite-state, but which is independently needed
as the most basic mechanism to combine constraints. The distinction
between contextual requirements and genuine lexical contribution is
reflected in differentiating between consumers and producers of
information, a move that introduces resource consciousness into
automata. Bounded local optimization supports the formulation of
optimizing analyses that seem to be called for in a range of empirical 
cases. A generic recipe for flattening prosodic constituency provides
the basis for maximally local access to higher phonological
structure. The architecture  supports both generation and
parsing. Finally, the proposal proved its worth in three practical, computer-implemented
case studies.

As is to be expected with any new proposal, however, there are
areas that deserve further investigation, limitations that need to be
overcome and alternative design decisions that should be explored.

First, the repeat-arc enrichment that facilitates reduplication assumed
that base strings are separately enriched, and not their union as a
whole, because only then can inadvertent repetition of parts of some other base be excluded.
Somewhat loosely speaking we have relied on the fact that the weaker
notion of type identity available in finite-state networks can still
mimick token identity if there is only one type to consider,  i.e.\/
in moving back and forth in time one sees only one base string at a
time. A drawback of this mode of enrichment is that the set of base strings
cannot be compressed much further, as would be possible in other cases 
of finite-state modelling. However, recall that the repeat-arc encoding imposes a highly regular
layer onto existing base lexicons. What is regular is also
predictable, hence can be virtualized: one could keep a minimized base
lexicon and devise an on-the-fly implementation of the
\code{add\_repeats} regular expression operator to add repeat arcs on
demand only to the currently pursued base hypothesis. With suitable diacritics one could even extend such an
approach to languages where exceptions that fail to reduplicate must
be taken into account. A variant of that approach would add 
global register-manipulating instructions to choice arcs in the minimized
base lexicon, e.g.\/ \citeN{beesley:98}'s \code{unify-test}. Injecting 
a little bit of memory into automata in this way would serve to
synchronize choices in the base and reduplicant part of the word form, 
acting as a kind of distributed disjunction. Because the same
instruction would be used twice in repetition-as-intersection, the use 
of a (limited kind of) unification is indeed necessary here. It
remains to be seen whether such a mixed approach would be feasible in practice.

Also, from the point of view of competence-based grammar, the proposed
encoding for reduplication is not as restricted as one might wish,
because it can also encode unattested types of reduplication such as
mirror image copying $ww^R$, a context-free language. Here is the
essential part of an actual piece of code showing how to do it:
\startpiece
\begin{verbatim}
realize_base :=  [boundarySymbol,interior_part,boundarySymbol].
mirror_image_total_reduplication :=   
  [realize_base, [repeat,repeat,contentSymbol] *,
   repeat, repeat, boundarySymbol, skip *].
\end{verbatim}
\stoppiece
When intersecting with an enriched version of a string like
\code[a,b,c,d] (bracketed with \code{boundarySymbol}), each character of the mirror-image part
\code{[d,c,b,a]} will be preceded by two \code{repeat} symbols. Moreover, because we
will ultimately have stepped back to the beginning of the string, we
finally need to  \code{skip} over the entire base to avoid the
realization of an additional conventional reduplicant. This additional
effort suggest one way to explain the absence of mirror-image
reduplication, namely by extending the general notion of resource
consciousness: traversing an arc carries a certain cost and there 
will be a maximum cost threshold  in production and acquisition. 
Under this view mirror-image reduplication fares significantly worse
than its naturally occuring competitor because it needs three times the amount of technical
arcs. Clearly, this sketch of a performance-based explanation of an important
gap in natural language patterns of reduplication needs to be worked
out more fully in future research, but the initial line of attack seems promising.

Another point worth noting is that the repeat-arc encoding is not
intimately tied to one-level automata. This is to be expected given
the fact that transducer composition \code{o} can simulate intersection \code{A 
  \& B} via \code{identity(A) o identity(B)}. As demonstrated in appendix
\ref{transducer-redup}, a version of our proposal that works for transducers is readily
constructed for those who prefer to work in a multilevel framework.

Finally, one might worry about the apparent need for a greater number
of online operations, especially runtime intersection. The radical
alternative, relying completely on offline computations, seems rather
unattractive in terms of storage cost and account of productivity for
cases like reduplication in Bambara, Indonesian, etc. However, we believe that with
lazy, on-the-fly algorithms for intersection etc.\/ the runtime cost
can still be kept at a moderate level. This is an interesting area for
further experimentation, especially when representative benchmarking procedures can 
be found.

There is also room for variation with respect to the ideas about resource-conscious automata.
First, instead of distinguishing between producers and consumers on the
level of entire segmental symbol, one could employ a finer subdivision 
into feature-level producers and consumers. So e.g.\/ a segment might
produce a nasal feature but consume place-of-articulation
features delivered from elsewhere. 

Second, one might play with
changing the logic of producer/consumer combination. For example, in a 
stricter setting corresponding to linear logic, resources could be
consumed only once and leftover unconsumed resources would result in
ungrammaticality. One potential application area might be to prevent coalescent
overlap, signalled by a doubly produced resource at the position of
coalescence. However, the intuitionistic behaviour embodied in the
present proposal seems to fit more naturally with the demands of
productive reduplication, where base segments are consumed two or
even more times. In contrast, linear behaviour would require explicit re-production 
of consumed resources here. More analyses need to be undertaken to
find out whether both types of behaviour (or even more) are needed to
formulate elegant grammars of a wider range of phenomena, or whether a 
single uniform combinatory logic suffices. 

Finally, there is a possibility that resource consciousness of the kind proposed here may
be reducible to classical arc intersection via global grammar
analysis/transformation.  This is not alltogether implausible given the fact 
that LFG's bipartite setup of constraint and
constraining equations is now seen as involving resource-conscious notions,
while at the same time a theoretical reduction of LFG grammars to simpler ones
employing only conventional constraint equations (i.e.\/ pure unification grammars) exists.% 
\footnote{This remarkable result is attributed to unpublished work by Mark Johnson 
  in \citeN[19]{shieber:87}. Mark Johnson (e.c.\/) explains that it involves a
somewhat complicated grammar transform.}
To be sure, even if such a reduction would prove to be feasible in our 
case, the value of resource-consciousness as a high level concept
would still be unaffected.

For Bounded Local Optimization, a brief note will do, namely that its local search for
minimal-weight arcs can profitably be integrated with the elimination of
consumer-only arcs in phase II of grammatical evaluation. As a result, we
now need only one pass over (a fraction of) the automaton, with a
subsequent gain in efficiency.

In conclusion, while there are certainly areas in need of future research, 
the one-level approach to prosodic morphology presented in this paper
already offers an attractive way of extending finite-state techniques to difficult
phenomena that hitherto resisted elegant computational analyses.
\phantom{\cite{roa}\cite{cla}}
{\small

}
\clearpage
\appendix
\section{\label{app:icopy}Alternative method:\\ Lexical transducer plus intelligent copy module}
Finite-state {\bf transducers} are currently the most popular
implementation device in application-oriented computational
morphology, mainly for reasons of low time complexity, reversibility
for generation and parsing and the existence of minimization methods.
They are used to model both the basic lexicon and its additional
morphological and phonological variation. However, to date there
exists no proposal that can handle reduplicative morphology in general,
apart from isolated attempts at simpler, rather local instances (cf.\/ a Tagalog case modelled by
\citeNP{antworth:90}).
In view of the fact that the copy language $ww$ is context sensitive,
some compromise or approximation must obviously be found, at least when the
finite-state assumption is to be maintained. The following ideas form
part of such an {\bf approximative solution}. 

The first part of the solution separates the copying aspect from
finite-state-based lexicon and underlying-to-surface variation.
\begin{Ex}
\item {\sc Separate copy}\\[1ex]
\label{SepCopy}
 $\rightarrow$\fbox{Lexicon + Phonology/Morphology
  (finite-state)}$\rightarrow$\fbox{\phantom{L}copy\phantom{L}}$\rightarrow$ Surface form 
\end{Ex}
To preserve regularity, the copy must be made outside of the lexicon + 
phonology FSA/FST. Since there are languages where
phonological rules apply to a reduplicated form, the copy should not be
done before the phonology. Also, while conceptually lexicon and
phonology may be separated, in practice they are normally
composed to form a single lexical transducer \cite{karttunen:94}. The key advantage of
this compilation step is that a possibly exponential grow of the rule
transducer is prevented in practice through the limiting contexts provided by the
lexicon. However, having only a single lexical transducer means that an intermediate
stage  is no longer available for copying before applying the
phonology.

Having secured the proper place of a copy module  {\em after} the
lexical FST or FSA, we next need to devise a way to control
copying. Since we cannot model the nested nonlocal dependencies
exhibited by $ww$-type reduplicative constructions directly, let us
instead {\em encode a  segment-local promise to reduplicate}, to be fulfilled by
the copy module. In \ref{techtelmechtel} we see a first example from German.
\begin{Ex}
\item {\sc Segment-local encoding of multiple realisations}\\
\label{techtelmechtel}
\begin{tabular}[t]{|l|c|} \hline
{\em surface} & techtelmechtel \\ \hline
{\em encoding} & \begin{tabular}[t]{l|lllllllllllllllllll}
                      {\em segment string }  & t & e & c & h & t & e & l & m \\ \hline
                       {\em \# copies} & 1 & 2 & 2 & 2 & 2 & 2 & 2 & -1
                    \end{tabular} \\ \hline
\end{tabular}
\end{Ex}
In a nutshell, the copy module now gets complex symbols, which consist 
of a segment proper and an annotation for the number of copies to be
made. The module scans a string of complex symbols from left to right, 
outputting each segment that has a nonzero number of copies. By convention,
whenever the number of copies is negative, a new scan is started after
outputting the current segment. The new scan performs the same actions 
as before, except that it first decrements the value of the
number of copies when positive while incrementing when negative. Thus, in our example we have:
\begin{examples}
\item {\sc Stepwise derivation of example \ref{techtelmechtel}} \\
\label{stepwise}
\begin{tabular}[t]{|l|l|llllllll|l|} \hline
Scan & \multicolumn{9}{c}{Input} & Output \\ \hline
1 & {\em segment string }  & t & e & c & h & t & e & l & m & {\em techtelm}\\ 
 & {\em \# copies} & 1 & 2 & 2 & 2 & 2 & 2 & 2 & -1 & {} \\[1ex] \hline
2 & {\em segment string }  & t & e & c & h & t & e & l & m & {\em echtel}\\
& {\em \# copies} & 0 & 1 & 1 & 1 & 1 & 1 & 1 & 0 & {} \\ \hline
\end{tabular}\\
\end{examples}
Note that rescans are only initiated upon encountering a negative
number, so that incrementing the last occurrence of $-1$ suffices to stop rescanning.
Here then is a more systematic tabulation of proper usage of the
above encoding \ref{encoding}, followed by pseudocode of the algorithm
interpreting it \ref{copyalg}.
\begin{Ex}
\item {\sc Number-of-copies encoding schema}\\
\label{encoding}
\begin{tabular}[t]{|l|c|c|c|}\hline
            & Scan 1 & Scan 2 & $| \# copies |$ \\ \hline\hline
output   & yes & yes & 2 \\ \cline{2-4}
segment & yes & no & 1 \\ \cline{2-4}
             &  no & yes & 0 \\ \hline
\end{tabular}
\item {\sc copy Algorithm}\\[-2ex]
\label{copyalg}
\startpiece
\begin{tabbing}
XXX \= XXX \= XXX \= \kill
\INPUT string of symbols Input[1\dots\/N] \CONDITION length(Input) $>$ 0 \\
ScanPos \ASSIGN TempPos \ASSIGN LastRescanPos \ASSIGN 0 \\
AtEnd \ASSIGN length(Input) \\
\REPEAT \\
    \>   ScanPos \ASSIGN ScanPos $+$ 1 \\
    \>   CurrentSymbol \ASSIGN Input[ScanPos] \\
    \>   \IF CurrentSymbol.NumCopies \UNEQUAL 0 \THEN \\
     \> \>        \OUTPUT CurrentSymbol.Segment \\
     \>   \ENDIF \\
      \>       \IF CurrentSymbol.NumCopies $<$ 0 \THEN \\
       \> \>             Rescan \ASSIGN true \\
       \> \>             TempPos \ASSIGN ScanPos \\
       \> \>             ScanPos \ASSIGN LastRescanPos  \\
       \> \>             \IF CurrentSymbol.NumCopies = -1 \THEN \\
      \> \> \>                     LastRescanPos \ASSIGN TempPos \\
        \> \>            \ENDIF \\
       \>      \ELSE Rescan \ASSIGN false \\
       \>      \ENDIF \\
        \>      \IF CurrentSymbol.NumCopies $>$ 0 \THEN \\
        \> \>                 CurrentSymbol.NumCopies \ASSIGN
        CurrentSymbol.NumCopies $-$ 1 \\
       \>        \ELSE                      \\
        \> \>                  CurrentSymbol.NumCopies \ASSIGN
        CurrentSymbol.NumCopies $+$ 1 \\
        \>      \ENDIF \\
\UNTIL ScanPos = AtEnd \AND \NOT Rescan  \\
\end{tabbing}\vspace*{-3ex}             
\stoppiece
\end{Ex}
It is not difficult to see that the algorithm always
terminates. Informally, because every input string is of finite length 
and the number of copies associated with each segmental positions must
become positive in a finite number of steps (by way of the last 
{\em  if-then-else} statement), the number of rescans initated by negative 
number-of-copies annotations will be finite as well.

What are salient properties of the solution just proposed? It is
\begin{itemize}
\item {\bf dynamic}, i.e.\/ the surface form of a given input {\em 
    token} can be computed by the algorithm, while the {\em set} of
  all such surface forms is not statically represented
\item {\bf not persistently linearised}, i.e.\/ the complete surface
  linearization of the input cannot be read off the input easily using 
  only finite-state mechanisms
\item {\bf not uniquely encoded}, i.e.\/ there is a countably infinite
  number of possible input codes for any given surface form. The chief source
  of infiniteness comes from 0-annotated material that is
  outside of a rescan domain, for example
  ${<}$t$_1$e$_2$c$_2$h$_2$t$_2$e$_2$l$_2$m$_{\text{-}1}$x$_0$x$_0$x$_0$x$_0$\dots${>}$. In this way one may add an  arbitrary amount of underlying material that will never be output by the copy algorithm.  

\end{itemize}
Let us comment on some of these aspects. The dynamic aspect seems
unavoidable as long as the finite-state assumption is upheld in its
most restricted form, i.e.\/ there are no online operations with
greater-than-finite-state power like composition, intersection etc.\/
during parsing and generation. 

The next  point about full
linearization being performed outside of phonology 
proper, however, causes more concern.  One reason  is that there are cases where the
shape of reduplicants depends on their surface position,  as we have
seen in Koasati (recall the contrast between penultimate heavy syllables in {\em ak-h{\bf
    o}-lat.lin} vs. {\em tahas-t{\bf oo}.-pin}). Of course this is just an instance 
of the  general fact that phonological
alternations often depend on surface syllable structure. But
those surface syllables may cut through the `folded' encoding of inputs:
e.g. ${<}$tech.tel.mech.tel${>}$ has ${<}$mech${>}$ as the third syllable, which
happens to be nonlocally represented under the input encoding
${<}$t$_1$\underline{e$_2$c$_2$h$_2$}t$_2$e$_2$l$_2$\underline{m$_{\text{-}1}$}${>}$.
Also, the example of Washo  shows that surface-derived syllable roles may differ
between reduplicant and base, with concomitant phonological
consequences (data taken from \citeNP[17]{wilbur:73}):
\begin{examples}
\item \label{washo} {\sc Surface linearization and Washo reduplication}\\
\begin{tabular}[t]{ll}
{\bf w\textprimstress e\underline{t}}we\underline{d}i & `it's quacking' \\
{\bf \ip{S}\textprimstress u\underline{p}}\ip{S}u\underline{b}i & `he's crying gently' \\
tum\textglotstop {\bf s\textprimstress o\underline{p}}so\underline{b}i & `he's splashing his feet' \\
{\bf b\textprimstress a\underline{k}}ba\underline{g}i & `he's smoking'
\end{tabular}
\end{examples}
In the data under \ref{washo} we can see that obstruents get devoiced
in the coda. Crucially, however, the relevant codas are indirectly created by
reduplication: a C$_1$VC$_2$ reduplicant prefixed to the base results in a
{\em VCCV} context that {\em surface-based} syllabification resolves by
assigning coda-onset roles to the {\em CC} cluster. Since {\em C$_2$}
in the base ends up being in an onset position, it is not devoiced,
resulting in the phonological difference between reduplicant and base
underlined in \ref{washo}.

Washo seems to present a problem for our proposed input encoding:
when using a representation like ${<}w_2e_2d_{\text{-}2}i_1{>}$ that
only covers the underlying base, the independence between base and reduplicant required for
phonological differences is lost. 
However, the last aspect about the non-uniqueness of input encodings --
which has not been illustrated so far -- actually provides hope for
such situations. Let us therefore have a look at some alternative encodings for
reduplicative forms in \ref{nonunique}.
\begin{figure}[htb]
\begin{examples}
\item {\sc Non-Uniqueness in Encoding of Surface Forms}\\
\label{nonunique}
\begin{tabular}[t]{lll}
a1. & techtelm{\bf echtel} & t e c h t e l m \\ 
     &                  & 1 2 2 2 2 2 2 -1 \\
a2.      &                 & t m e c h t e  l \\
       &                & 10 2 2 2 2 2 -2 \\[1ex]
b1. & schnick{\bf schn}a{\bf ck} & s c h n i a c k \\
    &                        & 2 2 2 2 1 0 2 -2 \\
b2.    &                        & s c h n a i c k \\
    &                        & 2 2 2 2 0 1 2 -2 \\[1ex]
c1. & tahas{\bf t}oopin & t a h a s o o p i n \\
   &                           & 2 1 1 1 -1 1 1 1 1 1 \\
c2.   &                           & t o o a h a s p i n \\
   &                           & 2 0 0 1 1 1 -1 1 1 1 \\
c3.   &                            & t a h a s p i n t o o \\
   &                            & 1 1 1 1 1 0 0 0 1 1 -1 \\
c4.   &                            & t a h a s p i n o o \\
   &                             & 2 1 1 1 -1 0 0 0 1 -1 \\
d1. & *{\bf wed}wedi & w e d i \\
     &   & 2 2 -2 1 \\
d2. & {\bf wet}wedi & w e t d i \\
      & & 2 2 -1 1 1  
\end{tabular}
\end{examples}
\end{figure}
Example \ref{nonunique}.a2 demonstrates an alternative, more local encoding for
{\em techtelmechtel} that brings original /t/ and its modification
/m/ into close proximity. It also allows to syllabify the string
/tmechtel/, as if it were a surface string -- e.g. using a finite-state version of the proposals of
\citeN{walther:92} et seq.
Unlike \ref{nonunique}.a2 
the critical /m/ will then become part of a complex onset, thus
correctly receiving the same
syllable role as in the linearized surface form. The examples
\ref{nonunique}.b1,2 show that locality vs nonlocality is not the only 
source of ambiguity in the linearization code:  original-modification pairs such
as ${<}i_1a_0{>}$ may be reversed in reduplicative constructions without affecting the surface
result. In general precedence relationships in the input string are only
relevant if the corresponding annotations target the same
scans. The next set of examples \ref{nonunique}.c1-4 shows a broad range of 
options  differing in the placement of fixed material.
\ref{nonunique}.c1 is approximately surface-true in so far as it has
the /oo/ in the penultimate syllable position. However, it simultaneously serves to
again show the potential for non-surface truth within individual syllable
{\em role} assignments: /s/ will be assigned an onset role, but the
surface realization has a coda instead. Example \ref{nonunique}.c2 
illustrates a different compromise in that it sees all of the reduplicative
affix as a non-distributed prefix. Example \ref{nonunique}.c3 shows
a slightly odd analysis that denies reduplicative
status to {\em tahastoopin} in annotating an input suffix to end up as an infix after
linearization. Finally, \ref{nonunique}.c4 separates reduplicated and
fixed-melody material, seeing only the latter as suffixed to the input 
base. The last two encodings share the property of leaving the base
uninterrupted, which may be advantageous in terms of compression
degree of the base lexicon. The last pair of examples
\ref{nonunique}.d1-2 illustrates how to proceed when syllable roles
and segmental realization differ between base and copy: instead of
incorrectly representing  just one token  \ref{nonunique}.d1, it is sometimes
possible to maintain two adjacent copies \ref{nonunique}.d2 in such a
way that the input encoding itself can be correctly syllabified.

As we have just seen, the encoding is quite versatile. In fact, it is not even
limited to describing reduplicative patterns alone but can also handle infixations,
circumfixations and truncations. \ref{moreex} shows a few more
examples from previous sections.
\begin{Ex}
\item {\sc More Encoded Examples}\\
\label{moreex}
\begin{tabular}[t]{ll}
\begin{tabular}[t]{ll}
{\em Encoding} & {\em Surface}\\\hline
w u l u o & wuluowulu \\
2 2 2 2 -1 \\[1ex]
c \textglotstop {} e e t & ctc\textglotstop eet \\
2 0 0 0 -2 \\[1ex]
s i l i n & silslin \\
2 1 -2 1 1 \\[1ex]
v e l o & velelo \\
1 2 -2 1 
\end{tabular} &
\begin{tabular}[t]{ll}
{\em Encoding} & {\em Surface}\\\hline
k r a n d h i & kanikrandh \\
2 0 2 2 0 0 -1 \\[1ex]
u m b a s a & bumasa \\
0 0 -1 1 1 1 \\[1ex]
b e r g g e t e & gebergte \\
0 0 0 0 1 -1 1 1 \\[1ex]
\ip{k K\;{} U S t S O f i} &  \ip{kK\;UStSi} \\
1 1 1 1 1 1 0 0 1 
\end{tabular}
\end{tabular}
\end{Ex}
The Bella Coola form {\em silslin} in  \ref{moreex} shows that straightforward
use of the encoding  undermines a succinct
analysis of cases where segmental realizations are the same, although
the syllable roles differ between a base 
segment and its corresponding partner in the reduplicant. Assuming the 
syllabification /.sil.-slin./, the /l/ is in coda position in the
reduplicative prefix (spelled out in scan 1), but in onset position in
the base (spelled out in scan 2) -- contrary to 
the unique onset position it receives when directly syllabifying the
input /silin/. Of course, by introducing a redundant extra /l/ like in 
the Washo case we could rescue surface-like syllabfication, albeit at
the cost of introducing considerable redundancy and with no other
motivation but to avoid technical difficulties. (Note that in this case,
though, the systematic {\em local} copy of a single consonant $C_iC_{i,copy}$ in each base would be
technically feasible using a finite-state rule). To conclude, 
while all the information to ultimately deduce surface position --
and thus, surface prosodic role -- {\em is} formally present in the input, it
is often difficult to exploit that information. Sometimes at least
the individual solutions would seem to include construction-specific,
non-surface-referring rules for finite-state syllabification that
preclude maximal  modularity and component reuse in analyses.

Thus far I have described the proposed local encoding method from a
generation or production perspective only, where a single string of
complex symbols gets spelled out on the surface. It remains to show
that the method is also usable for parsing or perception, where
matching against a finite-state-encoded {\em set} of strings is
required. Also, finite-state networks are attractive because
minimization algorithms exist to reduce storage requirements, and it is 
natural to ask whether these continue to produce good results under the new
encoding.

Turning first to the issue of {\bf parsing surface forms}, let us 
assume as before that the lexicon takes the form of an acyclic FSA or
FST, using complex symbols to encode segments. In a loop the parsing
algorithm would then read a single surface segment and try to match it 
to the segment symbol of an arc emanating from the current state of
the network. However, encountering a symbol with zero annotation
causes that arc to be skipped, repeating the matching attempt with the 
following arc(s). For various reasons it may become necessary to
nondeterministically choose which arc from a set of possible arcs
should be followed. As the linearisation of encoded strings works off a single
string only, and furthermore modifies its number-of-copies annotations 
during consecutive scans, we need to copy at least the annotation information of the
partial path hypothesis that is currently pursued to a separate
buffer. The copying can of course be done incrementally. All rescans then
will only exame this buffer, which contains the current 
state of number-of-copies annotations, whereas the network itself remains
unmodified.%
\footnote{Note that with very large networks -- such as
  that employed in AT\&T speech applications --, the network data
  themselves may reside on read-only disk and not fit into main memory at
  all. Rather, they will be swapped in on demand through
  memory-mapped files. Similarly, networks might be stored in ROM,
  again necessitating RAM-resident copies to record on-the-fly modifications.}
Also, the stored path hypothesis must be retracted upon
backtracking -- to be initiated when it turns out that the current hypothesis
cannot be correct because a segment match fails -- so that we actually 
need a stack of partial path buffers. A match failure pops off one
entry, whereas each nondeterministic choice records the arc chosen and 
provides a new partial path buffer which initially contains a copy of
the previous one. However, upon closer examination it becomes clear
that recording current scan number and {\tt ScanPos} actually suffice to
uniquely determine the state of the algorithm in \ref{copyalg}, hence
we might trade (recomputation) time for space in this way.
Details remain to be worked out how to exactly
define suitable data structures and the recognition algorithms itself, 
but it seems reasonably clear that this can be done. 

Actually, a
second route to take would be to make the copy algorithm \ref{copyalg} 
itself reversible, instead of implementing a separate version for
recognition. Borrowing from existing work in Prolog implementation it 
seems sufficient to implement a backtrackable destructive assignment
operation on top of the backtracking mechanism that is independently
needed for traversing nondeterministic networks. Again, this sketch
obviously needs more detail to facilitate full evaluation.

The final issue worth discussing is {\bf minimization and storage requirements} for
networks encoded with the proposed method. Obviously, if the same
segments carry different number-of-copies annotations, they are
different when viewed as complex symbols. Without further means this
simple fact diminishes the chance for sharing of common substrings.
This is illustrated  in  \ref{automata} using the two German words {\em techtelmechtel, technik}.
\begin{Ex}
\item \label{automata} {\sc Sharing in traditional and complex symbol automata}\\
\epsfig{file=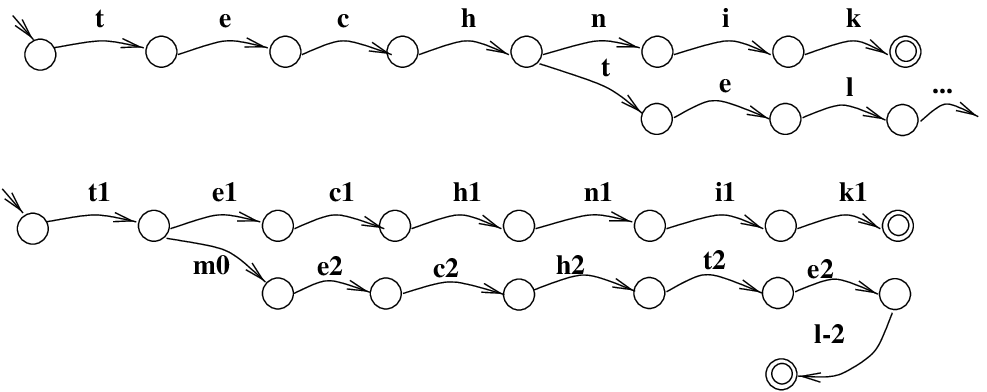}\hfill
\end{Ex}
However, only large-scale experiments with lexicons containing
reduplicative constructions can tell whether the loss of sharing
compared to a naive full-form lexicon is not outweighed in practice by the 
saving that comes with not having to repeat reduplicative parts on the
segmental level. 

In the face of these initial difficulties, a natural
question to ask is this: can we have the best of both worlds,
i.e.\/ minimization results approaching that of non-complex-encoded
networks {\em} and the space saving that potentially comes with the proposed method? 
To preview, it seems that only approximative solutions are possible. A first idea
that comes to mind is to linearize the encoding, following each
segment with the annotation that goes with it, e.g {\em
  t1m0e2c2h2t2e2l-2}.%
\footnote{Note that this linearized encoding generalizes to finite-state
  transducers, where $\underrightarrow{Input:Output}$ pairs can be sequentialized as
  $\underrightarrow{Input} _{_{\circ}} \underrightarrow{Output}$ to
  convert the transducer into an ordinary automaton (cf.\/ fig.\/ 9 in US patent
  5,625,554 granted to Xerox on April 29, 1997). Of course,
  application of underlying and surface strings must then skip odd and 
  even arcs to match and also output the following or preceding arc as
  result.}
 However, it is easy to see that -- given a
regular set $S$ containing only even-length strings -- the
`intercalated union' $U$ of the 
projection $O$ of $S$ containing only the letters at odd positions
(e.g. segments like $tmechtel$) with
the projection $E$ containing only even string positions
(e.g. annotations like $1022222$-$2$) is not equivalent 
to $S$ itself, thus precluding an approach that would couple separate offline minimization  with
parallel online traversal, i.e.\/ alternating between the $O$ and $E$ machines
depending on string position. In general $U$ is strictly larger than
$S$, because choices in $E$ and $O$ are not synchronized. One could
try to develop sophisticated schemes that implement a kind of
distributed disjunctions to effect synchronization, while simultaneously
taking into account that in our application $|E| \ll |U|$. However,
in order to decide which synchronized choice to make in $E$  when having done a
previous choice at a node $k \in O$, we  in general require knowledge of
the entire path $p_{k,O}$ travelled so far. Unfortunately, storing
such paths and synchronization marks is likely to become costly again.

A second approach to improving minimization quality would consist of
building a variant of the network that implements perfect hashing from 
each word $w_i$ into a natural number $i$ from $1 \dots N \in \cal{N}$, again assuming an
acyclic FSA/FST that contains a finite number $N$ of words (see
e.g. \citeNP{daciuk:98} for an implementation, or US patent 5,551,026
granted August 27, 1996 to Xerox). We could
then store the annotation vector $v_i$ for $w_i$ in a table indexed by
$i$. Unfortunately, this will definitely increase the amount of
search/backtracking in recognition, as the annotations are crucial in
defining surface shape, but will not become available incrementally in 
this approach. It might also not help much in saving storage space unless
clever compression is employed for the annotation table.

Another approach sees the annotations as weights in a weighted
FSA/FST. Again each word $w_i$ has an annotation vector $v_i$, but now 
we can view the vectors as being composed from suitable smaller parts
distributed over the network so as to facilitate maximal
compression. For example, weights could be strings from $($-$2|$-$1|0|1|2)^{*}$
and the weight-combining operator could be concatenation.
There exist minimization methods for the weighted case
that essentially comprise two phases \cite{mohri.riley.sproat:96}. The first phase pushes the
weights to the beginning of the FSA/FST as far as possible, thus
enabling greater similarity -- and therefore greater shareability -- towards the end. The 
second phase consists of ordinary FSA/FST minimization. This approach
then promises to adress the storage issue better than the previous one, 
although it is similar in that again one has to wait till the end of
each input string to get the composed total weight that drives the
inverse copy algorithm.

Finally, annotations could simply be output strings in an ordinary
FST, more precisely a sequential transducer, where the associated
input automataton is deterministic. \citeN{mohri:94} has described an
algorithm to make such transducers very compact, based on the closely related
idea of pushing partial output {\em strings} (represented as members of an
enlarged alphabet) towards the initial state of
the FST, and then mimimizing this FST in the sense of traditional
automata mimimization. Advantages and disadvantages of this approach
seem to be quite similar to the previous one, except that by indentifying
output strings with annotations instead of traditional surface strings
we have perhaps lost the ability to use canonical
underlying representations with morphological and other annotations.
\newpage
\section{\label{transducer-redup}Alternative method:\\ Reduplicative copying as transducer composition}
{\small
\begin{verbatim}
    1   %% This code assumes FSA Utilities (http://www.let.rug.nl/~vannoord/Fsa/)
    2   
    3   :- op(1200,xfx,':=').  %% separates name and definition of macro
    4   
    5   %% expand new macro notation into FSA Utilities format during file read-in
    6   
    7   :- multifile (user:term_expansion/2).
    8   user:term_expansion(':='(Head,Body), macro(Head,Body)).
    9   user:term_expansion(':-'(':='(Head,Body), PrologGoals),
   10                     (macro(Head,Body) :- PrologGoals)).
   11   
   12   %% the grammar proper
   13   
   14   % add_repeats/1 is best defined as manipulation of the underlying automaton:
   15   % for each noncylic arc A --> B, add B --[]/repeat--> A
   16   
   17   rx(add_repeats(E), FA) :- 
   18     (fsa_regex:rx(E, FA0)),
   19     (fsa_regex:add_symbols([repeat], FA0, FA1)),
   20     (fsa_data:copy_fa_except(transitions, FA1, FA, Transitions0, Transitions)),
   21     findall(trans(From, []/repeat, To), ( member(trans(To, _, From), Transitions0),
   22                                          To \== From ), RepeatTransitions),
   23     append(Transitions0, RepeatTransitions, Transitions1),
   24     sort(Transitions1, Transitions).
   25   
   26   boundary := escape(#).
   27   epsilon := [].
   28   technical_symbols := { boundary, repeat }.
   29   
   30   enriched_words([]) := {}.
   31   enriched_words([Word|Words]) :=
   32             { add_repeats([ epsilon:boundary, word(Word), epsilon:boundary ]),
   33               enriched_words(Words) }.
   34   
   35   % N.B. {} denotes the empty language, [] is the empty string/epsilon. The builtin 
   36   % word(Atom) yields an automaton where the characters of Atom are concatenated
   37   
   38   reduplicant := [boundary:epsilon, (? - technical_symbols) *, boundary:epsilon].
   39   
   40   % N.B. ? is the any (meta) symbol, A - B is set difference.
   41   % Regular languages are automatically coerced to relations/transducers
   42   % if necessary (e.g. in the context of composition o)
   43   
   44   total_reduplication(WordList) :=
   45                enriched_words(WordList)
   46                            o
   47                [reduplicant, repeat:epsilon *, reduplicant].
   48   test := total_reduplication([orang,utan]).
\end{verbatim}}

\begin{thebibliography}{}

\bibitem[\protect\citeauthoryear{??}{roa}{}]{roa}
[{\em CLA-XXX: paper no.\/ XXX from http://xxx.lanl.gov/abs/cs.CL}.

\bibitem[\protect\citeauthoryear{??}{cla}{}]{cla}
{\em ROA-XXX: paper no.\/ XXX from
  http://ruccs.rutgers.edu/ROA/showpaper.html}].

\bibitem[\protect\citeauthoryear{Abrusci, Fouquer\'{e} \& Vauzeilles}{Abrusci
  et~al.}{1999}]{abrusci.fouquere.vauzeilles:99}
Abrusci, V.~M., C.~Fouquer\'{e} \& J.~Vauzeilles (1999).
\newblock Tree Adjoining Grammars in a Fragment of the Lambek Calculus.
\newblock {\em Computational Linguistics\/}~{\em 25\/}(2), 209--236.

\bibitem[\protect\citeauthoryear{A{\"i}t-Kaci, Boyer, Lincoln \&
  Nasr}{A{\"i}t-Kaci et~al.}{1989}]{ait-kaci.et.al:89}
A{\"i}t-Kaci, H., R.~Boyer, P.~Lincoln \& R.~Nasr (1989).
\newblock Efficient Implementation of Lattice Operations.
\newblock {\em ACM Transactions on Programming Languages and Systems\/}~{\em
  11\/}(1), 115--146.

\bibitem[\protect\citeauthoryear{Albro}{Albro}{1997}]{albro:97}
Albro, D. (1997).
\newblock Evaluation, Implementation, and Extension of Primitive Optimality
  Theory.
\newblock Master's thesis, Department of Linguistics, UCLA.

\bibitem[\protect\citeauthoryear{Anderson}{Anderson}{1992}]{anderson:92}
Anderson, S.~R. (1992).
\newblock {\em A-Morphous Morphology}.
\newblock Cambridge: Cambridge University Press.

\bibitem[\protect\citeauthoryear{Antworth}{Antworth}{1990}]{antworth:90}
Antworth, E. (1990).
\newblock {\em PC-KIMMO: A Two-Level Processor for Morphological Analysis}.
\newblock Dallas: SIL.

\bibitem[\protect\citeauthoryear{Applegate}{Applegate}{1976}]{applegate:76}
Applegate, R.~B. (1976).
\newblock Reduplication in Chumash.
\newblock In: M.~Langdon \& S.~Silver (Eds.), {\em Hokan Studies: papers from
  the First Conference on Hokan Languages, held in San Diego, California, April
  23 - 25, 1970},  271--283. The Hague: Mouton.

\bibitem[\protect\citeauthoryear{Bayer \& Johnson}{Bayer \&
  Johnson}{1995}]{bayer.johnson:95}
Bayer, S.~L. \& M.~Johnson (1995).
\newblock Features and Agreement.
\newblock In: {\em Proceedings of the 33rd Annual Meeting of the Association
  for Computational Linguistics}.

\bibitem[\protect\citeauthoryear{Beesley, Buckwalter \& Newton}{Beesley
  et~al.}{1989}]{beesley.buckwalter.newton:89}
Beesley, K., T.~Buckwalter \& S.~Newton (1989).
\newblock Two-level finite-state analysis of Arabic morphology.
\newblock In: {\em Proceedings of the Seminar on Bilingual Computing in Arabic
  and English}, Cambridge. The Literary and Linguistic Computing Centre.

\bibitem[\protect\citeauthoryear{Beesley}{Beesley}{1996}]{beesley:96}
Beesley, K.~R. (1996).
\newblock Arabic Finite-State Morphological Analysis and Generation.
\newblock In: {\em Proceedings of COLING-96}, Vol.\/~I,  89--94.

\bibitem[\protect\citeauthoryear{Beesley}{Beesley}{1998}]{beesley:98}
Beesley, K.~R. (1998).
\newblock Constraining Separated Morphotactic Dependencies in Finite-State
  Grammars.
\newblock In: {\em Proceedings of FSMNLP'98, Bilkent University, Turkey}.

\bibitem[\protect\citeauthoryear{Belz}{Belz}{1998}]{belz:98}
Belz, A. (1998).
\newblock Discovering phonotactic finite-state automata by genetic search.
\newblock In: {\em Proceedings of COLING-ACL '98}.

\bibitem[\protect\citeauthoryear{Bendor-Samuel}{Bendor-Samuel}{1960}]{bendor-s%
amuel:60}
Bendor-Samuel, J.~T. (1960).
\newblock Segmentation in the phonological analysis of {T}erena.
\newblock {\em Word\/}~{\em 16\/}(3), 348--55.

\bibitem[\protect\citeauthoryear{Bird}{Bird}{1990}]{bird:90}
Bird, S. (1990).
\newblock {\em Constraint-Based Phonology}.
\newblock Ph.D. thesis, University of Edinburgh.

\bibitem[\protect\citeauthoryear{Bird \& Blackburn}{Bird \&
  Blackburn}{1991}]{bird.blackburn:91}
Bird, S. \& P.~Blackburn (1991).
\newblock A logical approach to {A}rabic phonology.
\newblock In: {\em Proceedings of the Fifth Meeting of the European Chapter of
  the Association for Computational Linguistics},  89--94. Association for
  Computational Linguistics.

\bibitem[\protect\citeauthoryear{Bird \& Ellison}{Bird \&
  Ellison}{1992}]{bird.ellison:92}
Bird, S. \& T.~M. Ellison (1992).
\newblock {O}ne-{L}evel {P}honology: Autosegmental Representations and Rules as
  Finite-State Automata.
\newblock Technical report, Centre for Cognitive Science, University of
  Edinburgh.
\newblock EUCCS/RP-51.

\bibitem[\protect\citeauthoryear{Bird \& Ellison}{Bird \&
  Ellison}{1994}]{bird.ellison:94}
Bird, S. \& T.~M. Ellison (1994).
\newblock {O}ne-{L}evel {P}honology.
\newblock {\em Computational Linguistics\/}~{\em 20\/}(1), 55--90.

\bibitem[\protect\citeauthoryear{Bird \& Klein}{Bird \&
  Klein}{1990}]{bird.klein:90}
Bird, S. \& E.~Klein (1990).
\newblock Phonological events.
\newblock {\em Journal of Linguistics\/}~{\em 26}, 33--56.

\bibitem[\protect\citeauthoryear{Blevins}{Blevins}{1995}]{blevins:95}
Blevins, J. (1995).
\newblock The Syllable in Phonological Theory.
\newblock In: J.~A. Goldsmith (Ed.), {\em The Handbook of Phonological Theory},
   206--244. Oxford: Basil Blackwell.

\bibitem[\protect\citeauthoryear{Blevins}{Blevins}{1996}]{blevins:96}
Blevins, J. (1996).
\newblock Mokilese Reduplication.
\newblock {\em Linguistic Inquiry\/}~{\em 27\/}(3), 523--530.

\bibitem[\protect\citeauthoryear{Bloomfield}{Bloomfield}{1933}]{bloomfield:33}
Bloomfield, L. (1933).
\newblock {\em Language}.
\newblock New York: Holt.

\bibitem[\protect\citeauthoryear{Boersma}{Boersma}{1998}]{boersma:98}
Boersma, P. (1998).
\newblock {\em Functional Phonology. Formalizing the interactions between
  articulatory and perceptual drives}.
\newblock Ph.D. thesis, University of Amsterdam.
\newblock [The Hague: Holland Academic Graphics].

\bibitem[\protect\citeauthoryear{Boersma \& Hayes}{Boersma \&
  Hayes}{1999}]{boersma.hayes:99}
Boersma, P. \& B.~Hayes (1999).
\newblock Empirical tests of the Gradual Learning Algorithm.
\newblock Ms., University of Amsterdam \& UCLA.
\newblock (ROA-348-1099).

\bibitem[\protect\citeauthoryear{Bouma \& Nerbonne}{Bouma \&
  Nerbonne}{1994}]{bouma.nerbonne:94}
Bouma, G. \& J.~Nerbonne (1994).
\newblock Lexicons for Feature-Based Systems.
\newblock In: H.~Trost (Ed.), {\em Tagungsband KONVENS '94}, Nummer~6 in
  Informatik Xpress,  42--51. Wien: {\"O}sterreichische Gesellschaft f{\"u}r
  Artificial Intelligence.

\bibitem[\protect\citeauthoryear{Bresnan \& Kaplan}{Bresnan \&
  Kaplan}{1982}]{bresnan.kaplan:82}
Bresnan, J. \& R.~Kaplan (1982).
\newblock Lexical Functional Grammar: A Formal System for Grammatical
  Representation.
\newblock In: J.~Bresnan (Ed.), {\em The Mental Representation of Grammatical
  Relations}. Cambridge, MA: MIT Press.

\bibitem[\protect\citeauthoryear{Brzozowski}{Brzozowski}{1962}]{brzozowski:62}
Brzozowski, J.~A. (1962).
\newblock Canonical regular expressions and minimal state graphs for definite
  events.
\newblock {\em Mathematical theory of automata\/}~{\em 12}, 529--561.

\bibitem[\protect\citeauthoryear{Carrier}{Carrier}{1979}]{carrier:79}
Carrier, J.~L. (1979).
\newblock {\em The interaction of morphological and phonological rules in
  Tagalog: a study in the relationship between rule components in grammar}.
\newblock Ph.D. thesis, MIT, Cambridge, MA.

\bibitem[\protect\citeauthoryear{Chomsky}{Chomsky}{1965}]{chomsky:65}
Chomsky, N. (1965).
\newblock {\em Aspects of the Theory of Syntax}.
\newblock Cambridge, MA: MIT Press.

\bibitem[\protect\citeauthoryear{Chomsky \& Halle}{Chomsky \&
  Halle}{1968}]{chomsky.halle:68}
Chomsky, N. \& M.~Halle (1968).
\newblock {\em Sound {P}attern of {E}nglish}.
\newblock New York: Harper and Row.

\bibitem[\protect\citeauthoryear{Culy}{Culy}{1985}]{culy:85}
Culy, C. (1985).
\newblock The complexity of the vocabulary of Bambara.
\newblock {\em Linguistics and Philosophy\/}~{\em 8}, 345--351.

\bibitem[\protect\citeauthoryear{Daciuk}{Daciuk}{1998}]{daciuk:98}
Daciuk, J. (1998).
\newblock {\em Incremental Construction of Finite-State Automata and
  Transducers, and their use in Natural Language Processing}.
\newblock Ph.D. thesis, Dep. of Electronics, Telecommunications and
  Informatics, Polytechnic, University of Gdansk, Poland.
\newblock [http://www.pg.gda.pl/~jandac/thesis/thesis.html].

\bibitem[\protect\citeauthoryear{Dahl, Tarau \& Li}{Dahl
  et~al.}{1997}]{dahl.tarau.li:97}
Dahl, V., P.~Tarau \& R.~Li (1997).
\newblock Assumption {G}rammars for {P}rocessing {N}atural {L}anguage.
\newblock In: {\em Proceedings of ICLP '97}.

\bibitem[\protect\citeauthoryear{Dalrymple}{Dalrymple}{1999}]{dalrymple:99}
Dalrymple, M. (Eds.) (1999).
\newblock {\em Semantics and Syntax in Lexical Functional Grammar}.
\newblock MIT Press.

\bibitem[\protect\citeauthoryear{Dijkstra}{Dijkstra}{1959}]{dijkstra:59}
Dijkstra, E.~W. (1959).
\newblock A note on two problems in connexion with graphs.
\newblock {\em Numerische Mathematik\/}~{\em 1}, 269--271.

\bibitem[\protect\citeauthoryear{Eisner}{Eisner}{1997a}]{eisner:97}
Eisner, J. (1997a).
\newblock Efficient generation in primitive Optimality Theory.
\newblock In: {\em Proceedings of the 35th Annual Meeting of the Association
  for Computational Linguistics and the 8th Conference of the European
  Association for Computational Linguistics}, Madrid.

\bibitem[\protect\citeauthoryear{Eisner}{Eisner}{1997b}]{eisner:97a}
Eisner, J. (1997b).
\newblock FootForm decomposed: Using primitive constraints in OT.
\newblock In: B.~Bruening (Ed.), {\em MIT Working Papers in Linguistics},
  Vol.\/~31. MIT.

\bibitem[\protect\citeauthoryear{Ellison}{Ellison}{1992}]{ellison:92}
Ellison, T.~M. (1992).
\newblock {\em Machine Learning of Phonological Representations}.
\newblock Ph.D. thesis, University of Western Australia, Perth.

\bibitem[\protect\citeauthoryear{Ellison}{Ellison}{1993}]{ellison:93}
Ellison, T.~M. (1993).
\newblock Concatenation vs conjunction in constraint-based phonology.
\newblock In: T.~M. Ellison \& J.~M. Scobbie (Eds.), {\em Computational
  Phonology}, Vol.\/~8 von {\em Edinburgh Working Papers in Cognitive Science},
   1--18. Centre for Cognitive Science, Edinburgh.

\bibitem[\protect\citeauthoryear{Ellison}{Ellison}{1994a}]{ellison:94a}
Ellison, T.~M. (1994a).
\newblock Constraints, Exceptions and Representations.
\newblock In: {\em Proceedings of ACL SIGPHON First Meeting},  25--32.
\newblock (ROA-76, CLA-9504022).

\bibitem[\protect\citeauthoryear{Ellison}{Ellison}{1994b}]{ellison:94}
Ellison, T.~M. (1994b).
\newblock Phonological Derivation in Optimality Theory.
\newblock In: {\em Proceedings of COLING '94}, Vol.\/~II,  1007--1013.
\newblock (ROA-75, CLA-9504021).

\bibitem[\protect\citeauthoryear{Ellison}{Ellison}{to
  appear}]{ellison:to_appear}
Ellison, T.~M. ({to appear}).
\newblock The Universal Constraint Set: Convention, not Fact.
\newblock In: J.~Dekkers, F.~{van der Leeuw} \& J.~{van de Weijer} (Eds.), {\em
  Conceptual Studies in Optimality Theory}. Oxford University Press.

\bibitem[\protect\citeauthoryear{F\'{e}ry}{F\'{e}ry}{1997}]{fery:97}
F\'{e}ry, C. (1997).
\newblock Uni und Studis: die besten W\"orter des Deutschen.
\newblock {\em Linguistische Berichte\/}~{\em 172}, 461--489.

\bibitem[\protect\citeauthoryear{Frank \& Satta}{Frank \&
  Satta}{1998}]{frank.satta:98}
Frank, R. \& G.~Satta (1998).
\newblock Optimality Theory and the Generative Complexity of Constraint
  Violability.
\newblock {\em Computational Linguistics\/}~{\em 24\/}(2), 307--315.

\bibitem[\protect\citeauthoryear{Goldsmith}{Goldsmith}{1990}]{goldsmith:90}
Goldsmith, J. (1990).
\newblock {\em Autosegmental and {Metrical} {P}honology}.
\newblock Oxford and Cambridge, MA: Basil Blackwell.

\bibitem[\protect\citeauthoryear{Golston \& Wiese}{Golston \&
  Wiese}{1996}]{golston.wiese:96}
Golston, C. \& R.~Wiese (1996).
\newblock Zero morphology and constraint interaction: subtraction and
  epenthesis in German dialects.
\newblock In: G.~Booij \& J.~van Marle (Eds.), {\em Yearbook of Morphology
  1995},  143--159. Kluwer Academic Publishers.

\bibitem[\protect\citeauthoryear{Green}{Green}{1999}]{green:99}
Green, T.~M. (1999).
\newblock {\em A Lexicographic Study of Ulwa}.
\newblock Ph.D. thesis, MIT, Cambridge, MA.
\newblock (\url{http://web.\/mit.\/edu/tmgreen/www/thesis.\/pdf)}.

\bibitem[\protect\citeauthoryear{Grimes}{Grimes}{1996}]{ethnologue:96}
Grimes, B.~F. (1996).
\newblock {\em Ethnologue}.
\newblock Dallas, TX: Summer Institute of Linguistics.
\newblock Thirteenth Edition, http://www.sil.org/ethnologue.

\bibitem[\protect\citeauthoryear{Hayes}{Hayes}{1995}]{hayes:95}
Hayes, B. (1995).
\newblock {\em Metrical stress theory: principles and case studies}.
\newblock University of Chicago Press.

\bibitem[\protect\citeauthoryear{Hayes}{Hayes}{1999}]{hayes:99}
Hayes, B.~P. (1999).
\newblock Phonetically Driven Phonology.
\newblock In: M.~Darnell, E.~Moravcsik, F.~Newmeyer, M.~Noonan \& K.~Wheatley
  (Eds.), {\em Functionalism and Formalism in Linguistics}, Vol.\/ I: General
  Papers,  243--285. Amsterdam: John Benjamins.

\bibitem[\protect\citeauthoryear{Hoeksema \& Janda}{Hoeksema \&
  Janda}{1988}]{hoeksema.janda:88}
Hoeksema, J. \& R.~D. Janda (1988).
\newblock Implications of Process-Morphology for Categorial Grammar.
\newblock In: R.~T. Oehrle, E.~Bach \& D.~Wheeler (Eds.), {\em Categorial
  grammars and natural language structures},  199--247. Dordrecht: Reidel.

\bibitem[\protect\citeauthoryear{Hopcroft \& Ullman}{Hopcroft \&
  Ullman}{1979}]{hopcroft:ullman:79}
Hopcroft, J. \& J.~Ullman (1979).
\newblock {\em Introduction to Automata Theory, Languages and Computation}.
\newblock Reading, MA: Addison-Wesley.

\bibitem[\protect\citeauthoryear{Johnson}{Johnson}{1997}]{johnson:97}
Johnson, M. (1997).
\newblock Features as Resources in R-LFG.
\newblock In: {\em Proceedings of the LFG-97 conference}. CSLI Press.

\bibitem[\protect\citeauthoryear{Kahn}{Kahn}{1976}]{kahn:76}
Kahn, D. (1976).
\newblock {\em Syllable-Based Generalizations in English Phonology}.
\newblock Bloomington: Indiana University Linguistics Club.
\newblock (= MIT Ph.D. dissertation).

\bibitem[\protect\citeauthoryear{Kaplan \& Kay}{Kaplan \&
  Kay}{1994}]{kaplan.kay:94}
Kaplan, R. \& M.~Kay (1994).
\newblock Regular models of phonological rule systems.
\newblock {\em Computational Linguistics\/}~{\em 20\/}(3), 331--78.

\bibitem[\protect\citeauthoryear{Karttunen}{Karttunen}{1994}]{karttunen:94}
Karttunen, L. (1994).
\newblock Constructing Lexical Transducers.
\newblock In: {\em Proceedings of COLING '94}, Vol.\/~I, Kyoto, Japan,
  406--411.

\bibitem[\protect\citeauthoryear{Karttunen}{Karttunen}{1996}]{karttunen:96}
Karttunen, L. (1996).
\newblock Directed Replacement.
\newblock In: {\em The Proceedings of the 34rd Annual Meeting of the
  Association for Computational Linguistics}, Santa Cruz, California.

\bibitem[\protect\citeauthoryear{Karttunen}{Karttunen}{1998}]{karttunen:98}
Karttunen, L. (1998).
\newblock The Proper Treatment of Optimality in Computational Phonology.
\newblock In: {\em Proceedings of FSMNLP'98. International Workshop on
  Finite-State Methods in Natural Language Processing}, Bilkent University,
  Ankara,  1--12.
\newblock (CLA-9804002,ROA-258-0498).

\bibitem[\protect\citeauthoryear{Karttunen, Kaplan \& Zaenen}{Karttunen
  et~al.}{1992}]{karttunen.kaplan.zaenen:92}
Karttunen, L., R.~M. Kaplan \& A.~Zaenen (1992).
\newblock Two-level morphology with composition.
\newblock In: {\em Proceedings of COLING '92}, Nantes,  141--148.

\bibitem[\protect\citeauthoryear{Kataja \& Koskenniemi}{Kataja \&
  Koskenniemi}{1988}]{kataja.koskenniemi:88}
Kataja, L. \& K.~Koskenniemi (1988).
\newblock Finite-state Description of Semitic morphology: A Case Study of
  Ancient Akkadian.
\newblock In: {\em COLING-88}, Vol.\/~1,  313--15.

\bibitem[\protect\citeauthoryear{Katamba}{Katamba}{1993}]{katamba:93}
Katamba, F. (1993).
\newblock {\em Morphology}.
\newblock Basingstoke: Macmillan.

\bibitem[\protect\citeauthoryear{Kenstowicz}{Kenstowicz}{1994}]{kenstowicz:94}
Kenstowicz, M. (1994).
\newblock {\em Phonology in Generative Grammar}.
\newblock Cambridge MA \& Oxford UK: Blackwell.

\bibitem[\protect\citeauthoryear{Kimball}{Kimball}{1988}]{kimball:88}
Kimball, G. (1988).
\newblock Koasati reduplication.
\newblock In: W.~Shipley (Ed.), {\em In honour of Mary Haas: from the Haas
  Festival Conference on Native American Linguistics},  431--42. Berlin: Mouton
  de Gruyter.

\bibitem[\protect\citeauthoryear{Kiraz}{Kiraz}{1994}]{kiraz:94}
Kiraz, G.~A. (1994).
\newblock Multi-tape two-level morphology: a case study in {S}emitic nonlinear
  morphology.
\newblock In: {\em Proceedings of COLING '94}, Vol.\/~1,  180--186.

\bibitem[\protect\citeauthoryear{Kiraz}{Kiraz}{1996}]{kiraz:96b}
Kiraz, G.~A. (1996).
\newblock S{\d{e}}mH{\d{e}}: A Generalised Two-Level System.
\newblock In: {\em Proceedings of ACL '96}.

\bibitem[\protect\citeauthoryear{Kiraz}{Kiraz}{1999}]{kiraz:99}
Kiraz, G.~A. (1999).
\newblock Compressed Storage of Sparse Finite-State Transducers.
\newblock In: {\em Proceedings of WIA '99}, Potsdam.

\bibitem[\protect\citeauthoryear{Kisseberth}{Kisseberth}{1970}]{kisseberth:70a}
Kisseberth, C.~W. (1970).
\newblock Vowel elision in Tonkawa and derivational constraints.
\newblock In: J.~L. Saddock \& A.~L. Vanek (Eds.), {\em Studies presented to
  Robert B. Lees by his students},  109--137. Champaign, IL: Linguistic
  Research.

\bibitem[\protect\citeauthoryear{Klein}{Klein}{1993}]{klein:93}
Klein, E. (1993).
\newblock Sierra {M}iwok verb stems.
\newblock In: T.~M. Ellison \& J.~M. Scobbie (Eds.), {\em Computational
  Phonology. Edinburgh Working Papers in Cognitive Science}, Vol.\/~8,  19--35.
  University of Edinburgh, Centre for Cognitive Science.

\bibitem[\protect\citeauthoryear{Kornai}{Kornai}{1996}]{kornai:96}
Kornai, A. (1996).
\newblock Vectorized Finite-State Automata.
\newblock In: A.~Kornai (Ed.), {\em Proceedings of the ECAI '96 Workshop {\em
  Extended Finite-State Models of Language}}, Budapest,  36--41. European
  Coordinating Committee for Artificial Intelligence.

\bibitem[\protect\citeauthoryear{Koskenniemi}{Koskenniemi}{1983}]{koskenniemi:%
83}
Koskenniemi, K. (1983).
\newblock {\em Two-{L}evel {M}orphology: {A} {G}eneral {C}omputational {M}odel
  for {W}ord-{F}orm {R}ecognition and {P}roduction}.
\newblock Ph.D. thesis, University of Helsinki.

\bibitem[\protect\citeauthoryear{Lieber}{Lieber}{1990}]{lieber:90}
Lieber, R. (1990).
\newblock {\em On the Organization of the Lexicon}.
\newblock New York/London: Garland.

\bibitem[\protect\citeauthoryear{Marantz}{Marantz}{1982}]{marantz:82}
Marantz, A. (1982).
\newblock Re Reduplication.
\newblock {\em Linguistic Inquiry\/}~{\em 13\/}(3), 435--482.

\bibitem[\protect\citeauthoryear{McCarthy \& Prince}{McCarthy \&
  Prince}{1991}]{mccarthy.prince:91}
McCarthy, J. \& A.~Prince (1991).
\newblock Lectures on prosodic morphology.
\newblock Linguistic Society of America Summer Institute, University of
  California, Santa Cruz.

\bibitem[\protect\citeauthoryear{McCarthy \& Prince}{McCarthy \&
  Prince}{1993}]{mccarthy.prince:93}
McCarthy, J. \& A.~Prince (1993).
\newblock Prosodic Morphology I: Constraint Interaction and Satisfaction.
\newblock Technical report RuCCS-TR-3, Rutgers University Center for Cognitive
  Science.

\bibitem[\protect\citeauthoryear{McCarthy \& Prince}{McCarthy \&
  Prince}{1994}]{mccarthy.prince:94}
McCarthy, J. \& A.~Prince (1994).
\newblock {T}he {E}mergence of the {U}nmarked: {O}ptimality in {P}rosodic
  {M}orphology.
\newblock In: M.~Gonzalez (Ed.), {\em Proceedings of the North-East Linguistics
  Society}, Vol.\/~24, Amherst, MA,  333--379. Graduate Linguistic Student
  Association. (ROA-13).

\bibitem[\protect\citeauthoryear{McCarthy \& Prince}{McCarthy \&
  Prince}{1995}]{mccarthy.prince:95a}
McCarthy, J. \& A.~Prince (1995).
\newblock Faithfulness and reduplicative identity.
\newblock In: J.~Beckman, L.~W. Dickey \& S.~Urbanczyk (Eds.), {\em Papers in
  Optimality Theory}, Vol.\/~18 von {\em University of Massachusetts Occasional
  Papers in Linguistics},  249--384. Amherst, MA: Graduate Linguistic Student
  Association.
\newblock (ROA-60).

\bibitem[\protect\citeauthoryear{Mohri}{Mohri}{1994}]{mohri:94}
Mohri, M. (1994).
\newblock Compact Representations by Finite-State Transducers.
\newblock In: {\em Proceedings of ACL'94}.
\newblock (CLA-9407003).

\bibitem[\protect\citeauthoryear{Mohri}{Mohri}{1997}]{mohri:97}
Mohri, M. (1997).
\newblock Finite-State Transducers in Language and Speech Processing.
\newblock {\em Computational Linguistics\/}~{\em 23\/}(2).

\bibitem[\protect\citeauthoryear{Mohri, Pereira \& Riley}{Mohri
  et~al.}{1998}]{mohri.pereira.riley:98}
Mohri, M., F.~Pereira \& M.~Riley (1998).
\newblock A rational design for a weighted finite-state transducer library.
\newblock In: D.~Wood \& S.~Yu (Eds.), {\em Automata Implementation. Second
  International Workshop on Implementing Automata, WIA '97}, Vol.\/ 1436 von
  {\em Lecture Notes in Computer Science},  144--58. Springer Verlag.

\bibitem[\protect\citeauthoryear{Mohri, Riley \& Sproat}{Mohri
  et~al.}{1996}]{mohri.riley.sproat:96}
Mohri, M., M.~Riley \& R.~Sproat (1996).
\newblock Algorithms for Speech Recognition and Language Processing.
\newblock (Slides of COLING'96 Tutorial, CLA-9608018).

\bibitem[\protect\citeauthoryear{Neef}{Neef}{1996}]{neef:96}
Neef, M. (1996).
\newblock {\em Wortdesign}.
\newblock T{\"u}bingen: Stauffenburg.

\bibitem[\protect\citeauthoryear{Noyer}{Noyer}{1993}]{noyer:93}
Noyer, R. (1993).
\newblock Mobile Affixes in {H}uave: optimality and morphological
  well-formedness.
\newblock In: E.~Duncan, M.~Hart \& P.~Spaelti (Eds.), {\em Proceedings of the
  Twelfth West Coast Conference on Formal Linguistics}, University of
  California, Santa Cruz.

\bibitem[\protect\citeauthoryear{Partee, ter Meulen \& Wall}{Partee
  et~al.}{1990}]{partee.meulen.wall:90}
Partee, B., A.~ter Meulen \& R.~E. Wall (1990).
\newblock {\em Mathematical Methods in Linguistics}.
\newblock Dordrecht: Kluwer.

\bibitem[\protect\citeauthoryear{Pereira \& Riley}{Pereira \&
  Riley}{1996}]{pereira.riley:96}
Pereira, F.~C. \& M.~D. Riley (1996).
\newblock Speech Recognition by Composition of Weighted Finite Automata.
\newblock Ms., AT\& T Research (CLA-9603001).

\bibitem[\protect\citeauthoryear{Pierrehumbert \& Nair}{Pierrehumbert \&
  Nair}{1996}]{pierrehumbert.nair:96}
Pierrehumbert, J. \& R.~Nair (1996).
\newblock Implications of Hindi prosodic structure.
\newblock In: J.~Durand \& B.~Laks (Eds.), {\em Current Trends in Phonology:
  Models and methods (= Proceedings of the Royaumont meeting 1995)}, Vol.\/~II,
   549--584. European Studies Research Institute, University of Salford
  Publications.

\bibitem[\protect\citeauthoryear{Pollard \& Sag}{Pollard \&
  Sag}{1994}]{pollard.sag:94}
Pollard, C. \& I.~Sag (1994).
\newblock {\em {H}ead-{D}riven {P}hrase {S}tructure {G}rammar}.
\newblock Chicago University Press.

\bibitem[\protect\citeauthoryear{Prince \& Smolensky}{Prince \&
  Smolensky}{1993}]{prince.smolensky:93}
Prince, A. \& P.~Smolensky (1993).
\newblock Optimality Theory. Constraint Interaction in Generative Grammar.
\newblock Technical report RuCCS TR-2, Rutgers University Center for Cognitive
  Science.

\bibitem[\protect\citeauthoryear{Scobbie}{Scobbie}{1991}]{scobbie:91}
Scobbie, J.~M. (1991).
\newblock Towards Declarative Phonology.
\newblock In: S.~Bird (Ed.), {\em Declarative Perspectives on Phonology},
  Vol.\/~7 von {\em Edinburgh Working Papers in Cognitive Science},  1--26.
  Centre for Cognitive Science, University of Edinburgh.

\bibitem[\protect\citeauthoryear{Scobbie, Coleman \& Bird}{Scobbie
  et~al.}{1996}]{scobbie.et.al:96}
Scobbie, J.~M., J.~S. Coleman \& S.~Bird (1996).
\newblock Key Aspects of Declarative Phonology.
\newblock In: J.~Durand \& B.~Laks (Eds.), {\em Current Trends in Phonology:
  Models and methods (= Proceedings of the Royaumont meeting 1995)}, Vol.\/~II,
   685--710. European Studies Research Institute, University of Salford
  Publications.

\bibitem[\protect\citeauthoryear{Selkirk}{Selkirk}{1980}]{selkirk:80}
Selkirk, E. (1980).
\newblock The Role of Prosodic Categories in English Word Stress.
\newblock {\em Linguistic Inquiry\/}~{\em 11\/}(3), 563--605.

\bibitem[\protect\citeauthoryear{Selkirk}{Selkirk}{1982}]{selkirk:82}
Selkirk, E. (1982).
\newblock The syllable.
\newblock In: H.~{van der Hulst} \& N.~Smith (Eds.), {\em The structure of
  phonological representations}, Vol.\/~II,  337--383. Dordrecht: Foris.

\bibitem[\protect\citeauthoryear{Shaw}{Shaw}{1987}]{shaw:87}
Shaw, P.~A. (1987).
\newblock Non-conservation of Melodic Structure in Reduplication.
\newblock In: A.~Bosch (Ed.), {\em Papers from the 23rd Annual Regional Meeting
  of the Chicago Linguistic Society, Part Two: Parasession on Autosegmental and
  Metrical Phonology},  291--306.

\bibitem[\protect\citeauthoryear{Shieber}{Shieber}{1987}]{shieber:87}
Shieber, S.~M. (1987).
\newblock Separating Linguistic Analyses from Linguistic Theories.
\newblock In: P.~Whitelock, M.~M. Wood, H.~L. Somers, R.~Johnson \& P.~Bennett
  (Eds.), {\em Linguistic Theory and Computer Applications},  1--36. London:
  Academic Press.

\bibitem[\protect\citeauthoryear{Sproat}{Sproat}{1992}]{sproat:92}
Sproat, R. (1992).
\newblock {\em Morphology and Computation}.
\newblock Cambridge, Mass.: MIT Press.

\bibitem[\protect\citeauthoryear{Sproat}{Sproat}{1996}]{sproat:96}
Sproat, R. (1996).
\newblock Multilingual Text Analysis for Text-to-Speech Synthesis.
\newblock In: W.~Wahlster (Ed.), {\em Proceedings of ECAI '96. 12th European
  Conference on Artifical Intelligence}. John Wiley \& Sons, Ltd.
\newblock (CLA-9608012).

\bibitem[\protect\citeauthoryear{Stevens}{Stevens}{1968}]{stevens:68}
Stevens, A. (1968).
\newblock {\em Madurese Phonology and Morphology}.
\newblock New Haven: American Oriental society.

\bibitem[\protect\citeauthoryear{Tarjan}{Tarjan}{1983}]{tarjan:83}
Tarjan, R. (1983).
\newblock {\em Data Structures and Network Algorithms}.
\newblock Philadelphia: SIAM.

\bibitem[\protect\citeauthoryear{Tesar}{Tesar}{1995}]{tesar:95}
Tesar, B. (1995).
\newblock Computing optimal forms in Optimality Theory: Basic syllabification.
\newblock Ms., University of Colorado and Rutgers University.

\bibitem[\protect\citeauthoryear{Tesar \& Smolensky}{Tesar \&
  Smolensky}{1993}]{tesar.smolensky:93}
Tesar, B. \& P.~Smolensky (1993).
\newblock The Learnability of Optimality Theory: An Algorithm and some Basic
  Complexity results.
\newblock (ROA-2).
\newblock Ms.

\bibitem[\protect\citeauthoryear{Touretzky \& Wheeler}{Touretzky \&
  Wheeler}{1990}]{touretzky.wheeler:90}
Touretzky, D. \& D.~Wheeler (1990).
\newblock A computational basis for phonology.
\newblock In: D.~Touretzky (Ed.), {\em Advances in Neural Information
  Processing Systems 2: The Collected Papers of the 1989 IEEE Conference on
  Neural Information Processing Systems}. Morgan Kaufmann.

\bibitem[\protect\citeauthoryear{Trommer}{Trommer}{1998}]{trommer:98}
Trommer, J. (1998).
\newblock Optimal Morphology.
\newblock In: M.~Ellison (Ed.), {\em Proceedings of SIGPHON '98, COLING-ACL'98
  post-conference workshop on the Computation of Phonological Constraints,
  August 15}, Montreal, Canada,  26--34.

\bibitem[\protect\citeauthoryear{Trommer}{Trommer}{1999}]{trommer:99}
Trommer, J. (1999, November).
\newblock Mende Tone Patterns Revisited: Tone Mapping as Local Constraint
  Evaluation.
\newblock In: {\em Linguistics in Potsdam}. University of Potsdam.

\bibitem[\protect\citeauthoryear{van Noord}{van Noord}{1997}]{vannoord:97}
van Noord, G. (1997).
\newblock FSA Utilities: A toolbox to manipulate finite-state automata.
\newblock In: D.~Raymond, D.~Wood \& S.~Yu (Eds.), {\em Automata
  Implementation}, Vol.\/ 1260 von {\em Lecture Notes in Computer Science},
  87--108. Springer Verlag.

\bibitem[\protect\citeauthoryear{van Noord \& Gerdemann}{van Noord \&
  Gerdemann}{1999}]{vannord.gerdemann:99}
van Noord, G. \& D.~Gerdemann (1999).
\newblock An Extendible Regular Expression Compiler for Finite-state Approaches
  in Natural Language Processing.
\newblock In: {\em Proceedings of WIA '99}, Potsdam.

\bibitem[\protect\citeauthoryear{Walther}{Walther}{1992}]{walther:92}
Walther, M. (1992).
\newblock {D}eklarative {S}ilbifizierung in einem constraintbasierten
  {G}rammatikformalismus.
\newblock Diplomarbeit Nr. 862, Institut f{\"u}r Informatik/Institut f{\"u}r
  maschinelle Sprachverarbeitung, Universit{\"a}t Stuttgart.

\bibitem[\protect\citeauthoryear{Walther}{Walther}{1993}]{walther:93}
Walther, M. (1993).
\newblock {D}eclarative {S}yllabification with {A}pplications to {G}erman.
\newblock In: T.~M. Ellison \& J.~M. Scobbie (Eds.), {\em Computational
  Phonology}, Vol.\/~8 von {\em Edinburgh Working Papers in Cognitive Science},
   55--79. Centre for Cognitive Science, University of Edinburgh.

\bibitem[\protect\citeauthoryear{Walther}{Walther}{1995}]{walther:95}
Walther, M. (1995).
\newblock {A} {S}trictly {L}exicalized {A}pproach to {P}honology.
\newblock In: J.~Kilbury \& R.~Wiese (Eds.), {\em Proceedings of DGfS/CL'95},
  108--113. {D\"usseldorf}: Deutsche Gesellschaft f{\"u}r Sprachwissenschaft,
  Sektion Computerlinguistik.
  (\url{http:\-//www.\-phil-fak.\-uni-duesseldorf.\-de\-/$\sim$walther\-/slp.p%
s.gz}).

\bibitem[\protect\citeauthoryear{Walther}{Walther}{1996}]{walther:96}
Walther, M. (1996).
\newblock {OT} {SIMPLE} -- {A} construction-kit approach to {O}ptimality
  {T}heory implementation.
\newblock Arbeiten des Sonderforschungsbereichs 282 Nr.~88, Seminar f. Allg.
  Sprachwissenschaft, Universit\"at D\"usseldorf.
\newblock (ROA-152, CLA-9611001).

\bibitem[\protect\citeauthoryear{Walther}{Walther}{1997}]{walther:97}
Walther, M. (1997).
\newblock {\em Deklarative prosodische Morphologie -- constraintbasierte
  Analysen und Computermodelle zum Finnischen und Tigrinya}.
\newblock Ph.D. thesis, Philosophische Fakult{\"a}t der
  Heinrich-Heine-Universit{\"at} D{\"u}sseldorf.
\newblock (Published 1999 by Niemeyer, T{\"u}bingen).

\bibitem[\protect\citeauthoryear{Walther}{Walther}{1998}]{walther:98}
Walther, M. (1998).
\newblock Computing Declarative Prosodic Morphology.
\newblock In: M.~Ellison (Ed.), {\em Proceedings of SIGPHON '98, COLING-ACL'98
  post-conference workshop on the Computation of Phonological Constraints,
  August 15}, Montreal, Canada,  11--20.
\newblock (CLA-9809107).

\bibitem[\protect\citeauthoryear{Werner}{Werner}{1996}]{werner:96}
Werner, A. (1996).
\newblock i-Bildungen im Deutschen.
\newblock Technical report~87, Arbeiten des Sonderforschungsbereichs 282
  `Theorie des Lexikons'.

\bibitem[\protect\citeauthoryear{Wiese}{Wiese}{1995}]{wiese:95}
Wiese, R. (1995).
\newblock {\em The {P}honology of {G}erman}.
\newblock Oxford: Oxford University Press.

\bibitem[\protect\citeauthoryear{Wilbur}{Wilbur}{1973}]{wilbur:73}
Wilbur, R.~B. (1973).
\newblock {\em The Phonology of Reduplication}.
\newblock Ph.D. thesis, University of Illinois, Urbana-Champaign.
\newblock [Reproduced by IULC].

\end{thebibliography}
\end{document}